\RequirePackage{fix-cm}

\documentclass[smallcondensed]{svjour3}     
\smartqed  

\usepackage{natbib}
\usepackage{booktabs}
\usepackage{color}
\usepackage[table]{xcolor}
\usepackage[cmex10]{amsmath}
\usepackage{amsfonts}
\usepackage{amssymb}
\usepackage{mathtools}
\usepackage[utf8]{inputenc}
\usepackage[english]{babel}
\usepackage{array}
\usepackage{mdwmath}
\usepackage{mdwtab}
\usepackage[caption=false,font=footnotesize]{subfig}
\usepackage{float}
\usepackage{multirow}
\usepackage{stfloats}
\usepackage{url}
\usepackage{wrapfig}
\usepackage{bm}
\usepackage{breqn}
\usepackage{etoolbox}
\usepackage{lipsum}
\usepackage{graphicx}
\usepackage[ruled,vlined,noend,linesnumbered]{algorithm2e}
\usepackage{morefloats}
\makeatletter
\patchcmd{\chapter}{\if@openright\cleardoublepage\else\clearpage\fi}{}{}{}
\makeatother

\newcommand{\hypothesis}{\ensuremath{\mathcal{H}}}
\newcommand{\dataset}{\ensuremath{\mathcal{D}}}

\newcommand{\fb}{\ensuremath{\text{FB}_5}}
\newcommand{\trans}{\ensuremath{^T}}

\newcommand{\fisher}{\ensuremath{\mathcal{F}}(\boldsymbol{\Theta})}
\newcommand{\fisherone}{\ensuremath{\mathcal{F}}_1(\boldsymbol{\Theta})}

\newcommand{\fisherij}{\ensuremath{\mathcal{F}}}

\newcommand{\boldx}{\ensuremath{\mathbf{x}}}

\newcommand{\boldtheta}{\ensuremath{\boldsymbol{\Theta}}}

\newcommand{\expect}{\ensuremath{\mathbb{E}}}

\newcommand{\fancyl}{\ensuremath{\mathcal{L}}}
\newcommand{\fancyf}{\ensuremath{\mathcal{F}}}

\newcommand{\calpha}{\ensuremath{\text{C}_{\alpha}}}

\journalname{Machine Learning}

\begin{document}



\title{
Mixtures of Bivariate von Mises Distributions with Applications 
to Modelling of Protein Dihedral Angles
}

\titlerunning{Modelling of angular data using bivariate von Mises distributions}        

\author{Parthan Kasarapu}

\institute{P. Kasarapu  \at
           Faculty of Information Technology, Monash University, VIC 3800, Australia \\
              \email{parthan.kasarapu@monash.edu}           
}

\date{ }

\maketitle

\begin{abstract}

The modelling of empirically observed data is commonly done using
mixtures of probability distributions.
In order to model angular data, directional probability distributions
such as the bivariate von Mises (BVM) is typically used.
The critical task involved in mixture modelling
is to determine the optimal number of component probability distributions.
We employ the Bayesian information-theoretic principle of minimum
message length (MML) to distingush mixture models by balancing 
the trade-off between the model's complexity and its goodness-of-fit to the
data.
We consider the problem of modelling angular data resulting from
the spatial arrangement of protein structures using 
BVM distributions. 
The main contributions of the paper include the development
of the mixture modelling apparatus along with the MML estimation
of the parameters of the BVM distribution.
We demonstrate that statistical inference using the MML framework
supersedes the traditional methods and offers a mechanism to objectively
determine models that are of practical significance. 


\end{abstract}

\section{Introduction}

The efficient and accurate modelling of data is crucial to support reliable 
analyses and to improve the solution to related problems.
Mixture probability distributions are commonly used 
in machine learning applications to model the
underlying, often unknown, distribution of the data 
\citep{titterington1985statistical,mclachlan1988mixture,jain2000statistical}.
They are widely used to describe data arising in various domains such as 
astronomy, biology, ecology, engineering, and economics, amongst many 
others \citep{mclachlan2000finite}.
In order to describe the given data, the problem of selecting a suitable 
statistical model has to be carefully addressed. 

The problem of mixture modelling is associated with the difficult
task of selecting the optimal number of mixture components
and estimating the parameters of the constituent probability distributions.
Mixtures with varying number of component distributions differ
in their model complexities and their goodness-of-fit to the data.
An increase in the complexity of the mixture model, corresponding
to an increase in the model parameters,
leads to better quality of fit to the data.
Various criteria have been proposed to address the trade-off arising
due to these two conflicting objectives
\citep{aic,bic,rissanen1978modeling,icomp,oliver1996unsupervised,roberts,icl,figueiredo2002unsupervised}.
As explained in \citet{multivariate_vmf}, 
these methods are not completely effective in addressing this trade-off
as the model complexity is approximated as a function of the \emph{number}
of parameters and not the actual parameters themselves.
While some of the methods aim to tune the criteria
used to evaluate a mixture model 
\citep{aic,bic,rissanen1978modeling,roberts,icl}, 
they do not provide an associated search strategy to infer the optimal
number of mixture components.
The methods that incorporate a rigorous search method 
for the mixture components
are based on dynamic perturbations of the mixture model 
\citep{ueda2000smem,figueiredo2002unsupervised,multivariate_vmf}.
A thorough review of the various approaches to 
mixture modelling methods and their limitations 
is outlined in \citet{multivariate_vmf}. 

The strategies based on Bayesian inference, and more specifically,
using the minimum message length (MML) framework have increasingly
found support in mixture modelling tasks 
\citep{wallace68,wallace1986improved,roberts,figueiredo2002unsupervised,multivariate_vmf}.
The MML-based inference framework decomposes a modelling
problem into two parts:
the first part determines the model complexity by encoding all
the parameters of the model, and the second part corresponds to
encoding of the observed data using the chosen parameters.
Thus, a two-part message length is obtained for a model under consideration.
A model that results in the least total message length is then
determined to be the optimal model under this framework
\citep{oliver1994mml}.

The MML-based search method developed by \citet{multivariate_vmf} 
is demonstrated
to outperform the traditionally used approaches and is the current
state-of-the art. \citet{multivariate_vmf} have designed the mixture
modelling apparatus to include Gaussian distributions to model data
in the Euclidean space and von Mises-Fisher (vMF) distributions
to model directional data distributed on the surface of a sphere.
While Gaussian mixtures are ubiquitously used 
because of their computational tractability \citep{mclachlan2000finite},
they are ineffective to model directional data.
In this context, analogues of the Gaussian distribution defined on the 
surfaces of the appropriate Reimannian manifolds are typically considered.
The vMF is the most fundamental directional probability distribution
defined on the spherical surface.
It is the spherical analogue of a symmetrical Gaussian wrapped around
a unit hypersphere \citep{fisher1953dispersion,watson1956construction}
and is demonstrated to be useful in large-scale text clustering
\citep{Banerjee:generative-clustering,multivariate_vmf} and gene expression 
analyses \citep{Banerjee:clustering-hypersphere}.
A general form of the vMF distribution is 
the Fisher-Bingham (\fb) distribution which is used to model asymmetrically
distributed data on the spherical surface \citep{kent1982fisher}.
Mixtures of \fb~distributions
have been employed by \citet{peel2001fitting} to identify joint sets in rock masses, and by
\citet{hamelryck2006sampling} to sample random protein conformations.
The \fb~distribution has increasingly
found support in machine learning tasks for structural bioinformatics
\citep{kent2005using,boomsma2006graphical,hamelryck2009probabilistic}.

A two-dimensional version of the vMF distributions called the von Mises
circular is used to model data distributed on the boundary of a circle.
Each data point on the circle has a domain $[-\pi,\pi)$. If such data occur as pairs, then the
resulting manifold in three dimensions would be a torus.
The bivariate von Mises (BVM) distributions are used to model
such data distributed on the toroidal surface and serve as the Gaussian
analogue. Motivated by its practical applications in bioinformatics,
the BVM distributions are widely studied.
The mixtures of BVM distributions have been previously used in modelling protein
dihedral angles \citep{dowe1996circular,mardia2007protein,mardia2008multivariate}.
However, these approaches have some limitations.
\citet{dowe1996circular} treat the pairs of angles to be independent
of each other and do not account for their correlation. This is akin
to conflating two von Mises circular distributions together.
As explained in Section~\ref{sec:bvm_sine_mixtures}, such an approximation
leads to inefficient mixtures. 
Although \citet{mardia2007protein} use BVM distributions that 
account for the correlation between the angular pairs, they do not
have a rigorous search method to determine the optimal mixtures.
This limits their ability to correctly distinguish among models 
that, while being of different type, 
have the same number of model parameters.

This paper develops the mixture modelling apparatus to address these limitations
using the MML framework.
Further, different variants of the BVM distribution
obtained by constraining some of its characterizing parameters
(see Section~\ref{sec:bivariate_vonmises}).
\citet{mardia2007protein} have evaluated the utility of these variants
in the context of modelling the protein dihedral angles.
We adopt the MML principle in objectively assessing the 
mixture distributions of these variants (see Section~\ref{sec:bvm_sine_mixtures}).
We have developed
a search method to determine the 
optimal number of mixture components and their parameters that describe the given data
in a completely unsupervised setting.
The use of the MML modelling paradigm and our proposed search method
is explored on 
real-world data corresponding to the dihedral, that is, torsion angles
of protein structures. 
We demonstrate that mixtures of BVM distributions facilitate the
design of reliable computational models for protein structural data. 

In addition to determining the optimal number of mixture components,
the parameters of the individual component distributions need to be estimated.
Traditionally, the optimum parameters are obtained by maximum likelihood 
(ML) or Bayesian maximum \emph{a posteriori} probability (MAP) estimation. 
For a mixture distribution, the parameters are estimated by
maximizing the likelihood
of the data by employing an expectation-maximization (EM) algorithm
that iteratively updates the mixture parameters \citep{dempster1977maximum}.
The key differences between ML, MAP and MML-based estimation is:
(1) unlike ML, MML uses a prior over the parameters and considers their precision while encoding; 
(2) unlike MAP, MML estimators are invariant under 
non-linear transformations of the parameters \citep{oliver1994mml}.
The estimation of parameters using ML ignores the cost of stating the parameters, and MAP based
estimation uses the probability \textit{density} of parameters instead of their
probability measure. In contrast, the MML inference process takes into
account the optimal precision to which parameters should be stated 
and uses it to determine a corresponding probability value. 
Parameter estimation using the MML framework has been carried out on
various probability distributions \citep{WallaceBook}.
\citet{multivariate_vmf} have demonstrated that the MML estimators
outperform the traditionally used estimators in the case of Gaussian
and vMF distributions.
Furthermore, for a \fb~distribution, \citet{kent_arxiv} have shown
that the MML estimators have lower bias and error as compared to
the ML and MAP estimators. 

\noindent\textbf{Contributions:} The main contributions of this paper are 
as follows:
\begin{itemize}
\item We derive the MML-based estimates of the parameters of a BVM 
distribution.
The MML estimators are demonstrated to have lower bias and mean squared
error when compared to their traditional counterparts.
We consider two variants of the BVM distribution, namely the
Independent \citep{dowe1996circular} and the 
Sine variant \citep{singh2002probabilistic}.

\item We design a search method to infer the optimal number of BVM mixture
components that best describe the angular data distributed on the
toroidal surface.

\item We demonstrate the utility of the MML framework in determining
the suitability of the two variants of the BVM distribution
in modelling the protein dihedral angle data.
We show that the Sine variant that includes the correlation term
explains the data much more effectively
than the Independent version.

\item We demonstrate the effectiveness of the mixture modelling method
by applying it to cluster protein dihedral angles. We demonstrate that
the resulting mixtures closely correspond to the commonly observed
secondary structural regions in protein structures.
\end{itemize}

The rest of the chapter is organized as follows:
Section~\ref{sec:bivariate_vonmises} describes the BVM distribution,
the Independent and the Sine variants,
and their relevance in modelling data distributed on the toroidal surface.
Section~\ref{sec:mml_framework} describes the 
MML framework and outlines the differences between parameter
estimation using ML, MAP and MML methods.
It includes the derivation of the MML estimators
of the parameters of the BVM distribution.
We empirically demonstrate that the MML estimators outperform 
the traditionally used ML and MAP estimators by having lower bias 
and mean squared error.
Section~\ref{sec:bvm_sine_mixtures} discusses the search and inference of 
mixtures of bivariate von Mises (BVM) distributions
using the MML framework. 
As a specific application, we employ the
mixtures to model protein dihedral angles. 
We demonstrate that our search method is able to infer meaningful
clusters that directly correspond to frequently occuring
conformations in protein structures.

\section{Bivariate von Mises probability distribution}
\label{sec:bivariate_vonmises}

The class of bivariate von Mises (BVM) distributions was introduced by
\citet{mardia1975statistics2,mardia1975statistics} to model data 
distributed on the surface of a 3D torus. The study of these
distributions has been partly motivated by biological research,
where it is required to model the protein dihedral angles
(see Section~\ref{sec:bvm_modelling_dihedrals}).
The probability density function of the BVM distribution has the general form
\begin{equation}
f(\boldx;\boldtheta) \propto \exp\{\kappa_1 \cos(\theta_1 - \mu_1) 
                                   + \kappa_2 \cos(\theta_2 - \mu_2)
+ (\cos\theta_1, \,\sin\theta_1) \mathbf{A} (\cos\theta_2, \,\sin\theta_2)\trans\}
\label{eqn:bvm_density}
\end{equation}
where $\boldx=(\theta_1,\theta_2)$, such that $\theta_1,\theta_2\in[-\pi,\pi)$
and the parameter vector $\boldtheta=(\mu_1,\mu_2,\kappa_1,\kappa_2,\mathbf{A})$, such that
$\mu_1,\mu_2\in[-\pi,\pi)$ are the mean angles, $\kappa_1\ge 0$ and $\kappa_2\ge 0$ are the 
concentration parameters, and $\mathbf{A}$ is a $2\times 2$ real-valued matrix.
The term $\exp\{\kappa_1 \cos(\theta_1 - \mu_1)\}$ corresponds to a 
von Mises distribution on a circle
characterized by the parameters $\mu_1$ and $\kappa_1$,
Hence, the BVM distribution (Equation~\ref{eqn:bvm_density}) can be explained as a 
product of two von Mises circular distributions, with an additional
exponential term involving $\mathbf{A}$, that accounts for the correlation.

The general form of the BVM distribution has 8 free parameters.
In order to draw an analogy to the bivariate Gaussian distribution
(with 5 free parameters), sub-models of the BVM distribution have been
proposed by restricting the values that $\mathbf{A}$ can take \citep{jupp1980general}.
A 6-parameter version was explored by \citet{rivest1988distribution} and has the form
\begin{align}
f(\boldx;\boldtheta) \propto \exp\{&\kappa_1 \cos(\theta_1 - \mu_1) 
                                   + \kappa_2 \cos(\theta_2 - \mu_2)
+ \alpha \cos(\theta_1 - \mu_1)\cos(\theta_2 - \mu_2) \notag\\ 
&+ \beta \sin(\theta_1 - \mu_1)\sin(\theta_2 - \mu_2)\}
\label{eqn:bvm_rivest}
\end{align}   
In particular, when $\alpha=0$ and $\beta=\lambda$, the above density reduces to
the following 5-parameter version, which is called the BVM \emph{Sine} model \citep{singh2002probabilistic}.
\begin{equation}
f(\boldx;\boldtheta) = c(\kappa_1,\kappa_2,\lambda)^{-1} \exp\{\kappa_1 \cos(\theta_1 - \mu_1) 
                                   + \kappa_2 \cos(\theta_2 - \mu_2)
+ \lambda \sin(\theta_1 - \mu_1)\sin(\theta_2 - \mu_2)\} 
\label{eqn:bvm_sine}
\end{equation}
where $c(\kappa_1,\kappa_2,\lambda)$ is the normalization constant of the distribution
defined as
\begin{equation}
\quad c(\kappa_1,\kappa_2,\lambda) 
= 4\pi^2 \sum_{j=0}^{\infty} \binom{2j}{j} \left(\frac{\lambda^2}{4\kappa_1\kappa_2} \right)^j I_j(\kappa_1)I_j(\kappa_2)
\label{eqn:bvm_sine_norm_constant}
\end{equation}
and $I_v$ is the modified Bessel function of first kind and order $v$.
The 5-parameter vector will be $\boldtheta=(\mu_1,\mu_2,\kappa_1,\kappa_2,\lambda)$
where $\lambda$ is a real number. If $\lambda=0$, the probability density
function (Equation~\ref{eqn:bvm_sine}) will just be the product of two independent
von Mises circular distributions, and corresponds to the case when there is
no correlation between the two variables $\theta_1$ and $\theta_2$.
The probability density function in such a case is given as 
\begin{equation}
f(\boldx;\boldtheta) = c(\kappa_1,\kappa_2)^{-1} \exp\{\kappa_1 \cos(\theta_1 - \mu_1) 
                                   + \kappa_2 \cos(\theta_2 - \mu_2)\}
\label{eqn:bvm_ind}
\end{equation}
where $c(\kappa_1,\kappa_2)$ is the normalization constant defined as
$c(\kappa_1,\kappa_2) = \dfrac{1}{2\pi I_0(\kappa_1)}\dfrac{1}{2\pi I_0(\kappa_2)}$.
and corresponds to the product of
the normalization constants
for the respective von Mises circular distributions. 

Alternatively, when $\alpha=-\beta$, the form of Equation~\ref{eqn:bvm_rivest}
results in a different reduced form called the BVM \emph{Cosine} model \citep{mardia2007protein}.
The Sine and the Cosine models serve as natural analogues of the bivariate
Gaussian distribution on the 3D torus. In fact, for huge concentrations, 
\citet{singh2002probabilistic} approximate 
the Sine model to a bivariate Gaussian distribution with the $2\times 2$ covariance matrix 
$\mathbf{C} = [c_{ij}], i,j\in\{1,2\}$, whose elements are
given by
\begin{equation*} 
c_{11} = \frac{\kappa_2}{\kappa_1\kappa_2 - \lambda^2}, \quad
c_{22} = \frac{\kappa_1}{\kappa_1\kappa_2 - \lambda^2}, \quad
c_{12} = c_{21} = \frac{\lambda}{\kappa_1\kappa_2 - \lambda^2}
\end{equation*} 
The limiting case approximation is valid when $\kappa_1\kappa_2>\lambda^2$. Also, from the
covariance matrix, the correlation coefficient $\rho$ can be determined as \citep{pearson1895note}:
\begin{equation}
\rho = \frac{c_{12}}{\sqrt{c_{11} c_{22}}} = \frac{\lambda}{\sqrt{\kappa_1\kappa_2}}
\quad\text{such that}\quad
|\rho| < 1
\label{eqn:bvm_sine_rho}
\end{equation}

\begin{figure}[ht]
\centering
\subfloat[$\rho=0.1$]{\includegraphics[width=0.3\textwidth]{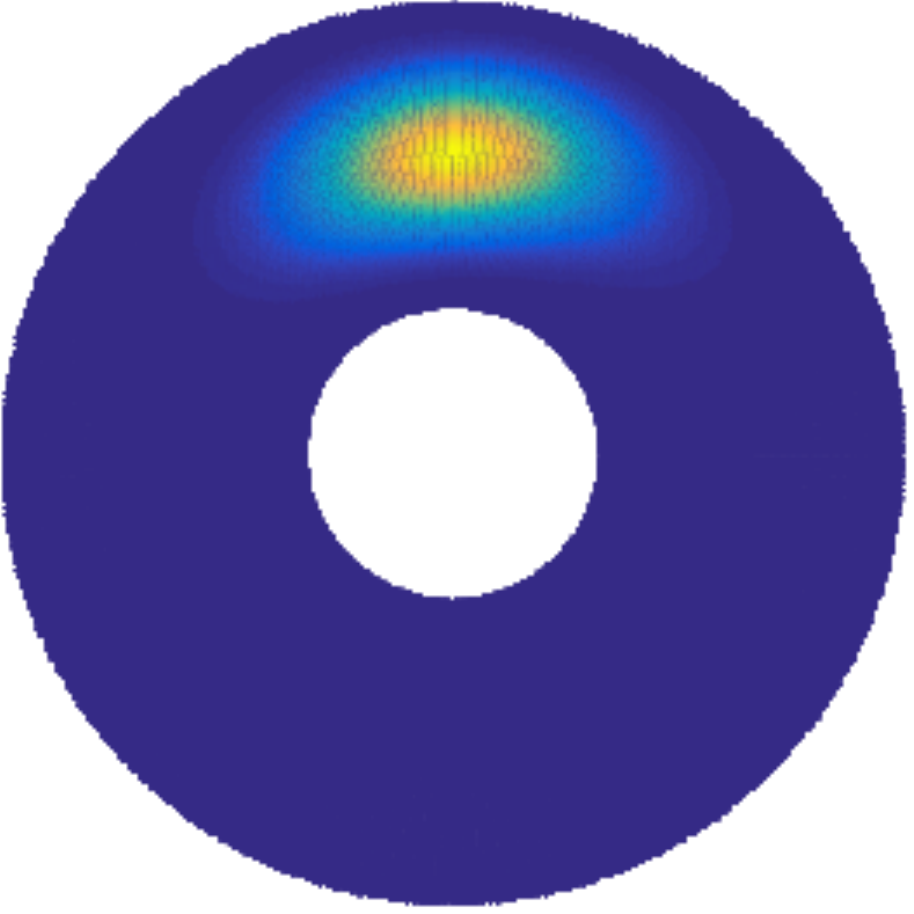}}\quad
\subfloat[$\rho=0.5$]{\includegraphics[width=0.3\textwidth]{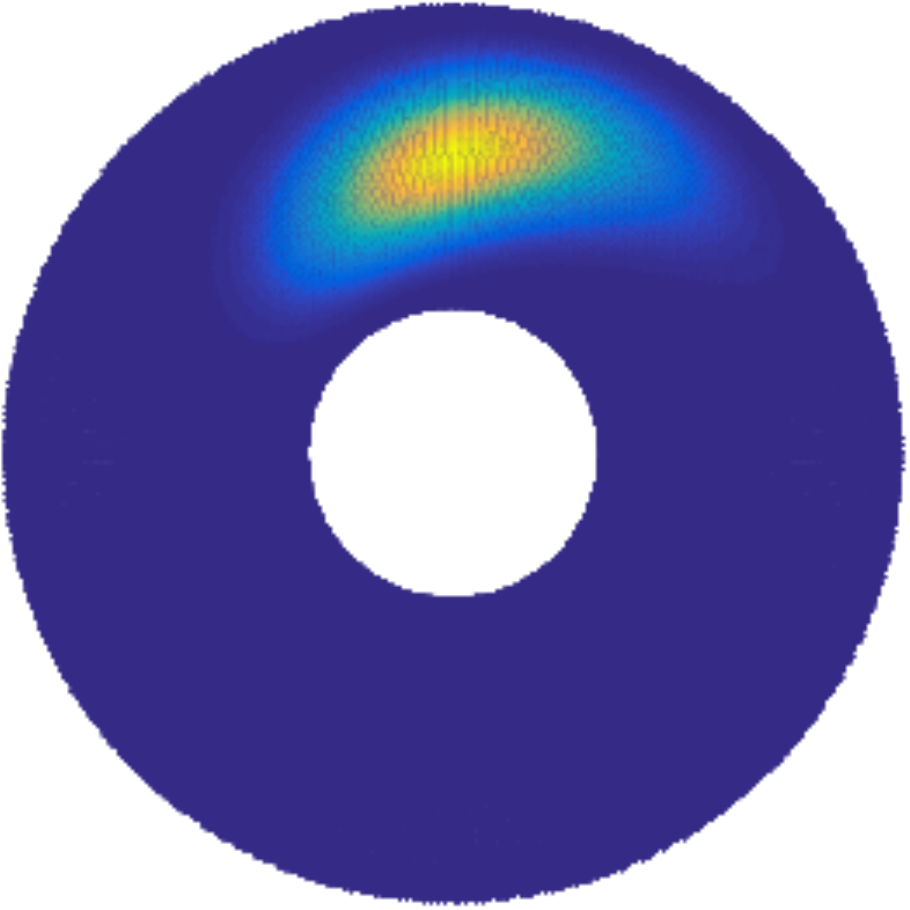}}\quad
\subfloat[$\rho=0.9$]{\includegraphics[width=0.3\textwidth]{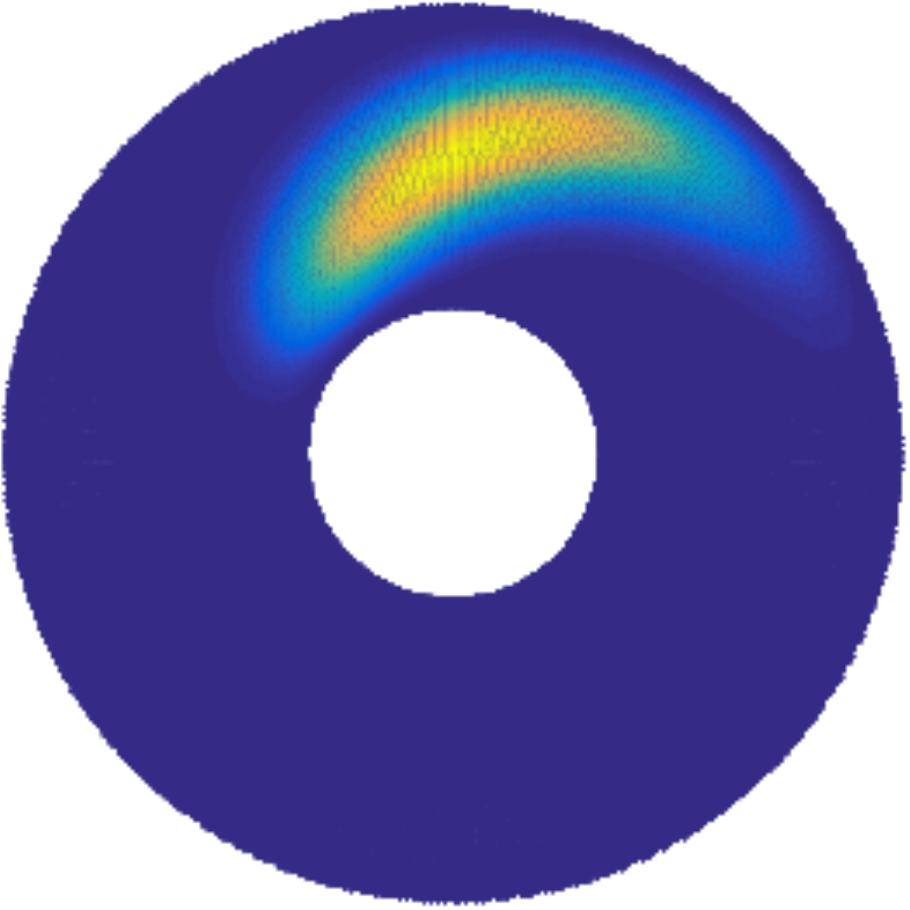}}
\caption[BVM Sine model showing different correlations.]
        {BVM Sine model showing different correlations.
         The distribution has $\mu_1=\mu_2=\frac{\pi}{2}$ and $\kappa_1=\kappa_2=10$.
         For each value of $\rho$, the corresponding value of $\lambda = \rho \sqrt{\kappa_1\kappa_2}$.
         } 
\label{fig:torus_diff_rho}
\end{figure}

In order to better understand the interaction of $\kappa_1,\kappa_2,$ and 
the correlation coefficient $\rho$,
we provide an example in Figure~\ref{fig:torus_diff_rho},
where the distribution is shown for values 
of $\rho=0.1$ (low correlation), $\rho=0.5$ (moderate correlation),
and $\rho=0.9$ (high correlation).
Note that $\rho$ can take negative values, in which case the resultant distribution
will just be a reflection in some axis \citep{mardia2007protein}.

The modelling of directional data using the BVM Sine and Cosine models has
been previously explored by \citet{mardia2007protein}.
For estimating the parameters of the distribution, maximum likelihood
based optimization is used. 
We discuss ML and MAP based estimators, which are the traditionally
used methods of parameter estimation.

\subsection{Maximum likelihood parameter estimation}
\label{subsec:bvm_sine_mle}

In applications involving modelling directional data using the BVM Sine
distributions, the maximum likelihood (ML) estimates are typically used
\citep{boomsma2006graphical,mardia2007protein,mardia2008multivariate}. 
For BVM Sine distributions, the moment and ML estimates
are the same, as the BVM Sine distribution belongs to the exponential
family of distributions \citep{mardia2008multivariate}.


Given data $\dataset = \{\boldx_1,\ldots,\boldx_N\}$, where $\boldx_i = (\theta_{i1},\theta_{i2})$,
the ML estimates of the parameter vector $\boldtheta=(\mu_1,\mu_2,\kappa_1,\kappa_2,\lambda)$
are obtained by minimizing the negative log-likelihood
expression of the data given by
\begin{align}
\fancyl(\dataset|\boldtheta) = &N\log c(\kappa_1,\kappa_2,\lambda) 
- \kappa_1 \sum_{i=1}^N \cos(\theta_{i1} - \mu_1)
- \kappa_2 \sum_{i=1}^N \cos(\theta_{i2} - \mu_2) \notag\\
&- \lambda \sum_{i=1}^N \sin(\theta_{i1} - \mu_1)\sin(\theta_{i2} - \mu_2)
\label{eqn:bvm_sine_negloglike_data}
\end{align}

The ML estimates satisfy $\dfrac{\partial\fancyl}{\partial\boldtheta} = 0$.
However, as no closed form soultions exist because of the complicated
form of $c(\kappa_1,\kappa_2,\lambda)$, an optimization library is used.
We use NLopt\footnote{\url{http://ab-initio.mit.edu/nlopt}},
a non-linear optimization library,
to compute the parameter estimates.

\subsection{Maximum \emph{a posteriori} probability (MAP) estimation}
\label{subsec:bvm_sine_map}

For an independent and identically distributed sample $\dataset$,
the MAP estimates are obtained by maximizing the posterior density $\Pr(\boldtheta|\dataset)$.
This requires the definition of a reasonable prior $\Pr(\boldtheta)$ on 
the parameter space.
The MAP estimates are sensitive to the nature of parameterization
of the probability distribution and this limitation is discussed here.
We demonstrate that the MAP estimators are inconsistent and
are subjective to the parameterization.
We consider two alternative parameterizations in the case of the
BVM Sine distribution. \\

\noindent\emph{Prior on the angular parameters $\mu_1$ and $\mu_2$:}
Since $\mu_1,\mu_2 \in [-\pi,\pi)$, a uniform prior can be assumed in this
range for each of the means. Further, assuming $\mu_1$ and $\mu_2$ to be
independent of each other, their joint prior will be
$\Pr(\mu_1,\mu_2) = \dfrac{1}{4\pi^2}$. \\

\noindent\emph{Prior on the scale parameters $\kappa_1, \kappa_2$, and $\lambda$:}
As discussed for Equation~\ref{eqn:bvm_density},
the BVM density function can be regarded as a product of two von Mises
circular distributions with an additional term that captures 
the correlation. In the Bayesian analysis of the von Mises
circular distribution, \citet{wallace1994estimation} used the prior
on the concentration parameter $\kappa$ as $\Pr(\kappa) = \dfrac{\kappa}{(1+\kappa^2)^{3/2}}$.
In the current context of defining priors on $\kappa_1$ and $\kappa_2$
for a BVM distribution, we use the prior $\Pr(\kappa)$. Assuming
$\kappa_1$ and $\kappa_2$ to be independent of each other, the joint prior
is given by
\begin{equation*}
\Pr(\kappa_1,\kappa_2) = \frac{\kappa_1\kappa_2}{(1+\kappa_1^2)^{3/2}(1+\kappa_2^2)^{3/2}}
\end{equation*}

In order to define a reasonable prior on $\lambda$, we use
the fact that $\lambda^2 < \kappa_1 \kappa_2$ (see Equation~\ref{eqn:bvm_sine_rho}).
Hence, the conditional probability density of $\lambda$
is given as: $\Pr(\lambda|\kappa_1,\kappa_2) = \dfrac{1}{2\sqrt{\kappa_1\kappa_2}}$.
Therefore, the joint prior density of the scalar parameters 
$\kappa_1, \kappa_2$ and $\lambda$ is
\begin{equation*}
\Pr(\kappa_1,\kappa_2,\lambda) = \Pr(\kappa_1,\kappa_2)\Pr(\lambda|\kappa_1,\kappa_2) 
= \frac{\sqrt{\kappa_1\kappa_2}}{2(1+\kappa_1^2)^{3/2}(1+\kappa_2^2)^{3/2}}
\end{equation*}

Using the product of 
the priors for the angular and the scale parameters, that is,
$\Pr(\mu_1,\mu_2)$ and $\Pr(\kappa_1,\kappa_2,\lambda)$,
the joint prior of the parameter vector $\boldtheta$, is given by
\begin{equation}
\Pr(\boldtheta) = \Pr(\mu_1,\mu_2,\kappa_1,\kappa_2,\lambda) 
= \frac{\sqrt{\kappa_1\kappa_2}}{8\pi^2(1+\kappa_1^2)^{3/2}(1+\kappa_2^2)^{3/2}}
\label{eqn:bvm_sine_prior1}
\end{equation}
The prior density $\Pr(\boldtheta)$ can be used along with the likelihood
function to formulate the posterior density 
as the product of the prior and the likelihood function, that is,
$\Pr(\boldtheta|\dataset) \propto \Pr(\boldtheta) \Pr(\dataset|\boldtheta)$.
The MAP estimates correspond to the 
maximized value of the posterior $\Pr(\boldtheta|\dataset)$.

\subsubsection{Non-linear transformations of the parameter space}

We consider non-linear transformations of the parameter space,
in order to demonstrate that the MAP estimates
are not invariant in different parameterizations of the probability distribution.
We discuss a simple 
non-linear transformation of the parameter space involving the
correlation parameter $\lambda$.
Additionally, we also describe a parameterization that transforms
all the five parameters. \\

\noindent\emph{An alternative parameterization involving $\lambda$:}
The BVM Sine probability density function (Equation~\ref{eqn:bvm_sine})
can be reparameterized in terms of the correlation coefficient $\rho$, instead of $\lambda$,
by using the relationship $\lambda = \rho\sqrt{\kappa_1\kappa_2}$ 
(as per Equation~\ref{eqn:bvm_sine_rho}).
If $\boldtheta' = (\mu_1,\mu_2,\kappa_1,\kappa_2,\rho)$ denotes the modified vector
of parameters, the modified prior density $\Pr(\boldtheta')$ is
obtained by dividing $\Pr(\boldtheta)$ with the Jacobian of the transformation
$J = \dfrac{\partial\rho}{\partial\lambda} = \dfrac{1}{\sqrt{\kappa_1\kappa_2}}$
as follows
\begin{equation}
\Pr(\boldtheta') = \frac{\Pr(\mu_1,\mu_2,\kappa_1,\kappa_2,\lambda)}{J}
= \frac{\kappa_1\kappa_2}{8\pi^2(1+\kappa_1^2)^{3/2}(1+\kappa_2^2)^{3/2}}
\label{eqn:bvm_sine_prior2}
\end{equation}
With this transformation, the posterior density $\Pr(\boldtheta'|\dataset)$
can be computed, and
subsequently used to determine the MAP estimates. \\

\noindent\emph{An alternative parameterization involving $\boldtheta$:}
In addition to the transformation of the correlation parameter $\lambda$, we study another 
transformation that was proposed by \citet{rosenblatt1952remarks}. The method transforms
a given continuous $k$-variate probability distribution into the uniform distribution
on the $k$-dimensional \emph{unit} hypercube. Such a transformation applied on the 
prior density of the parameter vector $\boldtheta$ results in
the prior transforming to a uniform distribution. Hence, estimation in this
transformed parameter space is equivalent to the corresponding 
maximum likelihood estimation.

For the 5-parameter vector $\boldtheta=(\mu_1,\mu_2,\kappa_1,\kappa_2,\lambda)$, 
the \citet{rosenblatt1952remarks} 
transformation to $\boldtheta''=(z_1,z_2,z_3,z_4,z_5)$ 
involves computing the cumulative densities $F_i,\,\forall\,i\in\{1,\ldots,5\}$
as follows
\begin{align*}
z_1 &= \Pr(X_1 \le \mu_1) = F_1(\mu_1)\\
z_2 &= \Pr(X_2 \le \mu_2 | X_1 = \mu_1) = F_2(\mu_2|\mu_1)\\
z_3 &= \Pr(X_3 \le \kappa_1 | X_2 = \mu_2, X_1 = \mu_1) = F_3(\kappa_1|\mu_2,\mu_1)\\
z_4 &= \Pr(X_4 \le \kappa_2 | X_3 = \kappa_1, X_2 = \mu_2, X_1 = \mu_1) = F_4(\kappa_2|\kappa_1,\mu_2,\mu_1)\\
z_5 &= \Pr(X_5 \le \lambda| X_4 = \kappa_2, X_3 = \kappa_1, X_2 = \mu_2, X_1=\mu_1) = F_5(\lambda|\kappa_2,\kappa_1,\mu_2,\mu_1)
\end{align*}
As the cumulative densities are bounded by 1,
the above transformation results in $0\le z_i \le 1, \,i = \{1,\ldots,5\}$. 
Further, \citet{rosenblatt1952remarks} argue that each $z_i$ is uniformly 
and independently distributed on $[0,1]$, so that the prior density 
in this transformed parameter space is 
\begin{equation}
\Pr(\boldtheta'') = \Pr(z_1,z_2,z_3,z_4,z_5) = 1
\label{eqn:bvm_sine_prior3}
\end{equation}
In order to achieve such a transformation, we need to express $z_i$ in terms of 
the original parameters. Based on the assumptions made in the 
formulation of the prior $\Pr(\boldtheta)$,
we derive the following relationships:
\begin{equation*}
z_1 = \int_{-\pi}^{\mu_1} \frac{1}{2\pi}\, d\mu_1 = \frac{\mu_1+\pi}{2\pi} \implies \mu_1 = \pi (2z_1-1)
\quad\text{and}\quad
z_2 = \frac{\mu_2+\pi}{2\pi} \implies \mu_2 = \pi (2z_2-1)
\end{equation*}
Based on the independence assumption in the formulation of priors of angular
and scale parameters, we have
$z_3 = F_3(\kappa_1|\mu_2,\mu_1) = F_3(\kappa_1)$,
and therefore we have
\begin{gather*}
z_3 = \int_{0}^{\kappa_1} \Pr(\kappa) \,d\kappa
= \int_{0}^{\kappa_1} \frac{\kappa}{(1+\kappa^2)^{3/2}} d\kappa = 1 - \cos(\arctan\kappa_1) \\
\text{Hence,}\,\,
\kappa_1 = \tan(\arccos(1-z_3))
\quad\text{and}\quad
\kappa_2 = \tan(\arccos(1-z_4))
\end{gather*}
Further, $F_5(\lambda|\kappa_2,\kappa_1,\mu_2,\mu_1) = F_5(\lambda|\kappa_2,\kappa_1)$,
as $\lambda$ is independent of $\mu_1$ and $\mu_2$.
Hence, the invertible transformation corresponding to
$\lambda$ is as follows
\begin{equation*}
z_5 = F_5(\lambda|\kappa_2,\kappa_1)
    = \int_{-\sqrt{\kappa_1\kappa_2}}^{\lambda} \frac{1}{2\sqrt{\kappa_1\kappa_2}}\,d\lambda
    = \frac{1}{2} \left( \frac{\lambda}{\sqrt{\kappa_1\kappa_2}} + 1 \right)
\end{equation*}
so that $\lambda$ can be expressed as a function of $z_3,z_4$, and $z_5$. The transformed
BVM Sine probability density function $f(\boldx,\boldtheta'')$ is obtained by
substituting the expressions of $\boldtheta$ in terms of $z_i, 1\le i \le 5$
in $f(\boldx,\boldtheta)$ (Equation~\ref{eqn:bvm_sine}).

In summary, we considered two additional parameterizations of the BVM Sine
probability density. For statistical invariance, the estimates of the parameters
should also be affected by the same transformation in alternative parameterizations.
The MAP estimation does not satisy this property, as illustrated
by the following example.

\subsubsection{An example demonstrating the effects of alternative parameterizations} 
\label{subsubsec:bvm_sine_map_example}

An example of estimating parameters using the posterior distributions
resulting from the various prior densities 
(Equations~\ref{eqn:bvm_sine_prior1} - \ref{eqn:bvm_sine_prior3})
is described here.
A random sample of size $N=10$ is generated from a BVM Sine distribution \citep{singh2002probabilistic}.
The true parameters of the distribution are $\mu_1 = \mu_2 = \pi/2$, 
$\kappa_1 = \kappa_2 = 10,$ and $\lambda = 9$ (corresponding to 
a correlation coefficient of $\rho = 0.9$).

The MAP estimators are obtained by maximizing the posterior
densities using
the non-linear optimization library NLopt 
\citep{nlopt} in conjunction with derivative-free
optimization \citep{powell1994direct}. 
The differences in the estimates are explained below. 

We observe that the estimates of the angular parameters, $\mu_1$
and $\mu_2$, are similar across the different parameterizations,
with values close to 1.730 and 1.695 radians respectively.
In the case of using $\boldtheta''$, the estimated values 
$\widehat{z}_1$ and $\widehat{z}_2$ are transformed back into $\widehat{\mu}_1$
and $\widehat{\mu}_2$ to allow comparison of similar quantities.
\begin{gather*}
\widehat{\mu}_1 = 1.730, \,\,\widehat{\mu}_2 = 1.695 \,\,\,\text{using } \Pr(\boldtheta) \\
\widehat{\mu}_1 = 1.731, \,\,\widehat{\mu}_2 = 1.696 \,\,\,\text{using } \Pr(\boldtheta') \\
\widehat{z}_1 = 0.276, \widehat{z}_2 = 0.270 \implies\widehat{\mu}_1 = 1.735, \,\,\widehat{\mu}_2 = 1.698 \,\,\,\text{using } \Pr(\boldtheta'')
\end{gather*}

The estimation of the scale parameters, $\kappa_1, \kappa_2,$ and $\lambda$ however,
results in different values. We observe that, in the case of $\Pr(\boldtheta'), \widehat{\rho} = 0.684$,
which translates to $\widehat{\lambda} = 6.565$. This is different from
the estimated value of $\widehat{\lambda} = 5.017$ using $\Pr(\boldtheta)$.
The values of $\widehat{\kappa}_1$ and $\widehat{\kappa}_2$ are also
different. Further, with $\Pr(\boldtheta'')$, the transformation
of estimated $z_i$ into the $\boldtheta$ parameter space result in
different estimates.
\begin{gather*}
\widehat{\kappa}_1 = 4.451, \,\,\widehat{\kappa}_2 = 14.158, \,\,\widehat{\lambda} = 5.017 \,\,\,\text{using } \Pr(\boldtheta) \\
\widehat{\kappa}_1 = 5.311, \,\,\widehat{\kappa}_2 = 17.338, \,\,\widehat{\rho} = 0.684 \implies \widehat{\lambda} = 6.565 \,\,\,\text{using } \Pr(\boldtheta') \\
\widehat{z}_3 = 0.900, \widehat{z}_4 = 0.970, \widehat{z}_5 = 0.924 \implies\widehat{\kappa}_1 = 9.998, \,\,\widehat{\kappa}_2 = 33.931, \,\,\widehat{\lambda} = 15.628 \,\,\,\text{using } \Pr(\boldtheta'')
\end{gather*}

The above example demonstrates a drawback of the MAP-based estimation
with respect to parameter invariance. 
The MAP estimator corresponds to the mode of the posterior distribution.
The mode is, however, not invariant under varying parameterizations.
We use the above parameterizations in analyzing the behaviour of 
the various estimators 
in the experiments section (Section~\ref{subsec:bvm_sine_experiments}).

\section{Minimum Message Length (MML) Inference}
\label{sec:mml_framework}

In this section, we describe the model selection paradigm using the
Minimum Message Length criterion and proceed to give an overview
of MML-based parameter estimation for any distribution.

\subsection{Model selection using minimum message length criterion}
\citet{wallace68} developed the first practical criterion
for model selection based on information theory.
As per Bayes's theorem:
\[\Pr(\hypothesis\&\dataset) = \Pr(\hypothesis) \times \Pr(\dataset|\hypothesis) = \Pr(\dataset) \times \Pr(\hypothesis|\dataset)\]
where \dataset~denotes observed data, and \hypothesis~some
hypothesis about that data. Further, $\Pr(\hypothesis\&\dataset)$ is the joint probability
of data \dataset~and hypothesis \hypothesis, 
$\Pr(\hypothesis)$ and $\Pr(\dataset)$ are the prior probabilities of
hypothesis \hypothesis~and data \dataset~respectively, $\Pr(\hypothesis|\dataset)$
is the posterior probability, and $\Pr(\dataset|\hypothesis)$ is the
likelihood.  

As per \citet{shannon1948},
given an event $E$ with probability
$\Pr(E)$, the length of the optimal lossless code to represent that
event requires $I(E) = -\log_2 (\Pr(E))$ bits.  
Applying Shannon's insight to
Bayes's theorem, \citet{wallace68} got the following relationship 
between conditional probabilities in terms of optimal message lengths: 
\begin{equation*} 
I(\hypothesis\&\dataset) = I(\hypothesis) + I(\dataset|\hypothesis) = I(\dataset) + I(\hypothesis|\dataset) 
\end{equation*}
The above equation 
can be intrepreted as the \emph{total} cost to encode a
message comprising of the following two parts:
\begin{enumerate}
\item \emph{First part:} the hypothesis $\hypothesis$, which takes $I(\hypothesis)$ bits,
\item \emph{Second part:} the observed data $\dataset$ using knowledge of $\hypothesis$, which takes $I(\dataset|\hypothesis)$ bits.
\end{enumerate}
As a result, given two competing hypotheses \hypothesis~and $\hypothesis^\prime$, 
\begin{gather*}
\Delta I = I(\hypothesis\&\dataset) - I(\hypothesis^\prime\&\dataset) = I(\hypothesis|\dataset) - I(\hypothesis^\prime|\dataset)\quad\text{bits.}\\
\text{Hence,}\,\Pr(\hypothesis^\prime|\dataset) = 2^{\Delta I} \Pr(\hypothesis|\dataset) 
\end{gather*}
gives the log-odds posterior ratio between the two hypotheses.
The framework provides a rigorous means to objectively compare two
competing hypotheses. 
Clearly, the message length can vary depending on the complexity of \hypothesis~and
how well it can explain \dataset. 
A more complex $\hypothesis$ may explain $\dataset$
better but takes more bits to be stated itself.  The trade-off comes
from the fact that (hypothetically) transmitting the message requires the encoding of both
the hypothesis and the data given the hypothesis, that is, the model
complexity $I(\hypothesis)$ and the goodness of fit $I(\dataset|\hypothesis)$.

\subsection{MML-based parameter estimation}  
\label{subsec:mml_parameter_estimation}

\citet{wallace87} introduced a generalized framework to estimate a set of 
parameters $\boldsymbol{\Theta}$ given data \dataset. The method
requires a reasonable prior $h(\boldsymbol{\Theta})$ on the hypothesis and 
evaluating the
\textit{determinant} of the Fisher information matrix $|\fisher|$ of the 
\textit{expected} second-order partial derivatives of the negative 
log-likelihood function, $\mathcal{L}(D|\boldsymbol{\Theta})$. 
The parameter vector $\boldsymbol{\Theta}$ that minimizes 
the message length expression (given by Equation~\ref{eqn:two_part_msg})
is the MML estimate according to \citet{wallace87}. 
\begin{equation}
I(\boldsymbol{\Theta},\dataset) = \underbrace{\frac{d}{2}\log q_d -\log\left(\frac{h(\boldsymbol{\Theta})}{\sqrt{|\mathcal{F}(\boldsymbol{\Theta})|}}\right)}_{\mathrm{I(\boldsymbol{\Theta})}} 
+ \underbrace{\mathcal{L}(\dataset|\boldsymbol{\Theta}) + \frac{d}{2}}_{\mathrm{I(\dataset|\boldsymbol{\Theta})}}
\label{eqn:two_part_msg}
\end{equation}
where $d$ is the number of free parameters in the model, and $q_d$ is the 
$d$-dimensional lattice quantization constant \citep{conwaySloane84}.
The total message length $I(\boldsymbol{\Theta},\dataset)$, therefore, comprises 
of two parts: (1)~the cost of encoding the parameters, $I(\boldsymbol{\Theta})$, and 
(2)~the cost of encoding the data given the parameters, $I(\dataset|\boldsymbol{\Theta})$.
A concise description of the MML method is presented in \citet{oliver1994mml}.

The key differences between ML, MAP, and MML estimation techniques are
as follows:
in ML estimation, the encoding cost of parameters is, in effect, considered constant,
and minimizing the message length corresponds to minimizing the negative
log-likelihood of the data (the second part).
In MAP based estimation, a probability \textit{density} 
rather than the probability is used.
It is self evident that continuous parameter values can
only be stated to some finite precision;
MML incorporates this in the framework
by determining the region of uncertainty in which the parameter is located.
The value of $V = \dfrac{q_d^{-d/2}}{\sqrt{|\mathcal{F}(\boldsymbol{\Theta})|}}$ gives a 
measure of the volume
of the region of uncertainty in which the parameter $\boldsymbol{\Theta}$ is centered.
This multiplied by the probability density $h(\boldsymbol{\Theta})$ gives the 
\emph{probability} of a particular $\boldsymbol{\Theta}$ as $\Pr(\boldsymbol{\Theta}) = h(\boldsymbol{\Theta})V$.
This probability is used to compute the message length associated with
encoding the continuous valued parameters (to a finite precision).

\subsection{MML estimation of the parameters of the BVM distribution}
\label{subsec:bvm_sine_mml}

In this section, we outline the derivation of the 
MML-based parameter estimates of a BVM Sine distribution. 
As explained in Section~\ref{subsec:mml_parameter_estimation}, the
derivation of the MML estimates requires the formulation of the 
message length expression (Equation~\ref{eqn:two_part_msg}) for encoding
some observed data using the BVM Sine distribution.

The formulation requires the use of a suitable prior density on the
parameters. We use the parameterization $\boldtheta$
and the corresponding prior $\Pr(\boldtheta)$ that was formulated
in the MAP analyses in Section~\ref{subsec:bvm_sine_map}.
It is to be noted that the MML estimation is invariant
to the parameterization used \citep{oliver1994mml}. \\

\noindent\textbf{Notations:}
\label{subsec:bvm_sine_notation}
Before describing the MML approach, the following notations
are defined as these are used in the following discussion.
The partial derivatives of the normalization constant $c(\kappa_1,\kappa_2,\lambda)$
of the BVM Sine distribution would be required later on. 
The following are the notations adopted to represent them.
\begin{gather*}
c(\kappa_1,\kappa_2,\lambda) = c, \quad
c_{\kappa_1} = \partial c/\partial\kappa_1, \quad 
c_{\kappa_2} = \partial c/\partial\kappa_2, \quad 
c_{\lambda} = \partial c/\partial\lambda \\
c_{\kappa_1\kappa_1} = \partial^2 c/\partial\kappa_1^2, \quad
c_{\kappa_2\kappa_2} = \partial^2 c/\partial\kappa_2^2, \quad
c_{\lambda\lambda} = \partial^2 c/\partial\lambda^2, \\
c_{\kappa_1\kappa_2} = \partial^2c/\partial\kappa_1\partial\kappa_2, \quad
c_{\kappa_1\lambda} = \partial^2c/\partial\kappa_1\partial\lambda, \quad
c_{\kappa_2\lambda} = \partial^2c/\partial\kappa_2\partial\lambda
\end{gather*}
We also require the determinant of the Fisher 
information for the MML estimation of parameters. 
We use the above notations in the following computation of the Fisher information.
The computation of these partial derivatives is explained
in Section~\ref{subsec:bvm_sine_norm_constant_derivatives}.

\subsubsection{Computation of Expectations}

In order to proceed with the derivation of the Fisher information, we first 
outline the derivation of some of the required  \emph{expectation} quantities.
For random variables $\theta_1,\theta_2$ sampled from the BVM Sine
distribution (Equation~\ref{eqn:bvm_sine}),
we compute the following quantities: 
$\expect[\cos(\theta_1-\mu_1)]$,
$\expect[\cos(\theta_2-\mu_2)]$,
$\expect[\cos(\theta_1-\mu_1)\cos(\theta_2-\mu_2)]$, and
$\expect[\sin(\theta_1-\mu_1)\sin(\theta_2-\mu_2)]$.

\citet{singh2002probabilistic} derived the normalization constant 
as an infinite series expansion given by Equation~\ref{eqn:bvm_sine_norm_constant}.
We use the following \emph{integral} form of the normalization constant
to derive the above mentioned expectations, as a function of $\kappa_1,\kappa_2,$
and $\lambda$.
\begin{gather*}
c(\kappa_1,\kappa_2,\lambda) = \int_{-\pi}^{\pi} \int_{-\pi}^{\pi} 
    \exp\{\kappa_1 \cos(\theta_1 - \mu_1) 
        + \kappa_2 \cos(\theta_2 - \mu_2)
        + \lambda \sin(\theta_1 - \mu_1)\sin(\theta_2 - \mu_2)\} \,\,d\theta_2\,d\theta_1 
\end{gather*}
On differentiating the above integral with respect to $\kappa_1$, we get  
\begin{dmath*}
\frac{\partial}{\partial\kappa_1} c(\kappa_1,\kappa_2,\lambda)
  = \int_{-\pi}^{\pi} \int_{-\pi}^{\pi} \cos(\theta_1 - \mu_1)
    \exp\{\kappa_1 \cos(\theta_1 - \mu_1) 
        + \kappa_2 \cos(\theta_2 - \mu_2)
        + \lambda \sin(\theta_1 - \mu_1)\sin(\theta_2 - \mu_2)\} \,\,d\theta_2\,d\theta_1 
= c(\kappa_1,\kappa_2,\lambda) \,\, \expect[\cos(\theta_1-\mu_1)] 
\end{dmath*}
Hence, the expectation can be represented using the above defined notation as 
\begin{gather}
\expect[\sin(\theta_1-\mu_1)] = 0 = \expect[\sin(\theta_2-\mu_2)] \notag\\
\expect[\cos(\theta_1-\mu_1)] = \frac{1}{c(\kappa_1,\kappa_2,\lambda)} \frac{\partial c(\kappa_1,\kappa_2,\lambda)}{\partial\kappa_1} = \frac{c_{\kappa_1}}{c} \notag\\
\text{Similarly,}\quad
\expect[\cos(\theta_2-\mu_2)] = \frac{c_{\kappa_2}}{c}
\quad\text{and}\quad
\expect[\sin(\theta_1-\mu_1)\sin(\theta_2-\mu_2)] = \frac{c_{\lambda}}{c} 
\label{eqn:expect_sincos1}
\end{gather}
On differentiating twice the integral form of $c(\kappa_1,\kappa_2,\lambda)$
with respect to $\kappa_1,\kappa_2,$  
and $\lambda$, we get the following relationships
\begin{gather}
\expect[\cos(\theta_1-\mu_1)\cos(\theta_2-\mu_2)] = \frac{c_{\kappa_1\kappa_2}}{c}, \notag\\
\expect[\cos(\theta_1-\mu_1)\sin(\theta_2-\mu_2)] = 0
= \expect[\sin(\theta_1-\mu_1)\cos(\theta_2-\mu_2)]
\label{eqn:expect_sincos2}
\end{gather}

\subsubsection{Computation of the Fisher information}

As described in Section~\ref{subsec:mml_parameter_estimation},
the computation of the \emph{determinant} of the Fisher information matrix
requires the evaluation of the second order partial derivatives
of the negative log-likelihood function with respect to the parameters of the distribution.
As per the density function (Equation~\ref{eqn:bvm_sine}),
the negative log-likelihood of a datum $\boldx = (\theta_1,\theta_2)$ is given by
\begin{equation}
\fancyl(\boldx|\boldtheta) = \log c(\kappa_1,\kappa_2,\lambda) 
- \kappa_1  \cos(\theta_{1} - \mu_1)
- \kappa_2  \cos(\theta_{2} - \mu_2) 
- \lambda   \sin(\theta_{1} - \mu_1)\sin(\theta_{2} - \mu_2)
\label{eqn:bvm_sine_negloglike}
\end{equation}
where $\boldtheta=(\mu_1,\mu_2,\kappa_1,\kappa_2,\lambda)$ as indicated before.
Let $\fisherone$ denote the Fisher information for a \emph{single} observation.
the Fisher information matrix $\fisherone$ in the case of an \fb~distribution
is a $5\times5$ \emph{symmetric} matrix.
Further, the determinant $|\fisherone|$ is decomposed as a product of $|\fisherij_A|$
and $|\fisherij_S|$, where $\fisherij_A$ is the Fisher matrix associated
with the angular parameters $\mu_1$ and $\mu_2$, and $\fisherij_S$ is the
Fisher matrix associated with the scale parameters $\kappa_1,\kappa_2$, and $\lambda$. \\

\noindent\emph{Fisher matrix ($\fisherij_{A}$) associated with $\mu_1,\mu_2$:}
$\fisherij_A$ is a $2\times 2$ symmetric matrix whose elements are the expected values 
of the second order partial derivatives of $\mathcal{L}$ with respect to $\mu_1$ and $\mu_2$.
On differentiating Equation~\ref{eqn:bvm_sine_negloglike} with respect to $\mu_1$, we get
\begin{align}
\frac{\partial\fancyl}{\partial\mu_1} &= -\kappa_1 \sin(\theta_{1} - \mu_1)
+ \lambda\cos(\theta_{1} - \mu_1)\sin(\theta_{2} - \mu_2) \label{eqn:bvm_sine_dl_dmu1}\\
\text{and}\quad
\frac{\partial^2\fancyl}{\partial\mu_1^2} &= \kappa_1 \cos(\theta_{1} - \mu_1)
+ \lambda\sin(\theta_{1} - \mu_1)\sin(\theta_{2} - \mu_2) \notag\\
\text{Hence,}\quad
\fancyf_{\mu_1\mu_1} = \expect\left[\frac{\partial^2\fancyl}{\partial\mu_1^2}\right] &=  \kappa_1\, \expect[\cos(\theta_{1} - \mu_1)]
+ \lambda \,\expect[\sin(\theta_{1} - \mu_1)\sin(\theta_{2} - \mu_2)]\notag\\
&= \kappa_1 \frac{c_{\kappa_1}}{c} + \lambda \frac{c_{\lambda}}{c}\notag\\
\text{Similarly,}\quad
\fancyf_{\mu_2\mu_2} = \expect\left[\frac{\partial^2\fancyl}{\partial\mu_2^2}\right] =
&= \kappa_2 \frac{c_{\kappa_2}}{c} + \lambda \frac{c_{\lambda}}{c}  
\label{eqn:bvm_sine_fisher_angles1}
\end{align}
On taking the derivative of Equation~\ref{eqn:bvm_sine_dl_dmu1} with respect to
$\mu_2$, we get
\begin{align}
\frac{\partial^2\fancyl}{\partial\mu_2\partial\mu_1} &= 
-\lambda\cos(\theta_{1} - \mu_1)\cos(\theta_{2} - \mu_2) \notag\\
\text{so that,}\quad
\fancyf_{\mu_2\mu_1} = \expect \left[\frac{\partial^2\fancyl}{\partial\mu_2\partial\mu_1}\right] &= 
-\lambda \expect [\cos(\theta_{1} - \mu_1)\cos(\theta_{2} - \mu_2)]
= -\lambda \frac{c_{\kappa_1\kappa_2}}{c}
\label{eqn:bvm_sine_fisher_angles2}
\end{align}

\noindent\emph{Fisher matrix ($\fisherij_S$) associated with $\kappa_1,\kappa_2,\lambda$:}
$\fisherij_S$ is a $3\times 3$ symmetric matrix whose elements are the expected values 
of the second order partial derivatives of $\mathcal{L}$ with respect to $\kappa_1,\kappa_2,$ and $\lambda$.
On differentiating Equation~\ref{eqn:bvm_sine_negloglike} with respect to $\kappa_1,\kappa_2,$ and $\lambda$, we get
\begin{align}
\frac{\partial\mathcal{L}}{\partial\kappa_1} = \frac{c_{\kappa_1}}{c} - \cos(\theta_{1} - \mu_1) &\quad\text{and}\quad
\frac{\partial\mathcal{L}}{\partial\lambda} = \frac{c_{\lambda}}{c} - \sin(\theta_{1} - \mu_1) \sin(\theta_{2} - \mu_2) \notag\\
\frac{\partial^2\mathcal{L}}{\partial\kappa_1^2} &= \frac{cc_{\kappa_1\kappa_1}-c_{\kappa_1}^2}{c^2} = \fisherij_{\kappa_1\kappa_1}\notag\\
\frac{\partial^2\mathcal{L}}{\partial\kappa_2^2} &= \frac{cc_{\kappa_2\kappa_2}-c_{\kappa_2}^2}{c^2} = \fisherij_{\kappa_2\kappa_2}\notag\\
\frac{\partial^2\mathcal{L}}{\partial\lambda^2} &= \frac{cc_{\lambda\lambda}-c_{\lambda}^2}{c^2} = \fisherij_{\lambda\lambda}\notag\\
\frac{\partial^2\mathcal{L}}{\partial\kappa_1\partial\kappa_2} &= \frac{cc_{\kappa_1\kappa_2}-c_{\kappa_1}c_{\kappa_2}}{c^2} = \fisherij_{\kappa_1\kappa_2}\notag\\
\frac{\partial^2\mathcal{L}}{\partial\lambda\partial\kappa_1} &= \frac{cc_{\lambda\kappa_1}-c_{\lambda}c_{\kappa_1}}{c^2} = \fisherij_{\lambda\kappa_1}\notag\\
\frac{\partial^2\mathcal{L}}{\partial\lambda\partial\kappa_2} &= \frac{cc_{\lambda\kappa_2}-c_{\lambda}c_{\kappa_2}}{c^2} = \fisherij_{\lambda\kappa_2}
\label{eqn:bvm_sine_fisher_scale}
\end{align}

\noindent\emph{Fisher matrix $\fisher$ associated with the 5-parameter vector $\boldtheta$:}
On differentiating Equation~\ref{eqn:bvm_sine_dl_dmu1}
with respect to $\kappa_1$ and computing the expectation of the differential, we get 
\begin{gather*}
\frac{\partial^2\fancyl}{\partial\kappa_1\partial\mu_1} = -\sin(\theta_1-\mu_1)
\quad\text{and}\quad
\frac{\partial^2\fancyl}{\partial\lambda\partial\mu_1} = \cos(\theta_1-\mu_1)\sin(\theta_2-\mu_2) \notag\\
\quad\text{Hence,}\quad
\expect\left[ \frac{\partial^2\fancyl}{\partial\kappa_1\partial\mu_1} \right] = 0 = \fisherij_{\kappa_1\mu_1} 
\quad\text{and}\quad
\expect\left[ \frac{\partial^2\fancyl}{\partial\lambda\partial\mu_1} \right] = 0 = \fisherij_{\lambda\mu_1} 
\end{gather*}
This allows for the computation of $|\fisherone|$ as the product of
$|\fisherij_A|$ and $|\fisherij_S|$, that is,
\begin{equation*}
|\fisherone| =  \begin{vmatrix}
                  \fisherij_{\mu_1\mu_1} &  \fisherij_{\mu_1\mu_2}  & 0 & 0 & 0 \\
                  \fisherij_{\mu_2\mu_1} &  \fisherij_{\mu_2\mu_2}  & 0 & 0 & 0 \\  
                  0 & 0 & \fisherij_{\kappa_1\kappa_1}  & \fisherij_{\kappa_1\kappa_2} & \fisherij_{\kappa_1\lambda} \\
                  0 & 0 & \fisherij_{\kappa_2\kappa_1}  & \fisherij_{\kappa_2\kappa_2} & \fisherij_{\kappa_2\lambda} \\
                  0 & 0 & \fisherij_{\lambda\kappa_1}  & \fisherij_{\lambda\kappa_2} & \fisherij_{\lambda\lambda} 
                \end{vmatrix}
          = |\fisherij_A| |\fisherij_S|
\end{equation*}
Then, the Fisher information for some observed data $\dataset=\{\boldx_1,\ldots,\boldx_N\}$ 
is given by 
\begin{equation}
|\fisher| = N^5 |\fisherone|
\label{eqn:bvm_sine_fisher}
\end{equation}
as each element in $|\fisherone|$ is multiplied by the sample size $N$.

\subsubsection{Message length formulation}

The message length to encode some observed data $\dataset$
can now be formulated by substituting the prior density $\Pr(\boldtheta)$
(Equation~\ref{eqn:bvm_sine_prior1}), the Fisher information $|\fisher|$ 
and the negative log-likelihood of the data (Equation~\ref{eqn:bvm_sine_negloglike_data})
in the message length expression (Equation~\ref{eqn:two_part_msg}).
The MML parameter estimates are the ones that minimize the total 
message length. As there is no analytical form of the MML estimates,
the solution is obtained, as for the maximum likelihood and MAP cases,
by using the NLopt 
optimization library \citep{nlopt}.
At each stage of the optimization routine, the Fisher information
needs to be calculated. However, this involves the computation of 
complex entities such as the normalization constant $c(\kappa,\beta)$ and its partial derivatives.
The computation of these intricate mathematical forms using numerical methods
is discussed next in Section~\ref{subsec:bvm_sine_norm_constant_derivatives}.

\subsection{Computation of the normalization constant and its derivatives}
\label{subsec:bvm_sine_norm_constant_derivatives}

The computation of the negative log-likelihood and the message length
requires the normalization constant and its associated derivatives. 
In this section, the description of the methods that can be employed to 
efficiently compute these complex functions is explored.
We will utilize the properties of Bessel functions 
to implement the normalization constant and the necessary partial 
derivatives as 
limiting order summations for the BVM Sine distribution.

\subsubsection{Computing $\boldsymbol{\log c(\kappa_1,\kappa_2,\lambda)}$ and the logarithm of 
the partial derivatives: $\boldsymbol{c_{\kappa_1}}, \boldsymbol{c_{\kappa_2}}, 
\boldsymbol{c_{\kappa_1\kappa_1}}, \boldsymbol{c_{\kappa_2\kappa_2}}$
and $\boldsymbol{c_{\kappa_1\kappa_2}}$}
\label{subsec:bvm_sine_kappa_der}

The expressions of $c,c_{\kappa_1},c_{\kappa_2},c_{\kappa_1\kappa_1},
c_{\kappa_2\kappa_2}$, and $c_{\kappa_1\kappa_2}$ are related
to each other. 
These expressions are explained by defining the quantity
$S^{(m,n)}_1$, a logarithm sum, 
\begin{equation}
S^{(m,n)}_1 = \log\delta_1 
+ \log\sum_{j=0}^\infty\underbrace{\binom{2j}{j} e^{j} I_{j+m}(\kappa_1) I_{j+n}(\kappa_2)}_{f_j} 
\label{eqn:bvm_sine_series1}
\end{equation}
where $m,n\in\{0,1,2\}$,
$\delta_1=4\pi^2$, and $e = \dfrac{\lambda^2}{4\kappa_1\kappa_2} < 1$ (by definition).

\noindent\emph{Computation of the series $S^{(m,n)}_1$:}
We first establish that $f_{j+1}<f_j \,\forall j \ge 0$ and 
show that $S^{(m,n)}_1$ converges to a finite sum as $j \to \infty$.
Consider the logarithm of the ratio of consecutive terms $f_j$ and $f_{j+1}$
in $S^{(m,n)}_1$
\begin{align}
\log\frac{f_{j+1}}{f_j} = \log\frac{\binom{2j+2}{j+1}}{\binom{2j}{j}} + \log e
+ \log\frac{I_{j+m+1}(\kappa_1)}{I_{j+m}(\kappa_1)}
+ \log\frac{I_{j+n+1}(\kappa_2)}{I_{j+n}(\kappa_2)}
\label{eqn:bvm_sine_ratio_series1}
\end{align}
for $p,v>0$, $I_{p+v} < I_{p}$, and the ratio $\frac{I_{p+v}}{I_p} \to 0$
as $p \to \infty$ \citep{amos1974computation}.
Further, $e < 1$ implies the above equation is the sum of negative terms. Hence,
$\log\frac{f_{j+1}}{f_j} < 0$, which means $f_{j+1}<f_j$. Also, 
\begin{equation*}
\lim_{j\to\infty} \log\frac{f_{j+1}}{f_j} = \log 4 + \log e 
+ \lim_{j\to\infty} \log\frac{I_{j+m+1}(\kappa_1)}{I_{j+m}(\kappa_1)}
+ \lim_{j\to\infty} \log\frac{I_{j+n+1}(\kappa_2)}{I_{j+n}(\kappa_2)}
= - \infty
\end{equation*}
Hence, as $\displaystyle\lim_{j\to\infty} \dfrac{f_{j+1}}{f_j} = 0$, $S^{(m,n)}_1$ is a convergent series.

For a practical implementation of the sum, we need to
express $S^{(m,n)}_1$ as the modified summation
\begin{equation}
S^{(m,n)}_1 = \log\delta_1 + \log f_0 + \log \sum_{j=0}^\infty t_j
\label{eqn:bvm_sine_series1_modified}
\end{equation}
where each $f_j$ is divided by the 
\emph{maximum} term $f_0$. For each $j > 0, \log f_{j}$ is calculated using
the previous term $\log f_{j-1}$ (Equation~\ref{eqn:bvm_sine_ratio_series1}).
The new term $t_j = f_j/f_0$ is then 
computed\footnote{Because of the nature of Bessel functions,
$\log f_j$ can get very large and can result in overflow when
calculating the exponent $\exp(\log f_j)$. However, dividing by $f_0$
results in $f_j/f_0 < 1$.
}
as $\exp(\log f_j - \log f_0)$. This is because computing the difference with the maximum value 
and then taking the exponent ensures numerical stability.
The summation is terminated when the ratio $\dfrac{t_j}{\sum_{k=0}^j t_k} < \epsilon$
(a small threshold $\sim10^{-6}$).

\begin{itemize}
\item Let $S(c)=\log c(\kappa_1,\kappa_2,\lambda)$:
Substituting $m=0$ and $n=0$ in Equation~\ref{eqn:bvm_sine_series1}
gives the logarithm of the normalization constant 
(given in Equation~\ref{eqn:bvm_sine_norm_constant}).
Hence, $S(c)=S^{(0,0)}_1$. 

\item Let the $j^{th}$ term dependent on $\kappa_1$ in Equation~\ref{eqn:bvm_sine_norm_constant}
be represented as
$g_j(\kappa_1) = I_j/\kappa_1^j$, where 
$I_j$ implicitly refers to $I_j(\kappa_1)$. 
Based on the relationship between the Bessel functions $I_{j},I_{j+1}$, and the derivative $I'_j$
in Equation~\ref{eqn:bvm_sine_bessel_identity} \citep{abramowitz1972handbook},
the expressions for the first and second derivatives of $g_j(\kappa_1)$
(Equation~\ref{eqn:gk1_derivatives}) are derived as
\begin{gather}
\kappa_1 I'_j = j I_j + \kappa_1 I_{j+1} \label{eqn:bvm_sine_bessel_identity}\\
g_j'(\kappa_1) = \frac{I_{j+1}}{\kappa_1^j} 
\quad\text{and}\quad g_j''(\kappa_1) = \frac{I_{j+2}}{\kappa_1^j} + \frac{1}{\kappa_1} . \frac{I_{j+1}}{\kappa_1^{j}} \label{eqn:gk1_derivatives}
\end{gather}

\item Let $S(c_{\kappa_1}) = \log c_{\kappa_1}$: 
Because of the similar forms of $g_j(\kappa_1)$ and $g_j'(\kappa_1)$,
the expression for $S(c_{\kappa_1})$ will be 
similar to that of $S(c)$ with a change in \emph{order} of the Bessel functions 
from $m=0$ in Equation~\ref{eqn:bvm_sine_series1} to $m=1$. 
Hence, $S(c_{\kappa_1})=S^{(1,0)}_1$ and
an expression similar to Equation~\ref{eqn:bvm_sine_series1_modified} 
can be derived for $S(c_{\kappa_1})$.

\item Let $S(c_{\kappa_2}) = \log c_{\kappa_2}$: 
Similar to the computation
of $S(c_{\kappa_1})$ above, if we substitute $m=0,n=1$ in 
Equation~\ref{eqn:bvm_sine_series1_modified}, we obtain the expression for 
$S(c_{\kappa_2})=S^{(0,1)}_1$.

\item Let $S(c_{\kappa_1\kappa_2}) = \log c_{\kappa_1\kappa_2}$:
Similar to the above computations
of $S(c_{\kappa_1})$ and $S(c_{\kappa_2})$, if we substitute $m=1,n=1$ in 
Equation~\ref{eqn:bvm_sine_series1_modified}, we obtain the expression for 
$S(c_{\kappa_1\kappa_2})=S^{(1,1)}_1$.

\item Let $S(c_{\kappa_1\kappa_1}) = \log c_{\kappa_1\kappa_1}$:
Substituting $m=2,n=0$ in Equation~\ref{eqn:bvm_sine_series1} gives the 
logarithm sum $S^{(2,0)}_1$ corresponding to the series with terms $\dfrac{I_{j+2}}{\kappa_1^j}$.
Based on the nature of $g_j''(\kappa_1)$ (Equation~\ref{eqn:gk1_derivatives}), 
and noting that $S(c_{\kappa_1}) > S^{(2,0)}_1$ (as $I_{j+1} > I_{j+2}\,\forall\,j\ge0$),
$S(c_{\kappa_1\kappa_1})$ is formulated as
\begin{equation*}
S(c_{\kappa_1\kappa_1}) = S(c_{\kappa_1}) + \log \left(\exp(S^{(2,0)}_1-S(c_{\kappa_1})) + \frac{1}{\kappa_1}\right)
\end{equation*}

\item Let $S(c_{\kappa_2\kappa_2}) = \log c_{\kappa_2\kappa_2}$: Based on the same 
reasoning as above, we have
\begin{equation*}
S(c_{\kappa_2\kappa_2}) = S(c_{\kappa_2}) + \log \left(\exp(S^{(0,2)}_1-S(c_{\kappa_2})) + \frac{1}{\kappa_2}\right)
\end{equation*}

\end{itemize}

\subsubsection{The logarithm of the partial
derivatives: $\boldsymbol{c_{\lambda}}$,
$\boldsymbol{c_{\kappa_1\lambda}}$, 
$\boldsymbol{c_{\kappa_2\lambda}}$, and
$\boldsymbol{c_{\lambda\lambda}}$}

The expressions of $c_{\lambda}, c_{\kappa_1\lambda},$ and $c_{\kappa_2\lambda}$ 
are related and are explained using 
the logarithm sum $S^{(m,n)}_2$ 
\begin{equation}
S^{(m,n)}_2 = \log\delta_2 + \log\sum_{j=1}^\infty\underbrace{\binom{2j}{j} j e^j I_{j+m}(\kappa_1) I_{j+n}(\kappa_2)}_{f_j}
\label{eqn:bvm_sine_series2}
\end{equation}
where $m,n\in\{0,1\}$, 
$\delta_2=\dfrac{8\pi^2}{\lambda}$, and 
$e = \dfrac{\lambda^2}{4\kappa_1\kappa_2}$. 
Note that $S^{(m,n)}_2$ is a convergent series (the proof is based on the same reasoning
as for $S^{(m,n)}_1$).

Let the $j^{th}$ term dependent on $\lambda,\kappa_1$ in 
Equation~\ref{eqn:bvm_sine_norm_constant} be represented as
$g_j(\lambda,\kappa_1) = \lambda^{2j}\dfrac{I_{j}}{\kappa_1^j}$. 
Its partial derivatives are given below.
These derivatives are the terms in the series $S^{(m,n)}_2$ (after
factoring out the common elements as $\delta_2$).
\begin{equation*}
\frac{\partial g_j}{\partial\lambda} = 2j \lambda^{2j-1}\frac{I_{j}}{\kappa_1^j} 
\quad\text{and}\quad
\frac{\partial^2 g_j}{\partial\kappa_1\partial\lambda} = 2j \lambda^{2j-1}\frac{I_{j+1}}{\kappa_1^j}
\end{equation*} 

\begin{itemize}
\item Let $S(c_{\lambda})=\log c_\lambda$: this is obtained by
substituting $m=0$ and $n=0$ in Equation~\ref{eqn:bvm_sine_series2}.
Hence, $S(c_{\lambda})=S^{(0,0)}_2$. 

\item Similarly, $S(c_{\kappa_1\lambda})=\log c_{\kappa_1\lambda}=S^{(1,0)}_2$
and $S(c_{\kappa_2\lambda})=\log c_{\kappa_2\lambda}=S^{(0,1)}_2$.

\item The expression to compute $S(c_{\lambda\lambda})=\log c_{\lambda\lambda}$ is given by
\begin{equation*}
S(c_{\lambda\lambda}) = \log\left(\frac{\delta_2}{\lambda}\right) 
+ \log\sum_{j=1}^\infty\underbrace{\binom{2j}{j} j(2j-1) e^j I_j(\kappa_1) I_j(\kappa_2)}_{f_j}
\label{eqn:bvm_sine_series3}
\end{equation*}
\end{itemize}
The practical implementation of $S_2^{(m,n)}$ and $S(c_{\lambda\lambda})$
is similar to that of $S^{(m,n)}_1$ 
given by Equation~\ref{eqn:bvm_sine_series1_modified}.
However, in these cases, the expressions of $f_j$ and consequently $t_j$, are modified
depending on their specific forms.
Also, the series begin from $j=1$ and, hence,
the respective maximum terms will correspond to $f_1$.

\subsection{Evaluation of the MML estimates}
\label{subsec:bvm_sine_experiments}

For a given BVM Sine distribution characterized by concentration parameters 
$\kappa_1,\kappa_2$ and correlation coefficient $\rho$,
a random sample of size $N$ is generated using the method proposed by \citet{mardia2007protein}.
The angular parameters of the true distribution are set to $\{\mu_1,\mu_2\} = \pi/2$.
The scale parameters $\kappa_1,\kappa_2,$ and $\rho$ are varied
to obtain different BVM Sine distributions and corresponding random samples.
The parameters are estimated using the sampled data and the different
estimation methods. The procedure is repeated
1000 times for each combination of $N,\kappa_1,\kappa_2,$ and $\rho$.

\subsubsection{Methods of comparison}

For every randomly generated sample from a BVM Sine distribution, 
we compute the the ML, MAP, and MML estimators of the parameters,
and these are compared with each other across all the simulations.
The results include the three versions of MAP
estimates resulting from the three forms of the posterior distributions 
(Equations~\ref{eqn:bvm_sine_prior1}-\ref{eqn:bvm_sine_prior3}): \emph{MAP1} corresponds to the posterior
with parameterization $\boldtheta=(\mu_1,\mu_2,\kappa_1,\kappa_2,\lambda)$, 
\emph{MAP2} corresponds to the posterior
with parameterization $\boldtheta'=(\mu_1,\mu_2,\kappa_1,\kappa_2,\rho)$, and 
\emph{MAP3} corresponds to the posterior with parameterization $\boldtheta''=(z_1,z_2,z_3,z_4,z_5)$.
As noted in Section~\ref{subsec:bvm_sine_map}, the MAP3 estimator
will be the same as the ML estimator due to the \citet{rosenblatt1952remarks}
transformation of $\boldtheta$ to $\boldtheta''$.

In order to compare the various estimators, we use 
the mean squared error (MSE) and Kullback-Leibler (KL) distance 
as the objective evaluation metrics.
The estimates are also compared using statistical hypothesis testing. 
For a parameter vector $\boldtheta$ characterizing a true BVM Sine distribution, 
and its estimate $\widehat{\boldtheta}$,
we analyze the MSE and KL distance of $\widehat{\boldtheta}$ with respect to
the true parameter vector $\boldtheta$.
The analytical form of the KL distance between two
BVM distributions is derived in Appendix~\ref{app:bvm_sine_kldiv}. 
We analyze the percentage of times (\emph{wins})
the KL distance of a particular estimator is smaller than that of others. 
When the KL distance of different estimates is compared, because of three different
versions of MAP estimation, three separate frequency plots are presented.
corresponding to the MAP1, MAP2, and MAP3 estimators. 

With respect to statistical hypothesis testing, 
the likelihood ratio test statistic is asymptotically approximated as 
an $\chi^2$ distribution with five degrees of freedom
For the various parameter estimates compared here, it is expected that
at especially large sample sizes, the estimates are close to the ML estimate.
In other words, the empirically determined test statistic is expected to be 
lower than the critical value $\tau=13.086$,
corresponding to a p-value greater than 0.01.

\subsubsection{Empirical analyses}

As per the experimental setup, we present the results for when the original
distribution from which the data is sampled has $\kappa_1=1$ and $\kappa_2=10$.
The correlation coefficient $\rho$ is varied between 0 and 1, so that we 
obtain different values for the correlation parameter $\lambda$ (Equation~\ref{eqn:bvm_sine_rho}).
We discuss the results for varying values of sample sizes $N$,
and $\rho=0.1,0.5,0.9$, corresponding to a low, moderate, and high correlation, respectively. \\

\begin{figure}[!htb]
\centering
\subfloat[Bias-squared]{\includegraphics[width=0.5\textwidth]{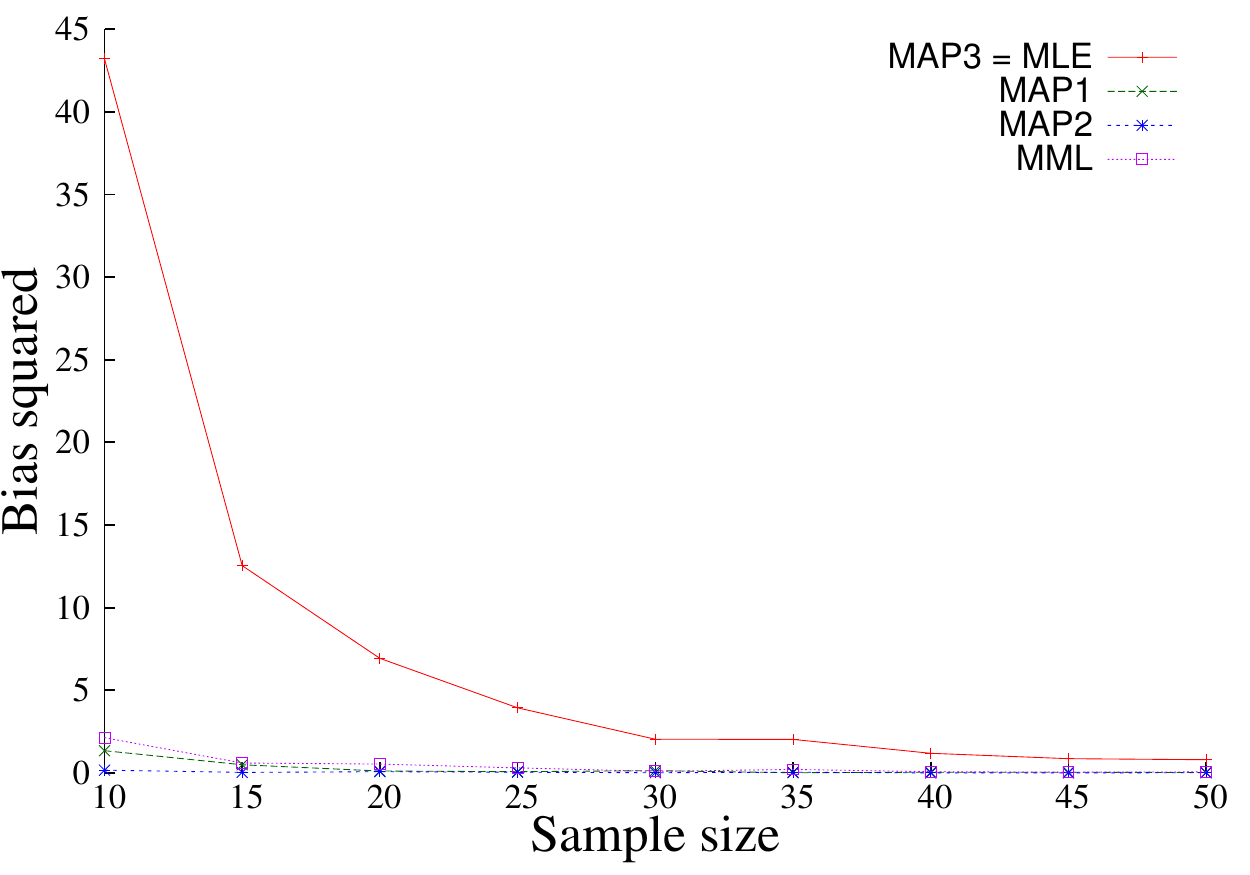}}
\subfloat[Mean squared error]{\includegraphics[width=0.5\textwidth]{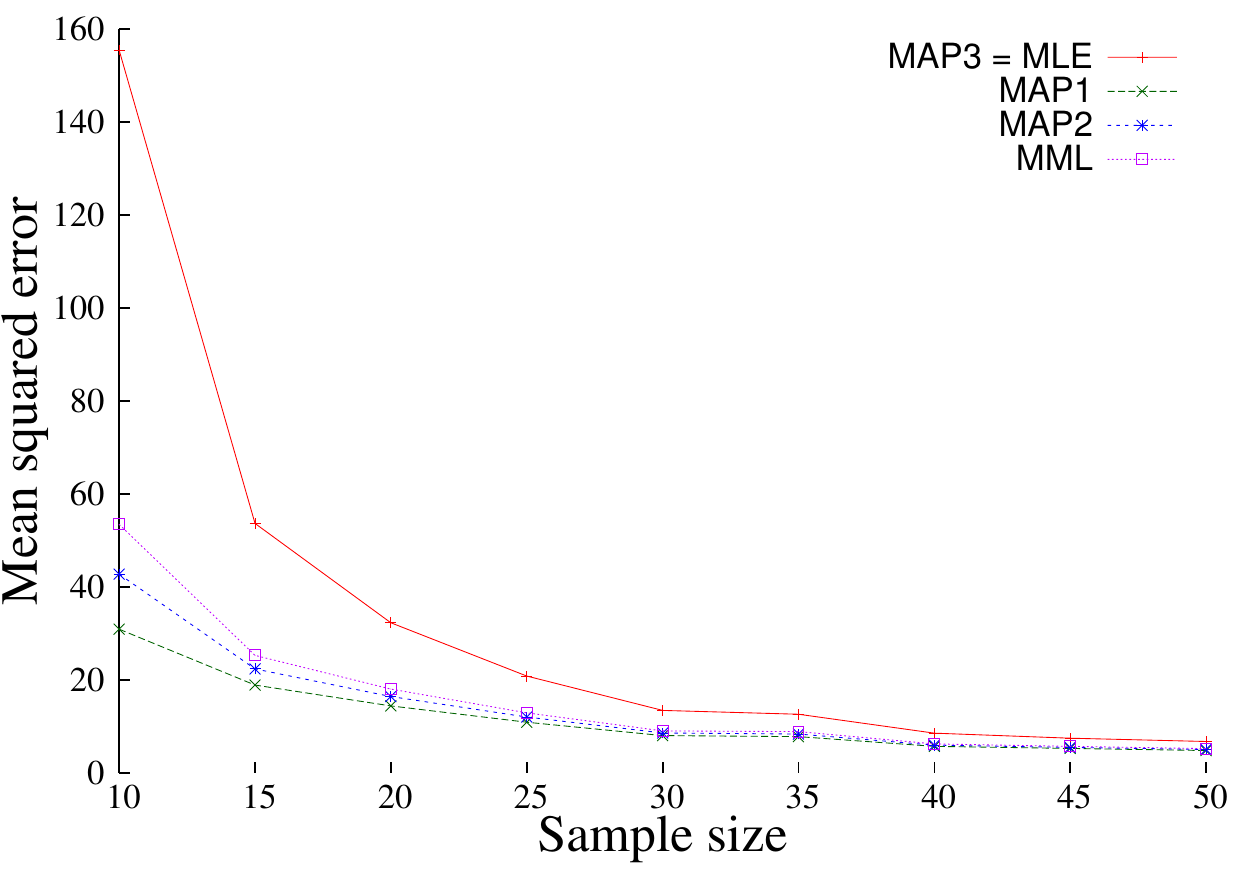}}\\
\subfloat[KL distance (MAP version 1)]{\includegraphics[width=0.33\textwidth]{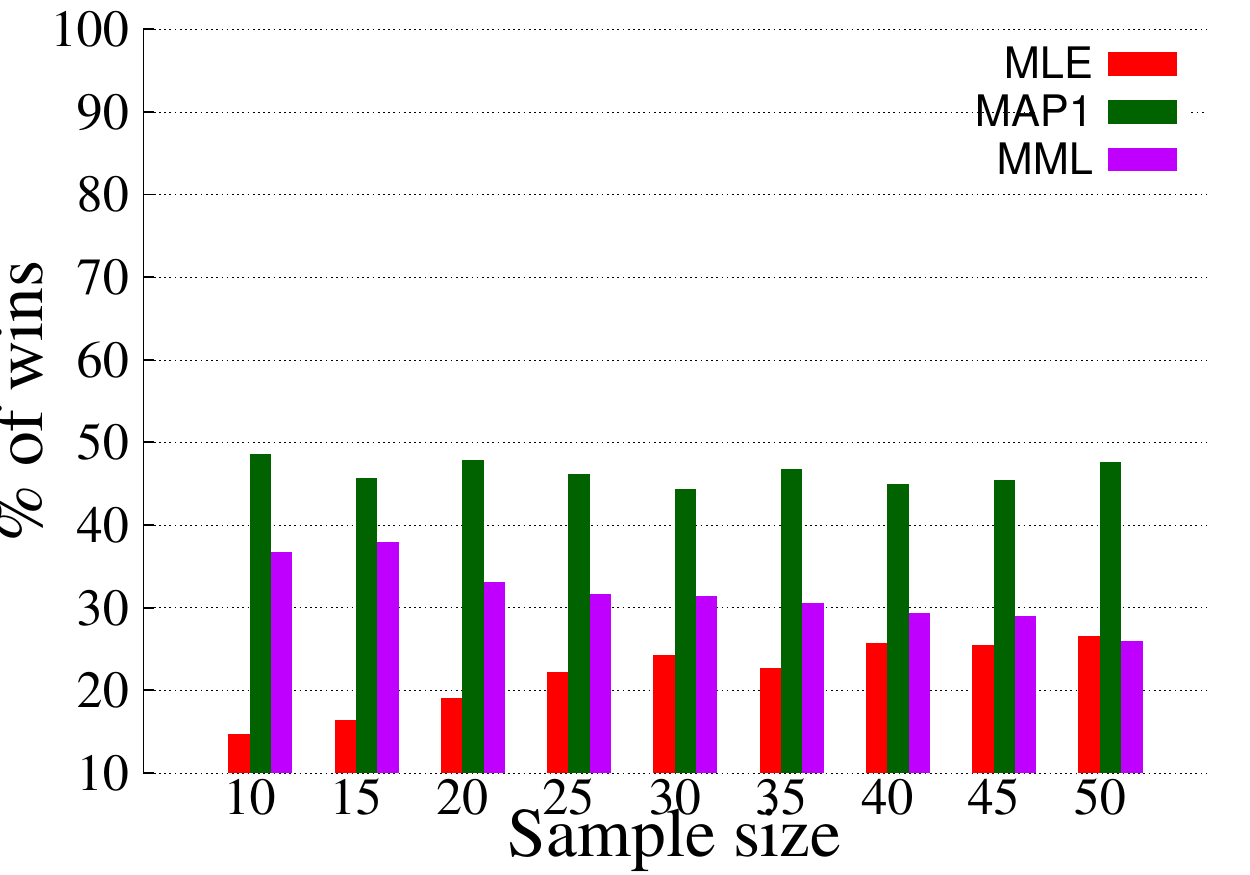}}
\subfloat[KL distance (MAP version 2)]{\includegraphics[width=0.33\textwidth]{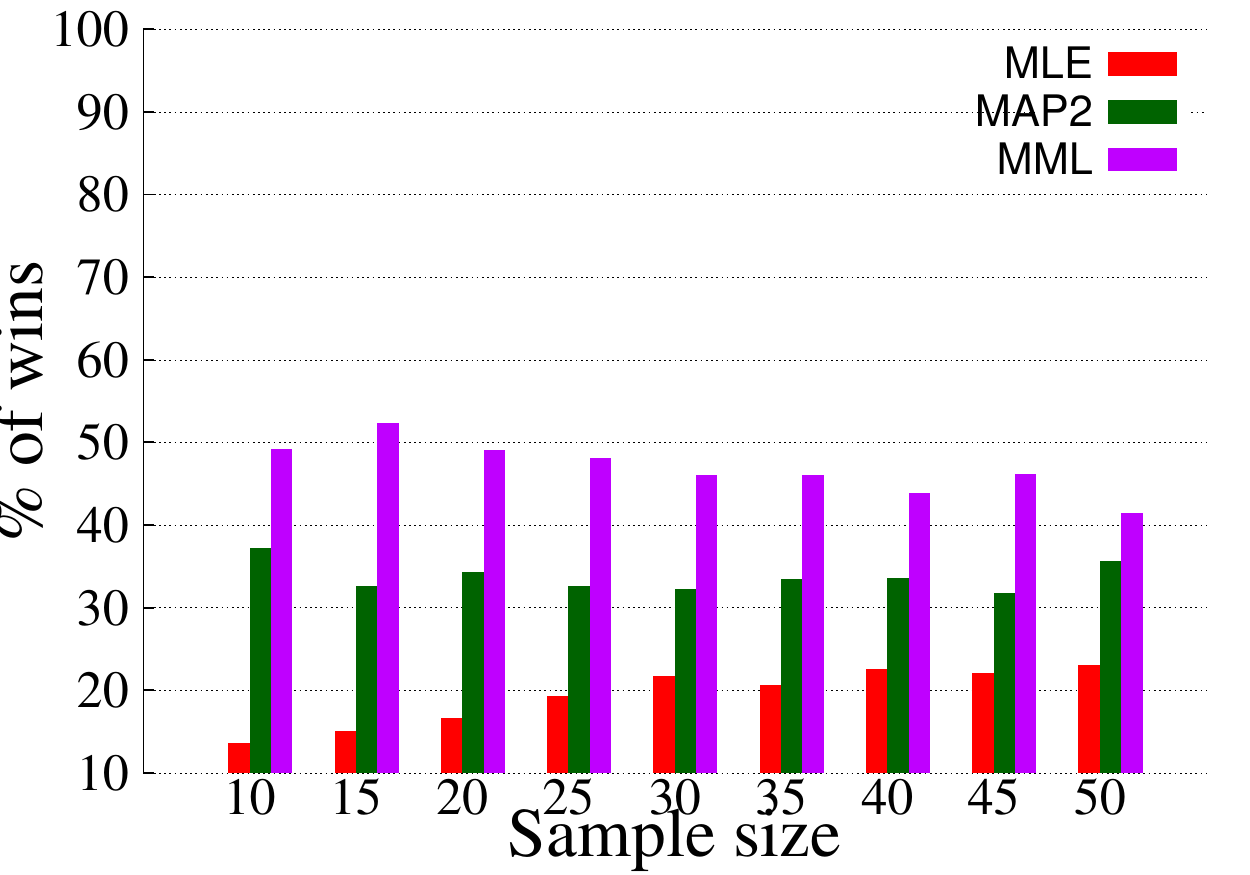}}
\subfloat[KL distance (MAP version 3)]{\includegraphics[width=0.33\textwidth]{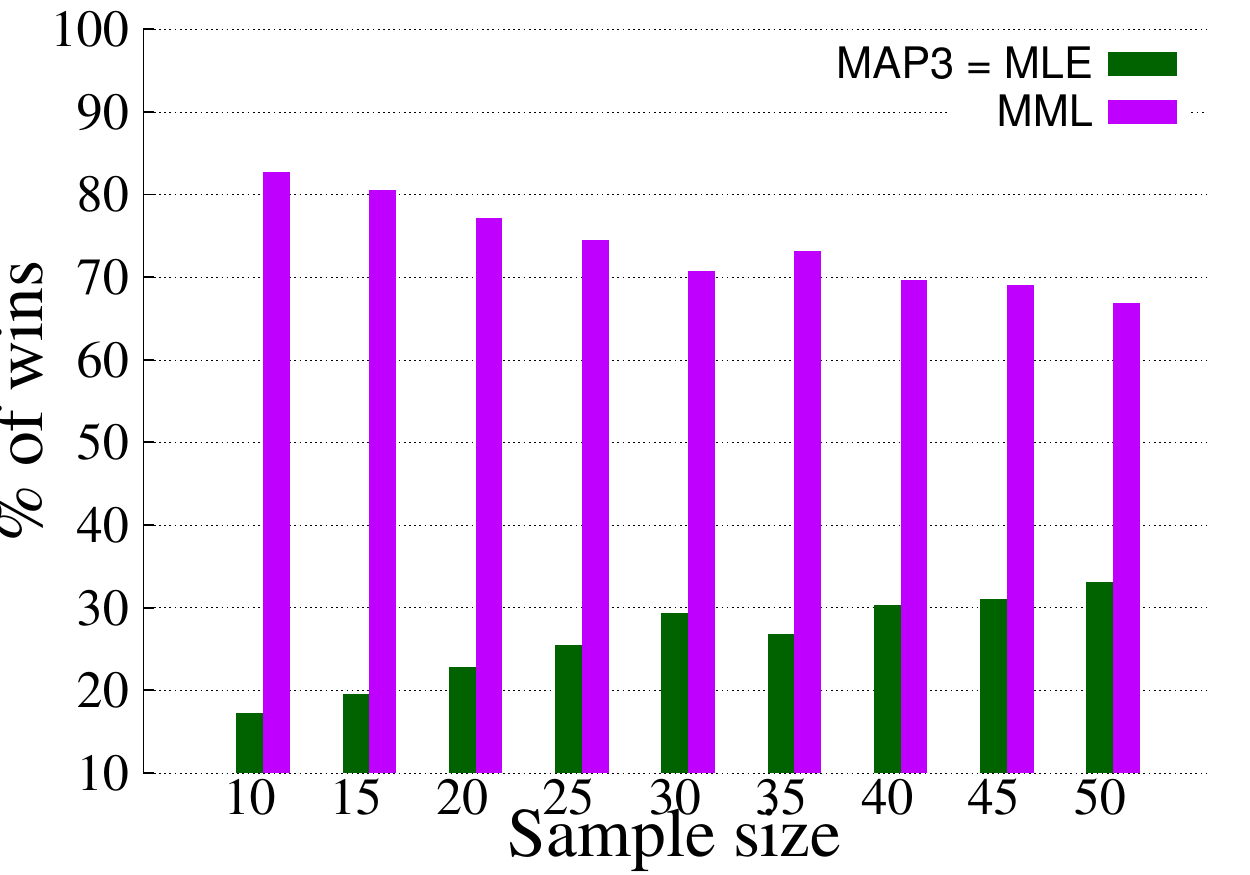}}\\
\subfloat[Variation of test statistics]{\includegraphics[width=0.5\textwidth]{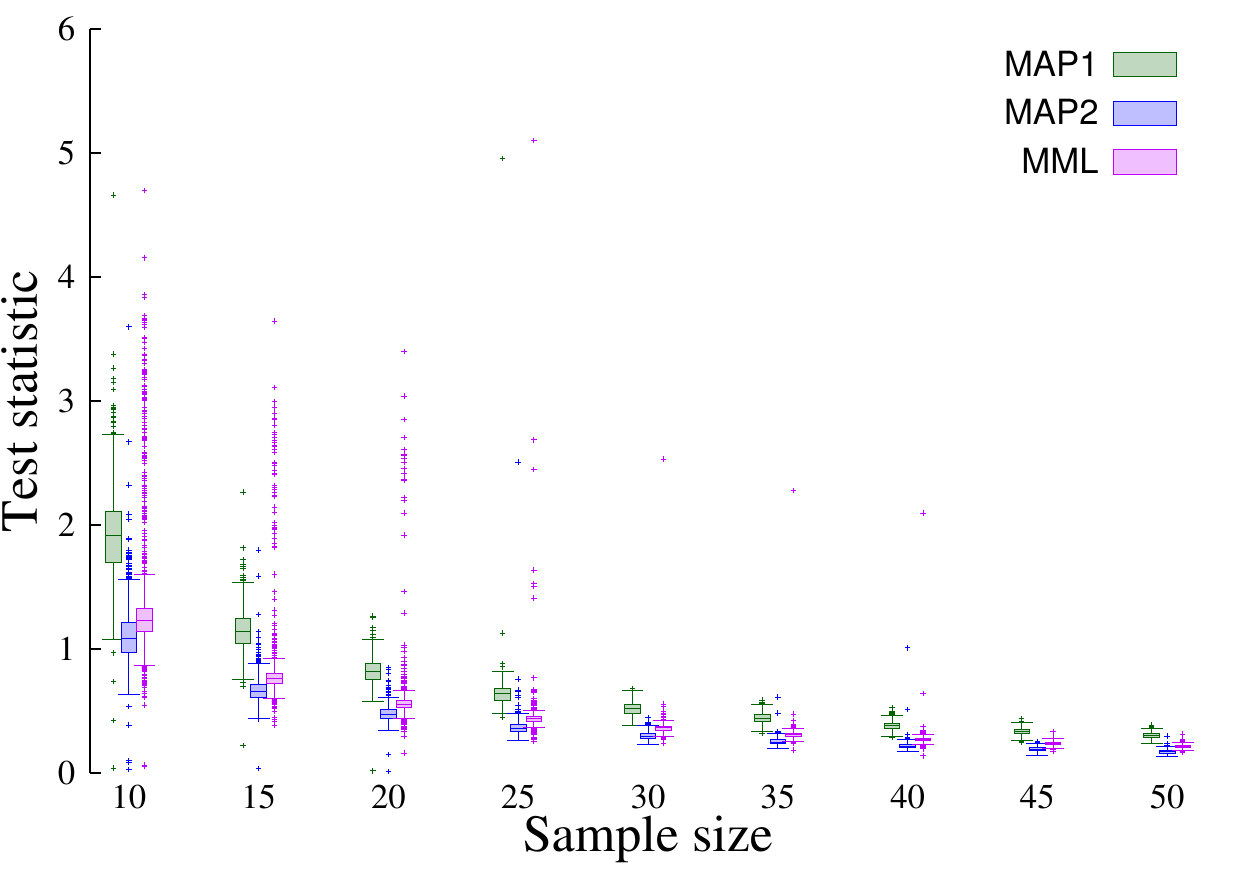}}
\subfloat[Variation of p-values]{\includegraphics[width=0.5\textwidth]{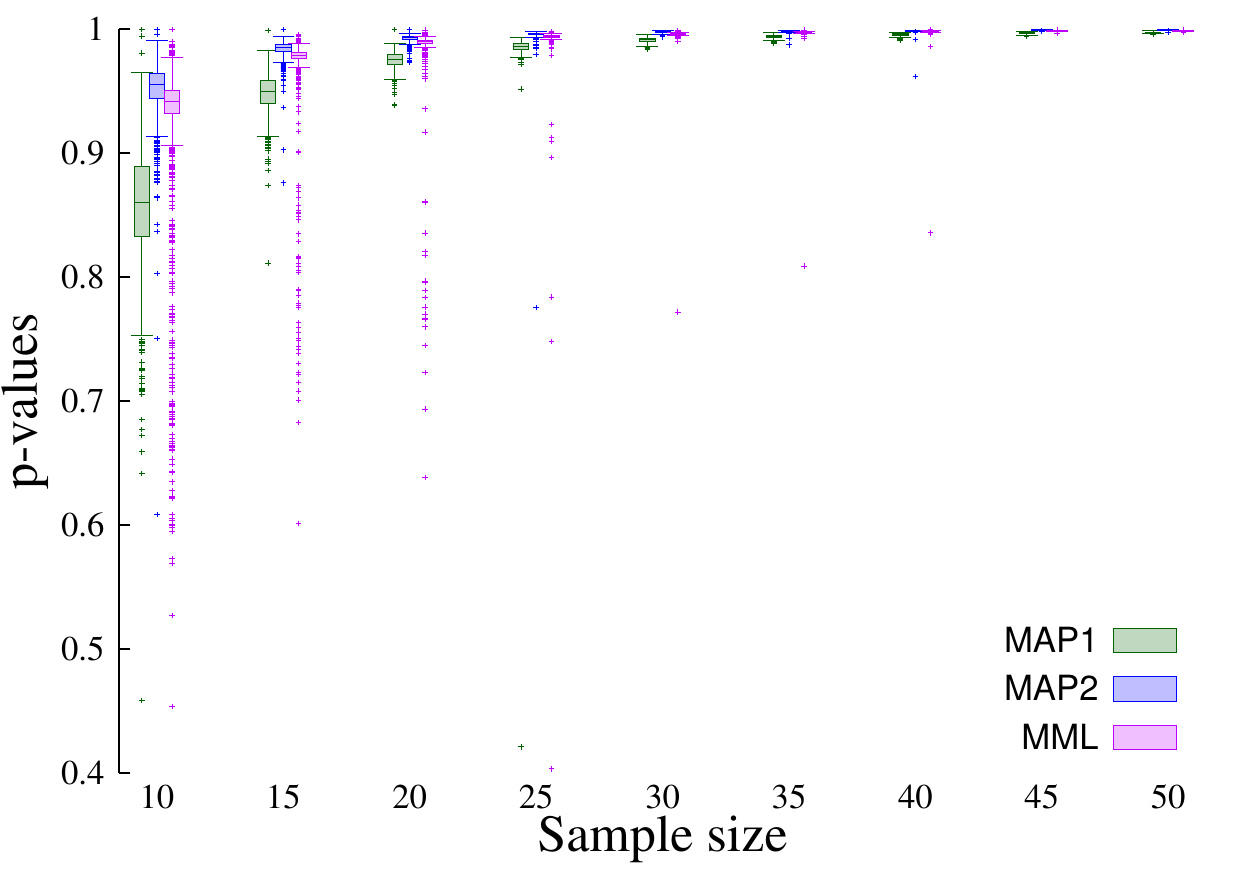}}
\caption
[Comparison of the parameter estimates when $\kappa_1=1,\kappa_2=10,\rho=0.1$.]
{Comparison of the parameter estimates when $\kappa_1=1,\kappa_2=10,\rho=0.1$.}
\label{fig:k1_1_k2_10_r_1}
\end{figure}

\noindent\textbf{For} $\pmb{\rho=0.1:}$
The results are presented in Figure~\ref{fig:k1_1_k2_10_r_1}.
Compared to the ML estimators, the MAP and MML estimators result
in lower bias and MSE for all values of $N$. Both the bias and MSE
continue to decrease as the sample size increases, as 
the estimation improves with more evidence for all methods. When compared with MAP1 and MAP2,
the MML estimators have greater bias and greater MSE. As with the \fb~distribution,
we observe that that MAP1 and MAP2 result in different estimators, and
therefore, result in different bias and MSE values.

The KL distance with respect to MAP1 is in favour of the MAP1 estimators.
The MAP1 estimates result in lower KL distance
as compared to the other estimators 
almost 50\% of the 1000 simulations for each $N$ (Figure~\ref{fig:k1_1_k2_10_r_1}c).
However, the MML estimators win when the MAP2 and MAP3 versions are used.
When MAP3 is used, the MML estimators have a smaller KL distance in close to 
70\% of the simulations (Figure~\ref{fig:k1_1_k2_10_r_1}e).
Further analysis using statistical hypothesis testing illustrates
that the null hypotheses corresponding to the
MAP and MML estimators are accepted (p-values greater than 0.01 in Figure~\ref{fig:k1_1_k2_10_r_1}f).
at the 1\% significance level. \\

\begin{figure}[!htb]
\centering
\subfloat[Bias-squared]{\includegraphics[width=0.5\textwidth]{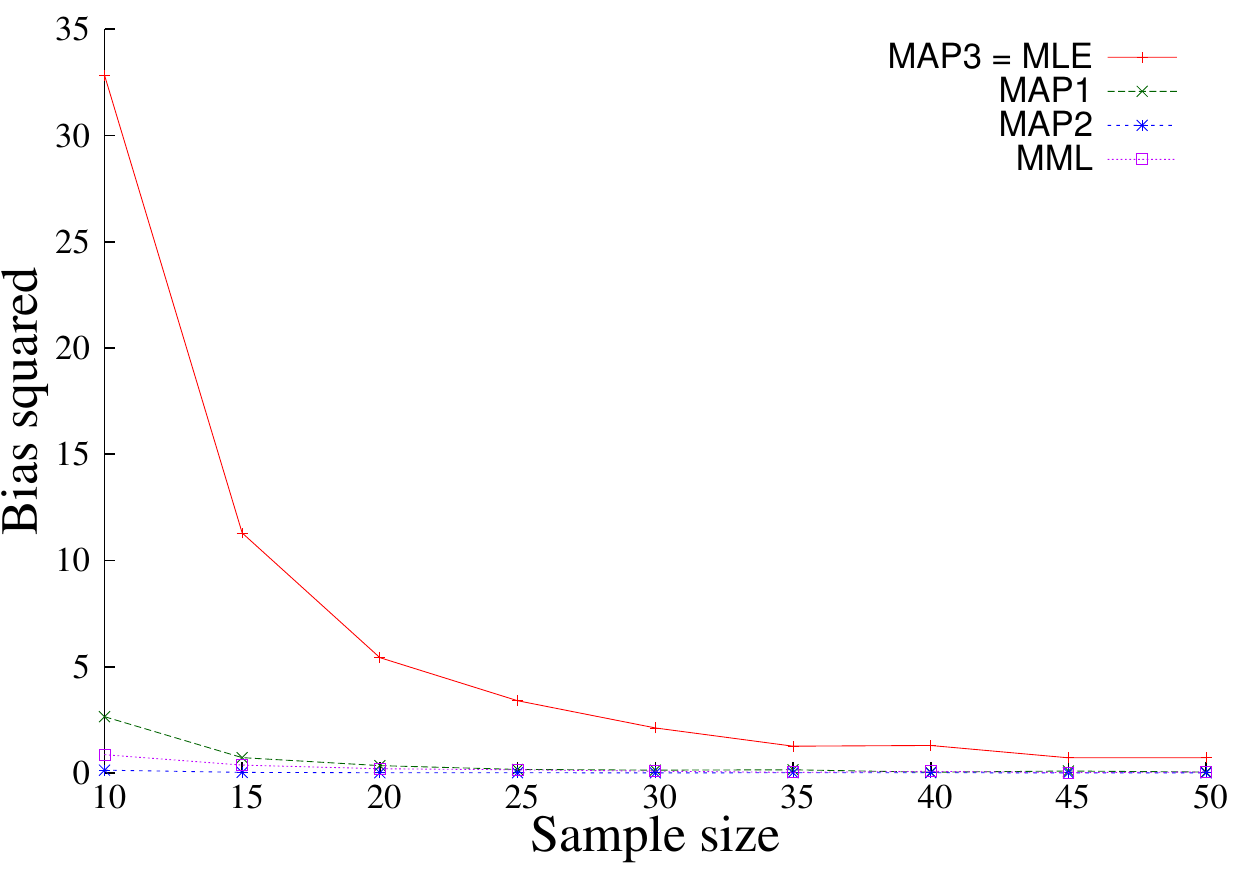}}
\subfloat[Mean squared error]{\includegraphics[width=0.5\textwidth]{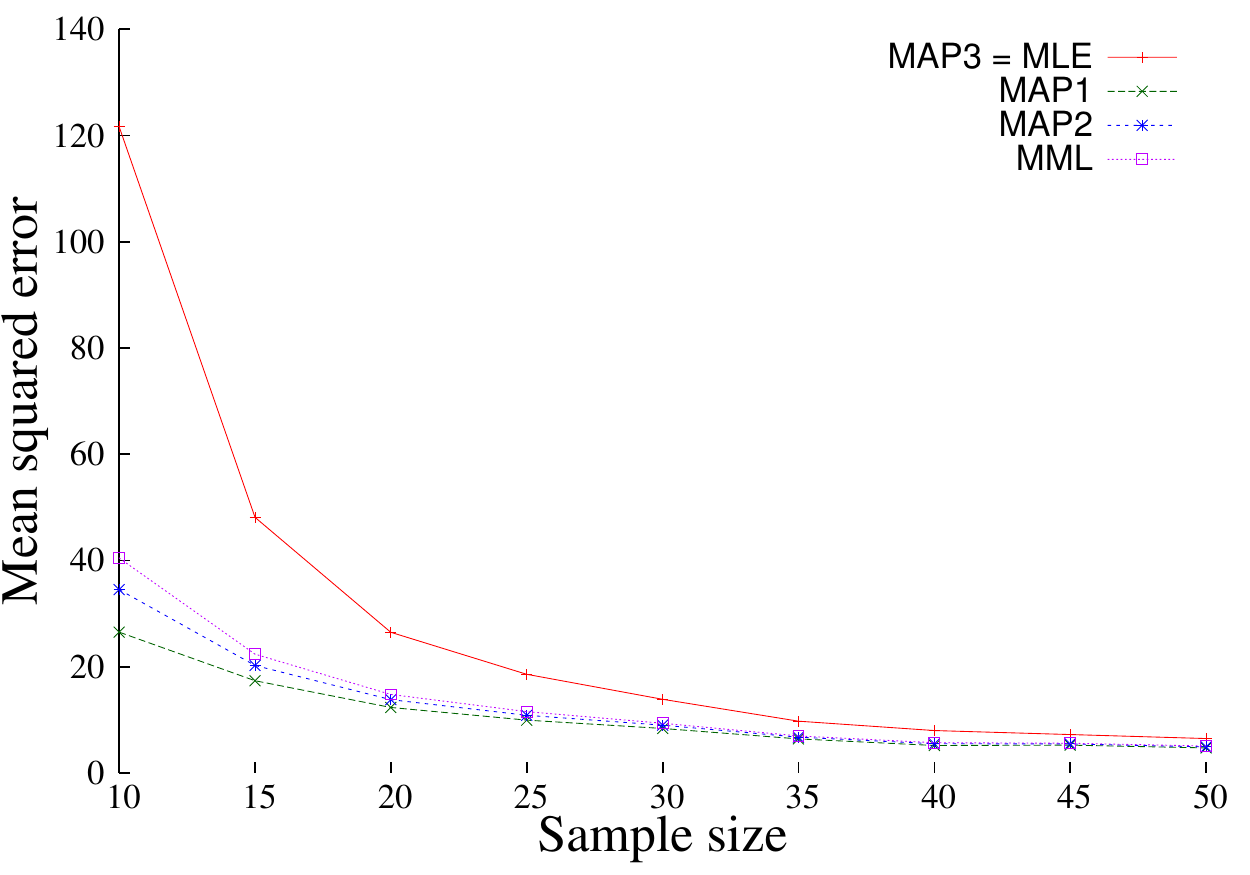}}\\
\subfloat[KL distance (MAP version 1)]{\includegraphics[width=0.33\textwidth]{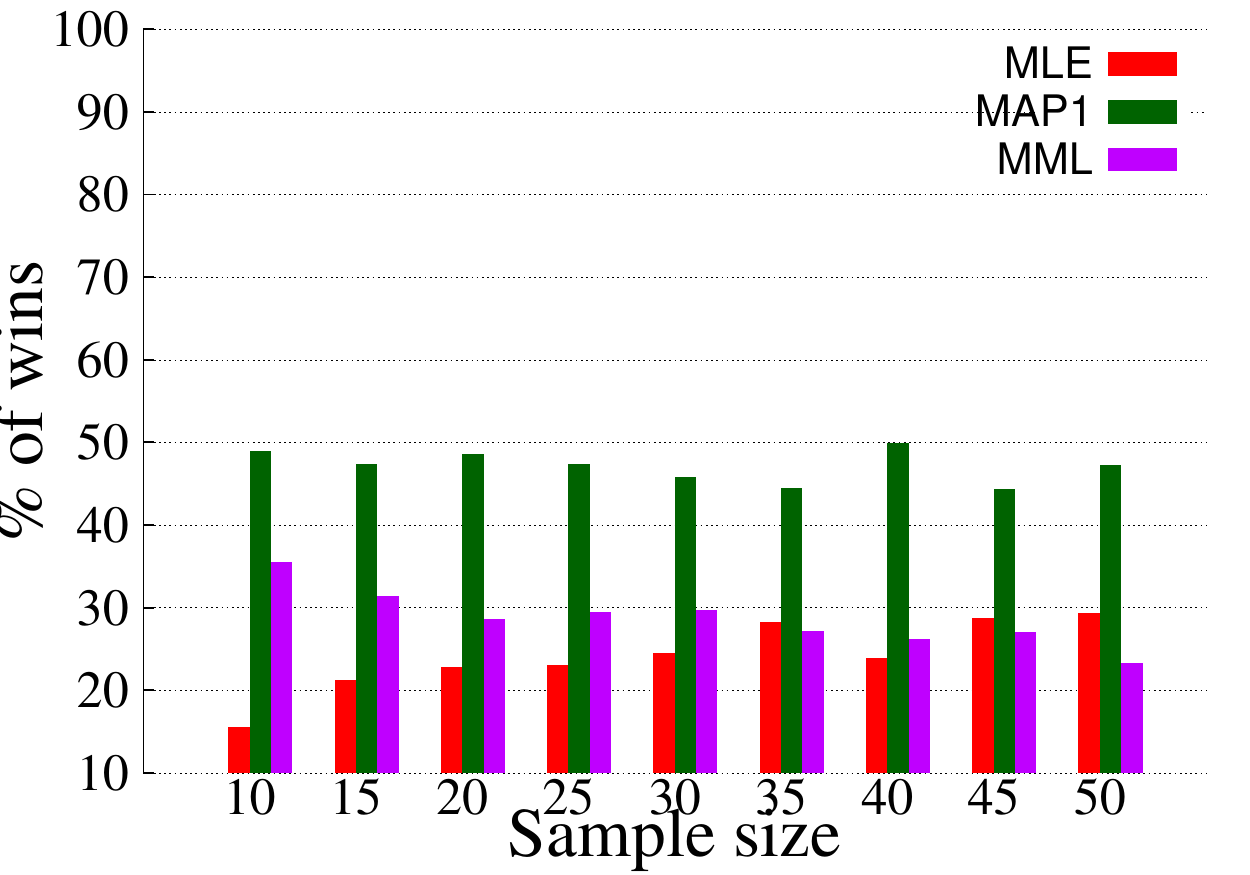}}
\subfloat[KL distance (MAP version 2)]{\includegraphics[width=0.33\textwidth]{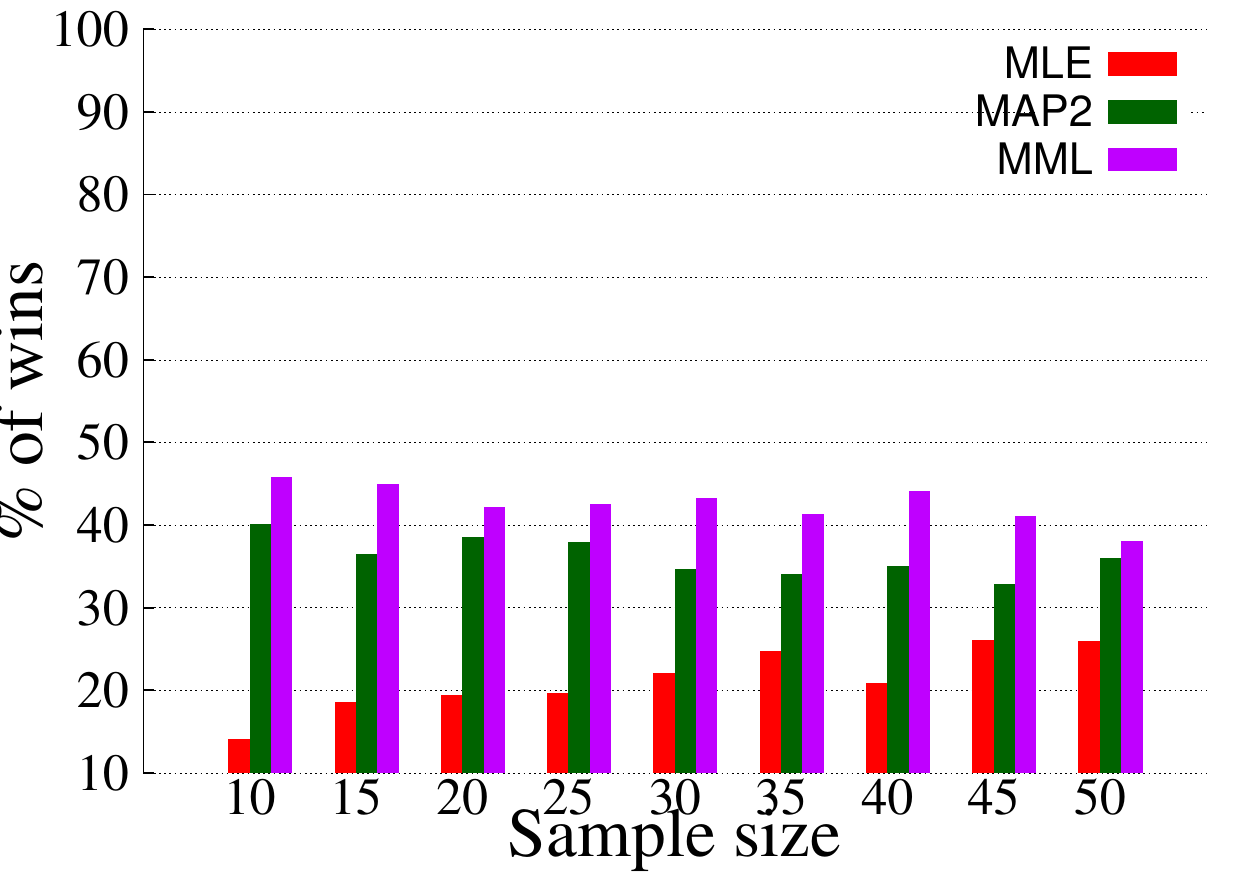}}
\subfloat[KL distance (MAP version 3)]{\includegraphics[width=0.33\textwidth]{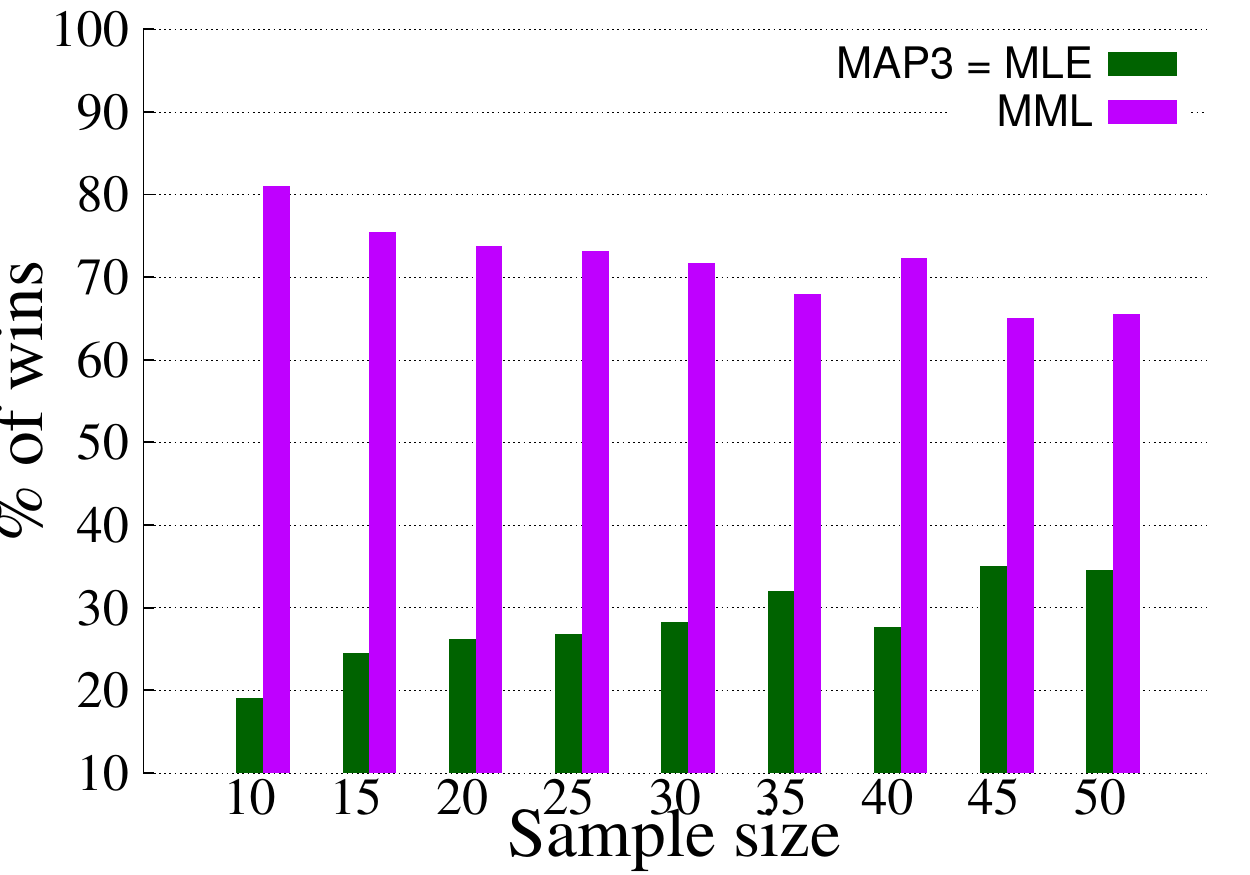}}\\
\subfloat[Variation of test statistics]{\includegraphics[width=0.5\textwidth]{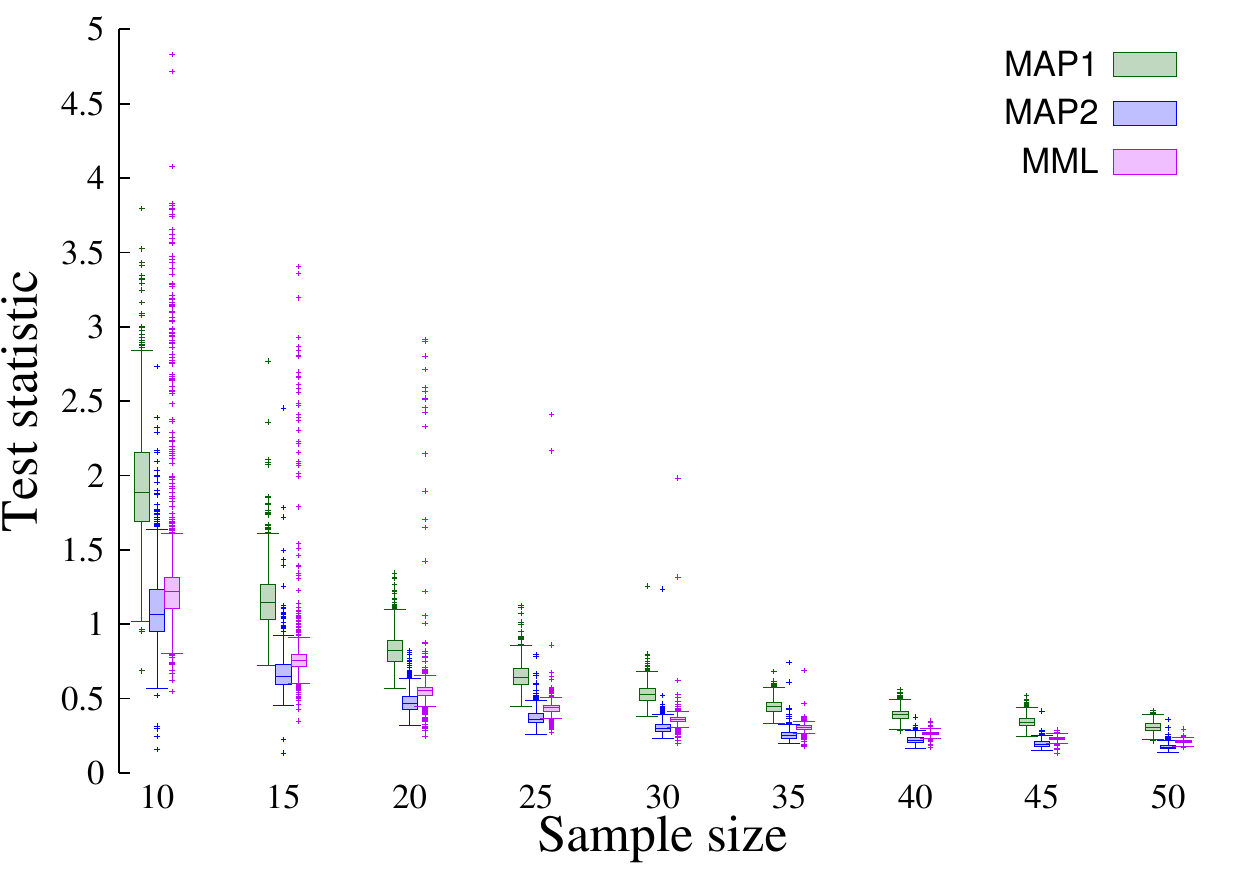}}
\subfloat[Variation of p-values]{\includegraphics[width=0.5\textwidth]{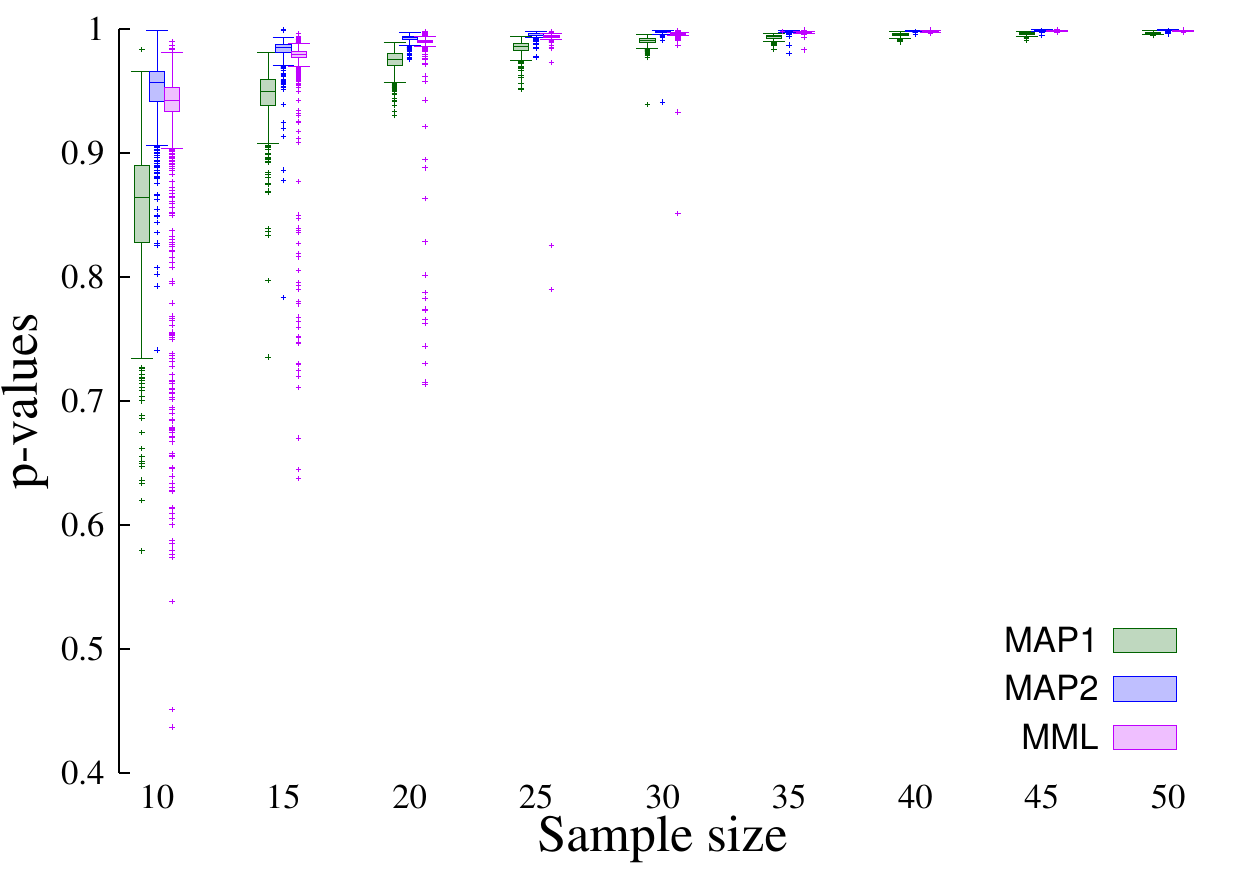}}
\caption
[Comparison of the parameter estimates when $\kappa_1=1,\kappa_2=10,\rho=0.5$.]
{Comparison of the parameter estimates when $\kappa_1=1,\kappa_2=10,\rho=0.5$.}
\label{fig:k1_1_k2_10_r_5}
\end{figure}

\noindent\textbf{For} $\pmb{\rho=0.5:}$
Similar to when $\rho=0.1$, we observe that the bias and MSE of the MAP
and MML estimators are lower than the ML estimators for different values of $N$.
In contrast to $\rho=0.1$, the bias of the MML estimator is lower than the MAP1
estimator but higher than the MAP2 estimator (Figure~\ref{fig:k1_1_k2_10_r_5}a). 
As with the previous case, MAP-based estimation result in  different
estimators. Further analysis of the estimators using KL distance 
and statistical hypothesis testing 
follow the same pattern as when $\rho=0.1$. \\

\begin{figure}[!htb]
\centering
\subfloat[Bias-squared]{\includegraphics[width=0.5\textwidth]{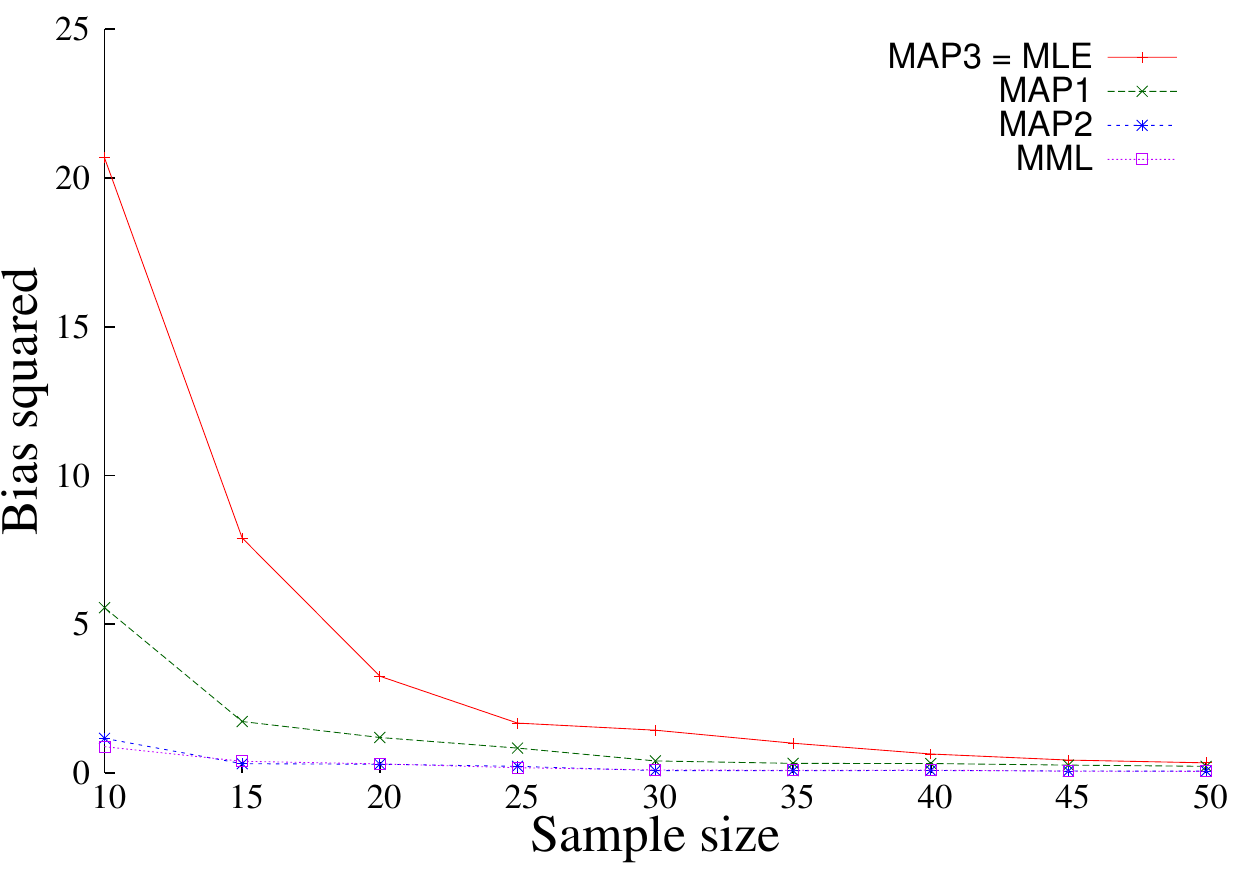}}
\subfloat[Mean squared error]{\includegraphics[width=0.5\textwidth]{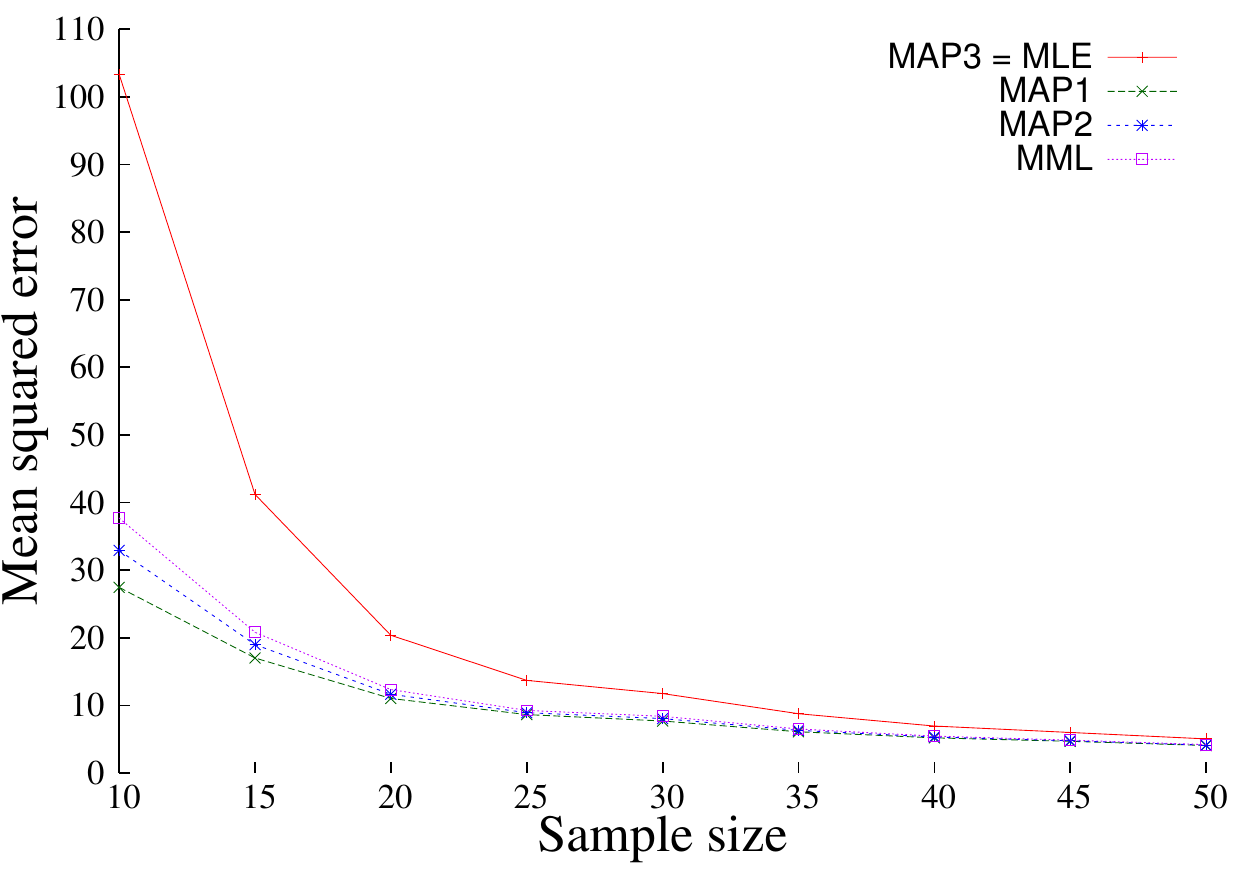}}\\
\subfloat[KL distance (MAP version 1)]{\includegraphics[width=0.33\textwidth]{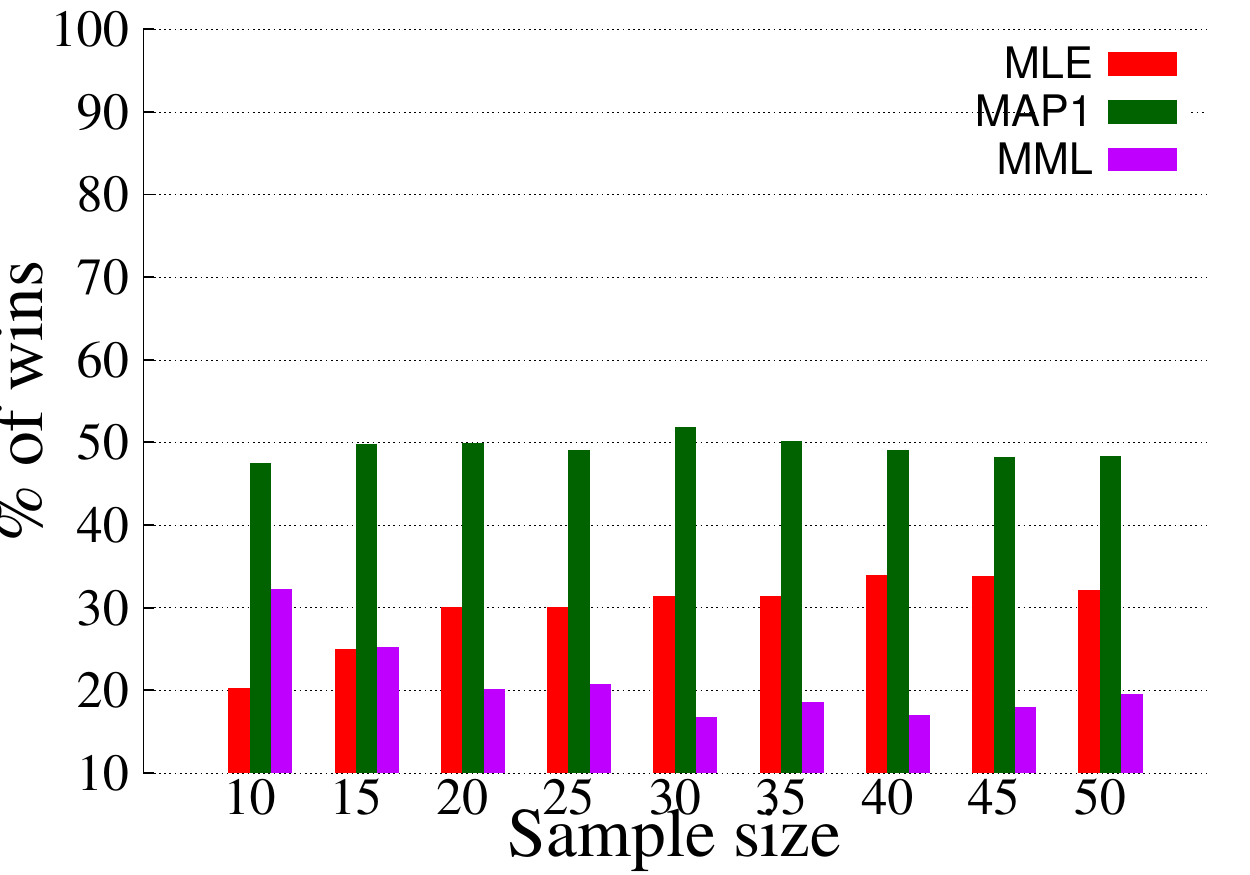}}
\subfloat[KL distance (MAP version 2)]{\includegraphics[width=0.33\textwidth]{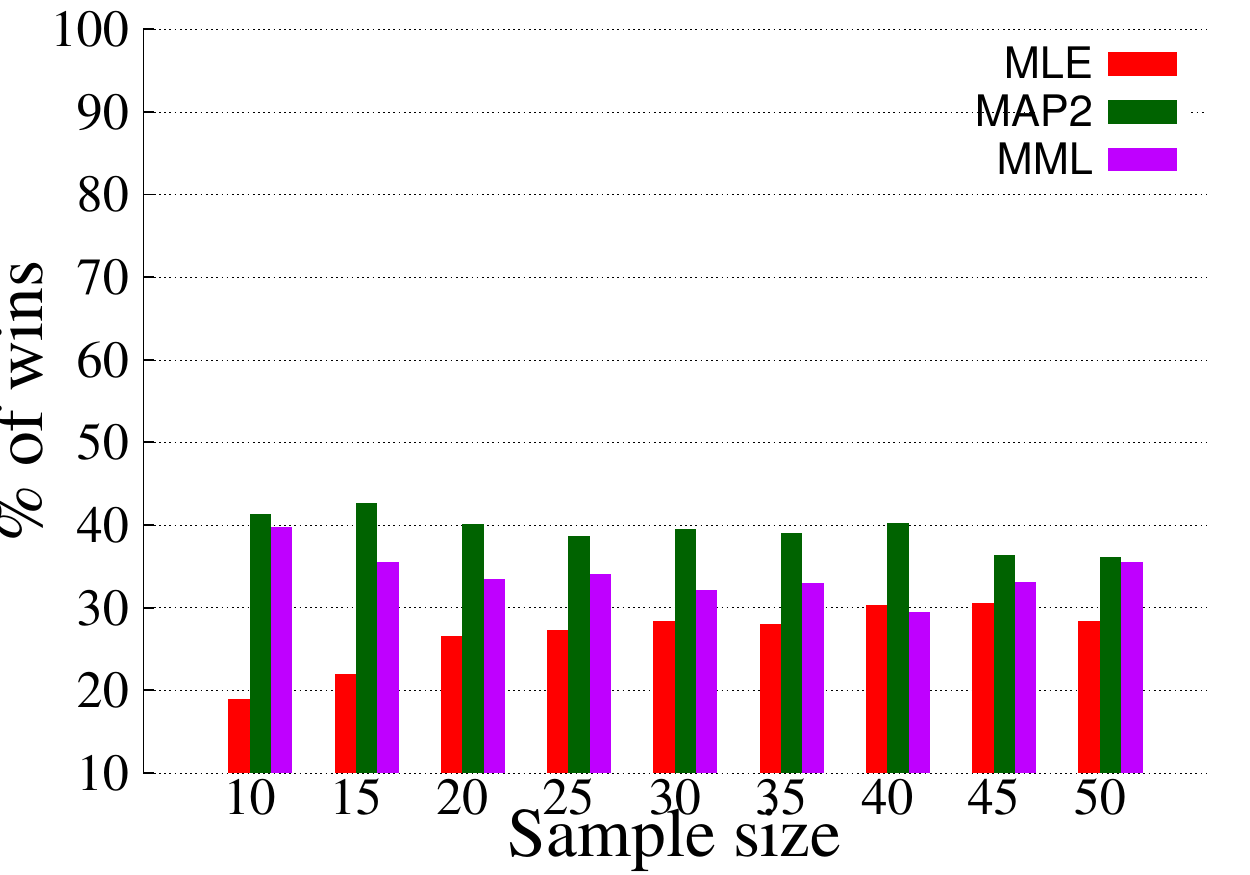}}
\subfloat[KL distance (MAP version 3)]{\includegraphics[width=0.33\textwidth]{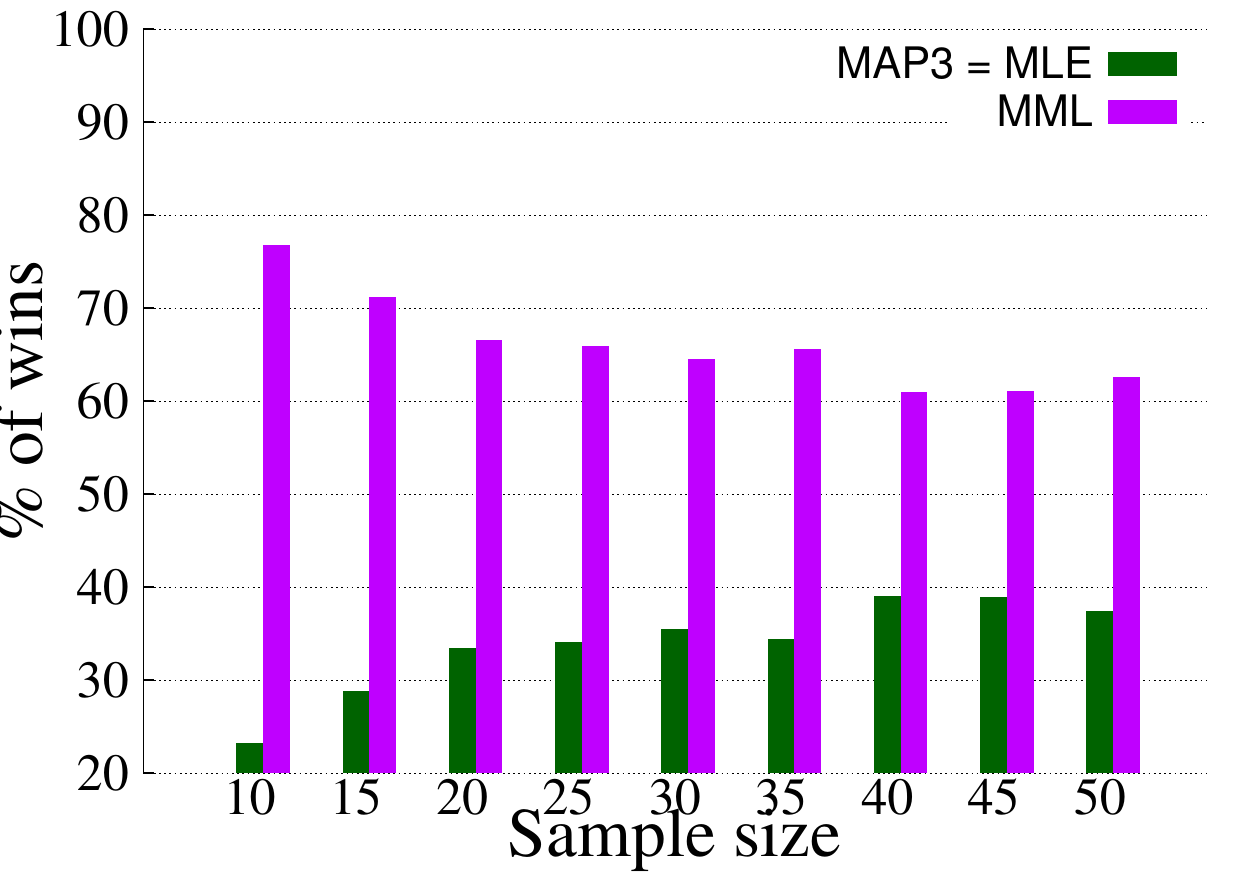}}\\
\subfloat[Variation of test statistics]{\includegraphics[width=0.5\textwidth]{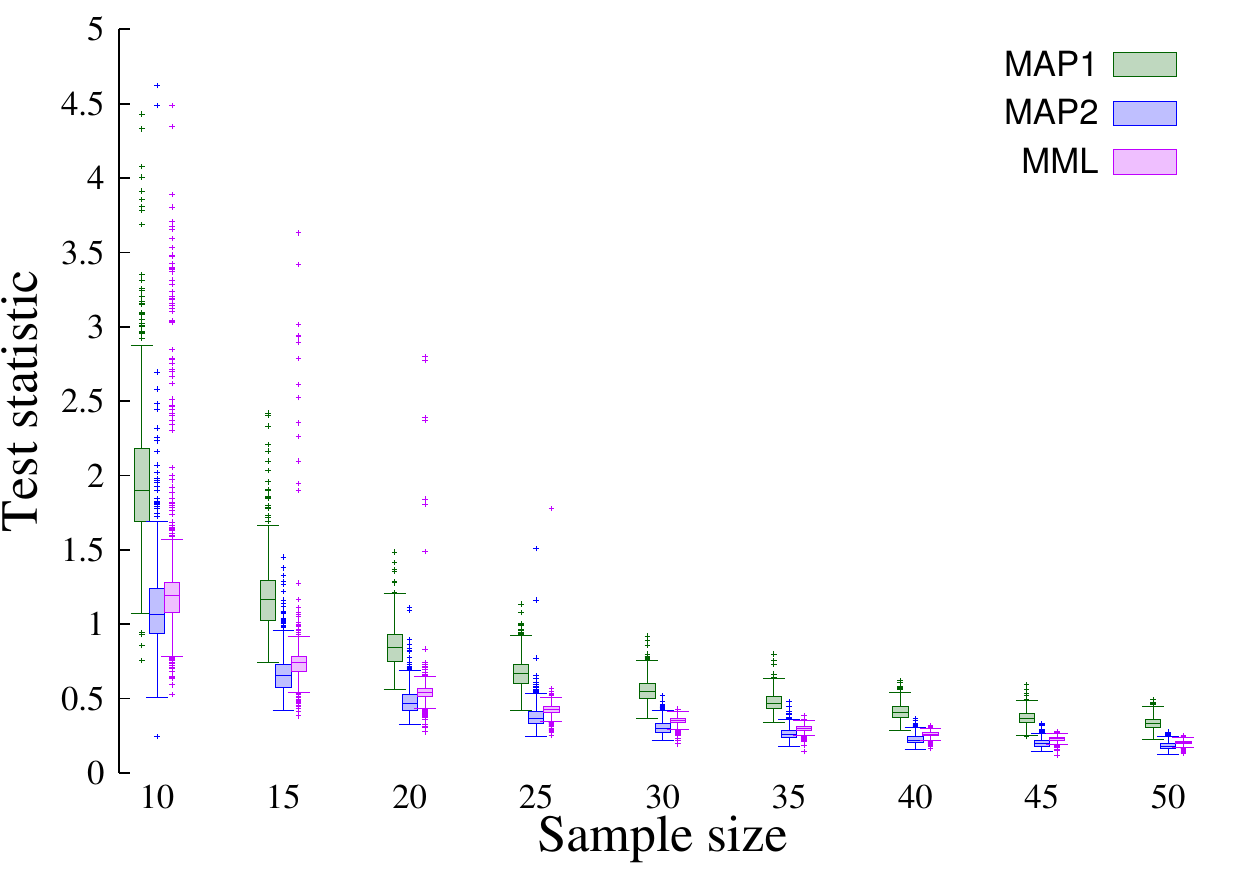}}
\subfloat[Variation of p-values]{\includegraphics[width=0.5\textwidth]{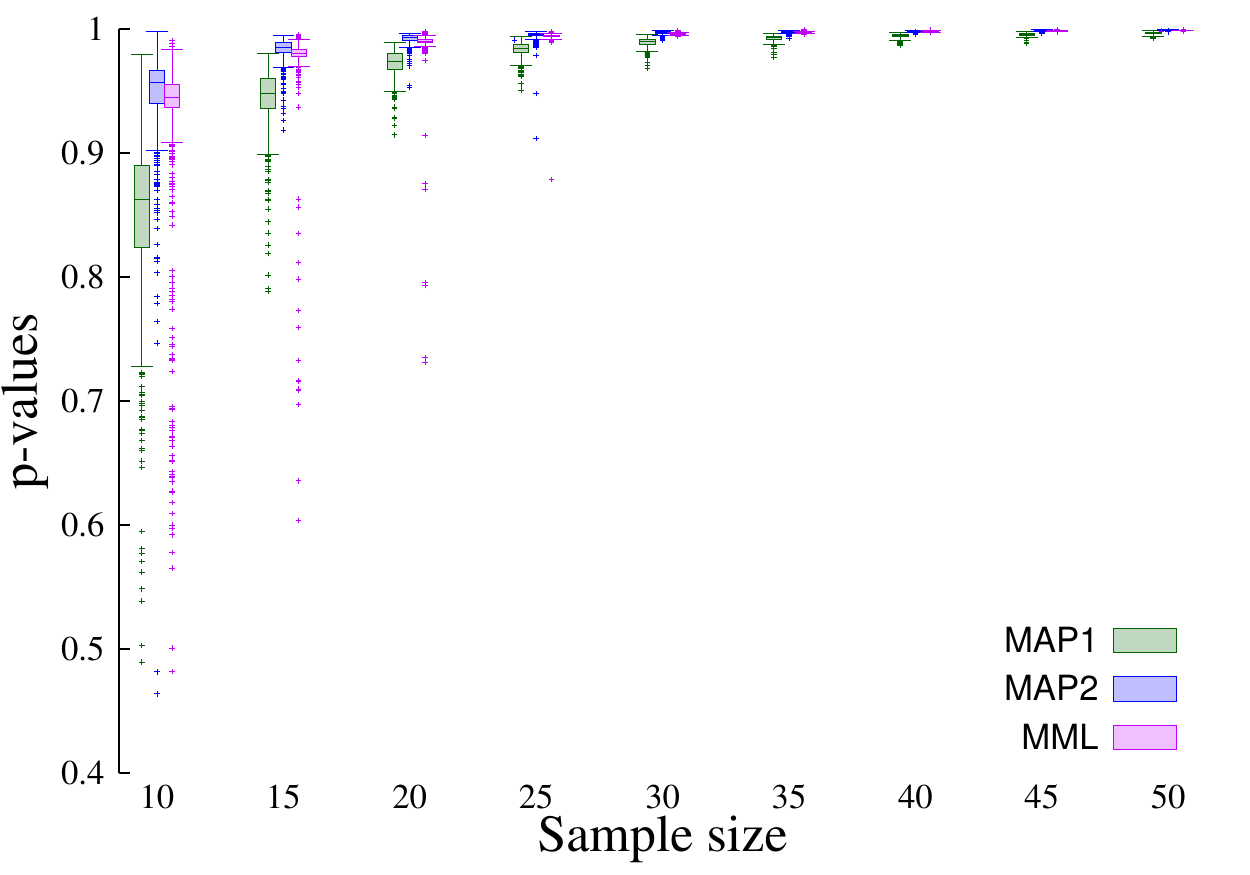}}
\caption
[Comparison of the parameter estimates when $\kappa_1=1,\kappa_2=10,\rho=0.9$.]
{Comparison of the parameter estimates when $\kappa_1=1,\kappa_2=10,\rho=0.9$.}
\label{fig:k1_1_k2_10_r_9}
\end{figure}

\noindent\textbf{For} $\pmb{\rho=0.9:}$
The results are presented in Figure~\ref{fig:k1_1_k2_10_r_9}.
As with the previous two cases, we observe that the ML estimators have the greatest
bias and MSE for all values of $N$.
The bias of the MML estimators is lower than all the MAP estimators. However,
the MSE of the MML estimators is greater compared to the MAP1 or MAP2 estimators.
Contrary to the previous two cases, we observe that the frequency of wins of KL distance
for the MML estimators is lower when compared to MAP2 estimation (Figure~\ref{fig:k1_1_k2_10_r_9}e).
Further, the results following the statistical hypothesis testing follow the same 
trend as the previous two cases. As the same size increases,
the different estimators converge to the ML estimators as seen from the
high p-values (Figure~\ref{fig:k1_1_k2_10_r_9}g).

The empirical analyses of the controlled experiments discussed above
indicate that the ML estimators of the parameters of a BVM distribution
are biased. The same was 
observed with other directional probability distributions such as
the vMF \citep{multivariate_vmf} and \fb~\citep{kent_arxiv}. Also, we
observe that the MAP estimation method result in different estimators
depending on how the distribution is parameterized. 
We have shown that the MAP estimators are not invariant
under non-linear transformations of the parameter space. 
In this context, the MML
estimators are empirically demonstrated to have lower bias than the 
traditional ML estimators and are invariant to alternative 
parameterizations unlike the MAP estimators.

\section{Mixtures of bivariate von Mises distributions}
\label{sec:bvm_sine_mixtures}

We consider two kinds of bivariate von Mises (BVM) distributions in 
mixture modelling.
In addition to the Sine variant (Equation~\ref{eqn:bvm_sine})
that has the correlation parameter $\lambda$, we also consider
the independent variant obtained when $\lambda = 0$. 
The independent version assumes zero correlation between the data
distributed on the torus (see Equation~\ref{eqn:bvm_ind}).
We provide a comparison for the mixture models obtained using both
versions of the BVM distributions. 

Previous work on MML-based modelling of protein dihedral angles
used independent BVM distributions \citep{dowe1996circular}.
Their work used the Snob mixture modelling software 
\citep{mml_classification}.
As pointed out by \citet{dowe1996circular}, Snob does not have
the functionality to account for the correlation between the data.
We therefore study the BVM Sine distributions and demonstrate
how they can be integrated with our generalized MML-based mixture modelling method.

\subsection{Approach for BVM distributions}
\label{subsec:bvm_mix_approach}

We extend the search method described in \citet{multivariate_vmf} 
to infer mixtures of BVM distributions. 
To infer the optimal number of mixture components, the mixture modelling apparatus
is now modified to handle the directional data distributed
on the surface of a torus.
As in the case of the vMF and \fb~distributions, the split operation detailed in 
\citet{kent_arxiv} is tailored for the BVM mixtures.
The basic idea behind splitting a parent component is to identify the means of the child components
so that they are on either side of the parent mean and are reasonably apart from each other.
Recall that for a Gaussian parent component, we computed the direction of maximum variance 
and selected the initial means, along this direction, that are one standard deviation away on either 
side of the parent mean. 
We employ the same strategy for BVM distributions. 
For data $\dataset=\{\boldx_1,\ldots,\boldx_N\}$, where $\boldx_i = (\phi_i,\psi_i)$
such that $\phi_i,\psi_i\in[-\pi,\pi)$,
we compute the direction of maximum variance in the $(\phi,\psi)$-space.
This allows us to compute the initial means of the child components.

The delete and merge operations are carried out in the same spirit.
During merging BVM components, the KL distance is evaluated to determine
the closest pair. We derive the KL distance for BVM Sine and BVM Independent 
distributions as shown in Appendix~\ref{app:bvm_sine_kldiv}.
Further, in all the operations, the MML estimators of the BVM Sine distribution,
derived in Section~\ref{subsec:bvm_sine_mml}
are used in the update step of the EM algorithm. 

\subsection{Mixture modelling of protein main chain dihedral angles}
\label{sec:bvm_modelling_dihedrals}

\begin{figure}[htb]
\centering
\subfloat[]{\includegraphics[width=0.7\textwidth]{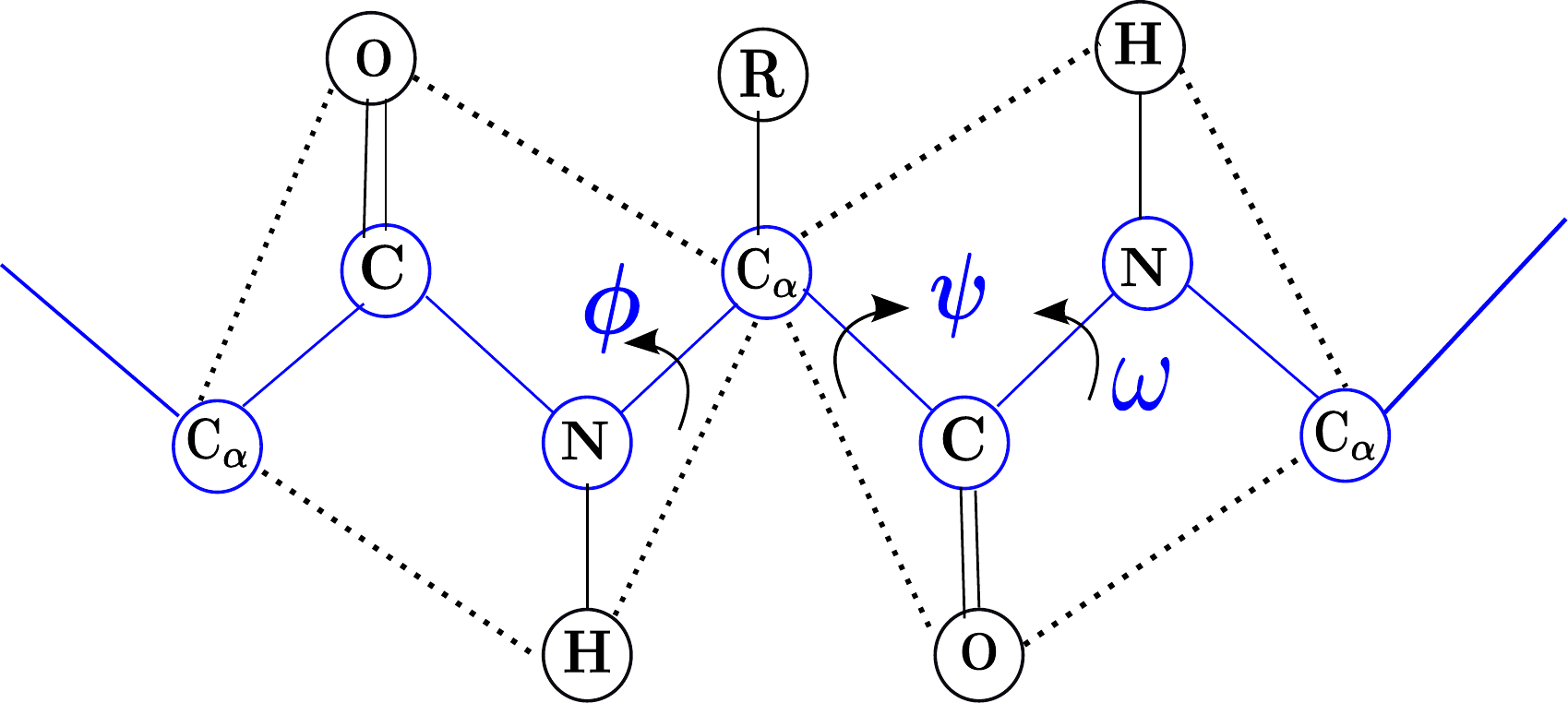}}\hfill
\subfloat[]{\includegraphics[width=0.22\textwidth]{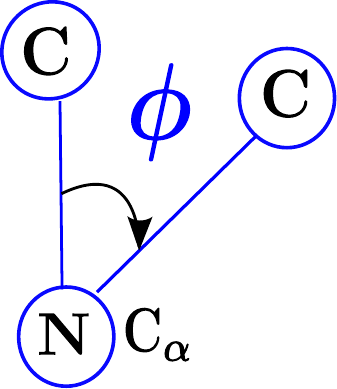}}
\caption[Protein main chain dihedral angles denoted by $(\phi,\psi)$.]
{Protein main chain dihedral angles denoted by $(\phi,\psi)$.}
\label{fig:protein_dihedrals}
\end{figure}

We consider the spatial orientations resulting from 
the interactions of the main chain atoms in protein structures.
A protein main chain is comprised of a chain of amino acids, each
os which is characterized by a central carbon \calpha.
The angular data corresponds to the spatial orientations of the 
the planes containing the atoms from successive amino acids.
A protein main chain is characterized by a sequence of 
$\phi,\psi,$ and $\omega$
angles. These angles uniquely determine the geometry of the protein
backbone structure \citep{Richardson1981167}.
However, in a majority of protein structures, $\omega=180^{\circ}$
and, hence the sequence of $\calpha$-C-N-$\calpha$ atoms lie in a plane
(see the dotted planar representation in Figure~\ref{fig:protein_dihedrals}a).
As a result, the angles $\phi$
and $\psi$ are typically analyzed \citep{ramachandran1963stereochemistry}.

The angles $\phi$ and $\psi$ are called the dihedral angle pair
corresponding to an amino acid residue with a central carbon atom $\calpha$
along the protein main chain.
Geometrically, a dihedral angle is the angle between any two planes
defined using four non-collinear points.
In Figure~\ref{fig:protein_dihedrals}(a), $\phi$ is the angle between the two planes
formed by C-N-\calpha~and N-\calpha-C. Similarly, 
$\psi$ is the angle between the two planes formed by
N-\calpha-C and \calpha-C-N.

The dihedral angles $\phi$ and $\psi$ are measured in a consistent manner.
For example, in order to measure $\phi$, the four atoms C-N-\calpha-C
are arranged such that $\phi$ is calculated as the deviation 
between N-C and \calpha-C when viewed in some consistent orientation.
As an illustration, in Figure~\ref{fig:protein_dihedrals}(b), view the
arrangement of the four atoms 
through the N-\calpha bond such that \calpha~is behind the plane of the paper
and N eclipses the \calpha~atom.
Also, the C atom directly attached to N is at the 12 o' clock position.
In this orientation, $\phi$ is given as the angle of rotation required
to align the N-C bond with the \calpha-C bond in the plane of the paper. Further, if it is
a clockwise rotation, it is considered a positive value. This ensures
that $\phi\in[-\pi,\pi)$. The dihedral angle $\psi$ is measured
by following the same convention with the four atoms being
N-\calpha-C-N.
\begin{figure}[!h]
\centering
\subfloat[Identifying the cross-section at $\phi$]{\includegraphics[width=0.4\textwidth]{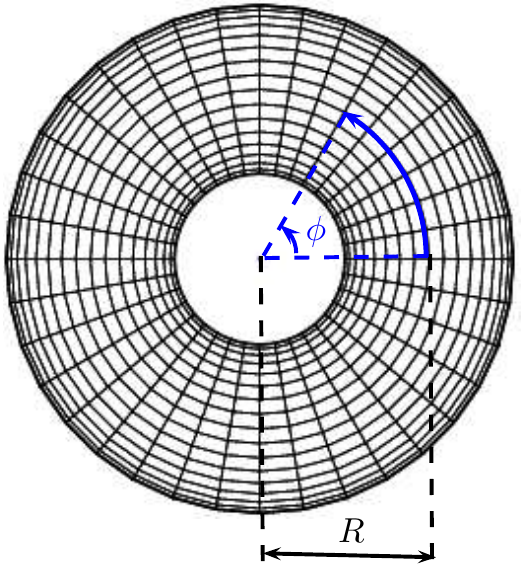}}\quad
\subfloat[Circular cross-section at $\phi$]{\includegraphics[width=0.385\textwidth]{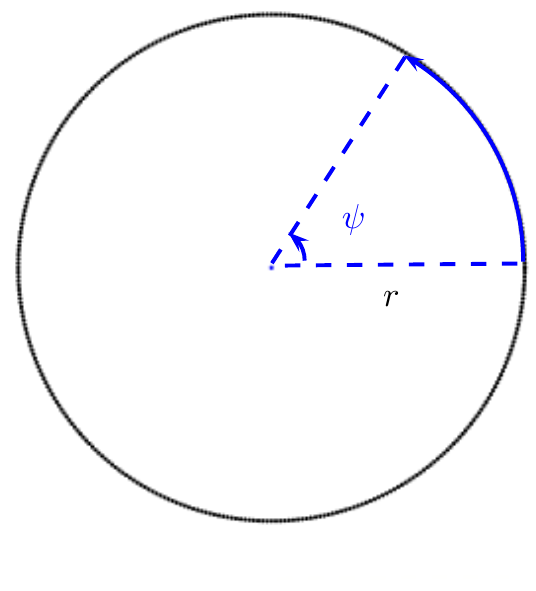}}
\caption{Representing a $(\phi,\psi)$ point on the torus.}
\label{fig:torus_phi_psi}
\end{figure} 
The $(\phi,\psi)$ pair measured in this way can be plotted on the surface
of a 3D torus. Each $(\phi,\psi)$ pair corresponds to a point on the toroidal surface.
The angle $\phi$ is used to identify a particular cross-section (circle)
of a torus,
while $\psi$ locates a point on this circle (see Figure~\ref{fig:torus_phi_psi}).

We generate the entire set of dihedral angle data from the  
1802 experimentally determined protein structures 
in the ASTRAL SCOP-40 (version 1.75) database \citep{murzin1995scop}
representing the ``$\beta$ class'' proteins. 
The number of $(\phi,\psi)$ dihedral angle pairs resulting from this data set is 253,165.
We model this generated set of dihedral angles using BVM Sine distributions.

\begin{figure}[!htb]
\centering
\includegraphics[width=0.4\textwidth]{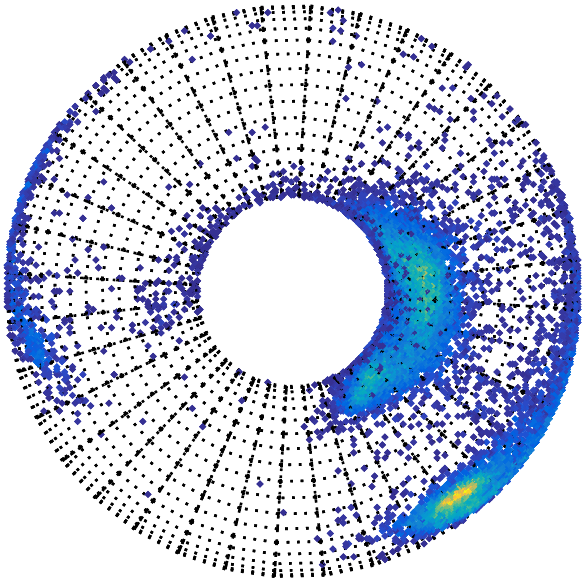}\quad
\includegraphics[width=0.4\textwidth]{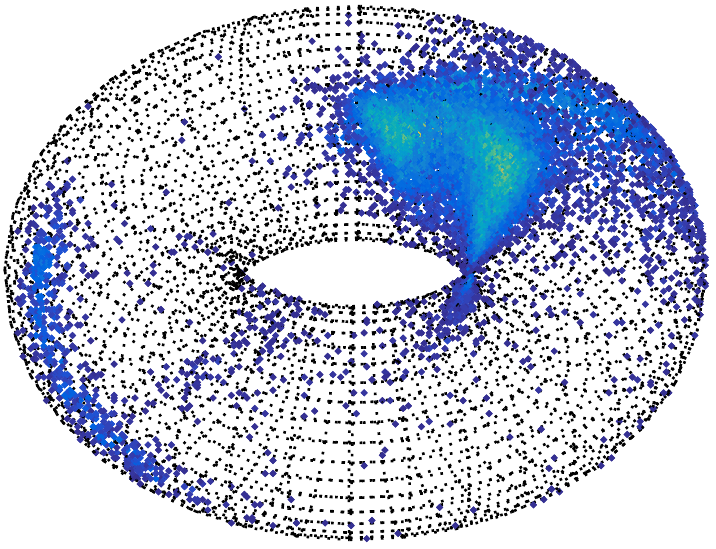}
\caption[Random sample from the empirical distribution on the 3D torus.]
{A sample of 10,000 points randomly generated from the empirical distribution
of ($\phi,\psi$) pairs. The figure shows the random sample from different viewpoints.}
\label{fig:empirical_distribution_torus}
\end{figure}

A random sample from this empirical distribution 
consisting of 10,000 points is shown in Figure~\ref{fig:empirical_distribution_torus}.
The plot is a heat map showing the density of the data distribution on
the toroidal surface.
Note that there are regions on the torus which are highly concentrated
(yellow), corresponding to the helical regions in the protein.
The ellipse-like patches (mostly in blue) roughly correspond
to the $\beta$ strands in proteins.
Furthermore, the data is multimodal which motivates its modelling 
using mixtures of BVM distributions.
We consider the effects of using the BVM Sine distribution as compared
to the BVM Independent variant in this context.

\subsubsection{Search of BVM Independent and BVM Sine mixtures}

The search method inferred a 32-component BVM Independent mixture and terminated
after 42 iterations involving split, delete, and merge operations. 
In the case of modelling using BVM Sine distributions,
our search method inferred
21 components and terminated after 29 iterations.
In each of these iterations, for every intermediate
$K$-component mixture, each constituent component is split, deleted, and merged
(with an appropriate component) to generate improved mixtures.

\begin{figure}[htb]
\centering
\subfloat[BVM Independent]{\includegraphics[width=0.45\textwidth]{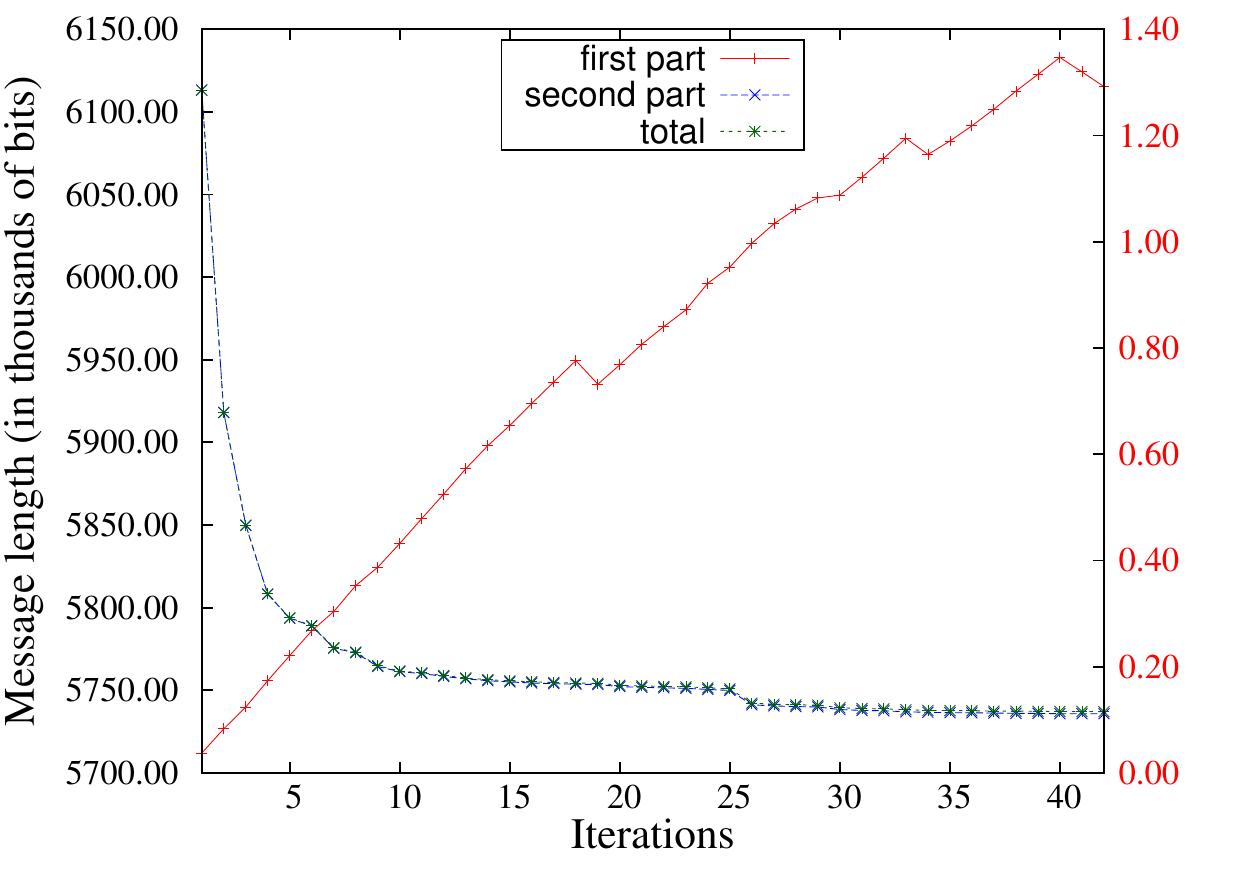}}
\subfloat[BVM Sine]{\includegraphics[width=0.45\textwidth]{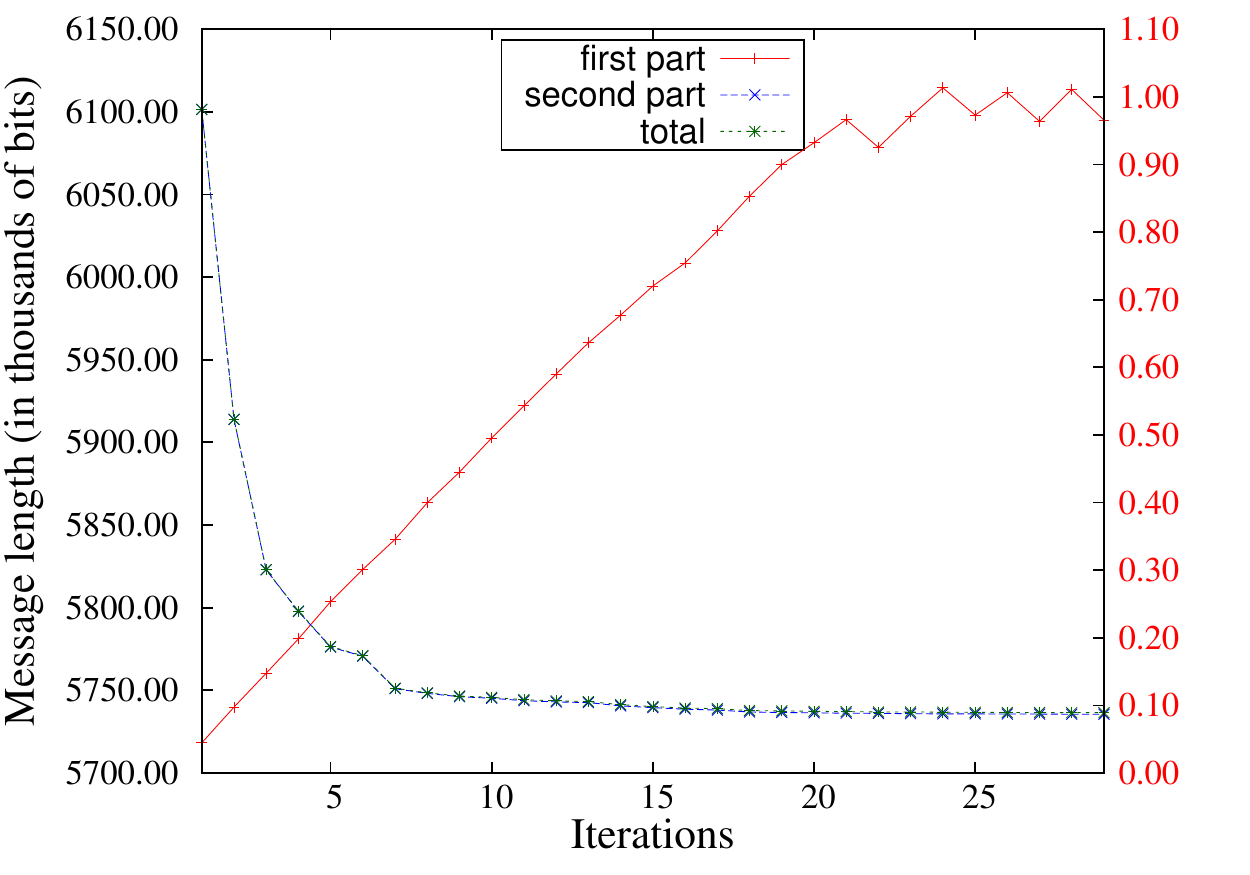}}
\caption
[Progression of the quality of the BVM mixtures inferred by our proposed search method.]
{Progression of the quality of the BVM mixtures inferred by our proposed search method.
         Note there are two Y-axes in both (a) and (b) with 
         different scales: the first part of the message follows the right side Y-axis (red);
         while the second part and total message lengths
         follow the left side Y-axis (black).
}
\label{fig:protein_bvm_mix_evolution}
\end{figure}

The progression of the search method for the optimal BVM Independent mixture
begins with a single component. The search method results in
continuous split operations until the $17^{\text{th}}$ iteration
when a 17-component mixture is inferred (see Figure~\ref{fig:protein_bvm_mix_evolution}a).
This corresponds to a progressive increase in the first part of the
message (red curve). Between the $17^{\text{th}}$ and the $21^{\text{st}}$
iterations, we observe a series of delete/merge and split operations
leading to a stable 19-component mixture. The search method
again continues to favour the split operations until the 
$28^{\text{th}}$ iteration when a 26-component mixture is inferred.
Thereafter, a series of deletions and splits yield a stable
29-component mixture at the end of the $35^{\text{th}}$ iteration.
The search method eventually terminates when a 32-component mixture
is inferred with a characteristic step-like behaviour towards the end
indicating perturbations involving split and delete/merge operations
(see Figure~\ref{fig:protein_bvm_mix_evolution}a).

In the case of searching for the optimal BVM Sine mixture,
our proposed search method continues to split
the components thereby increasing the mixture size. This occurs until
21 iterations. At this stage, there are 21 mixture components.
This can be observed in Figure~\ref{fig:protein_bvm_mix_evolution}(b),
when the first part of the message (red curve) continually increases
until the $21^{\text{st}}$ iteration. During this period, observe that the 
second part (blue) and the total message length (green) continually
decrease signifying an improvement to the mixtures.

After the $21^{\text{st}}$ iteration, we observe a step-like behaviour
as in the case of mixture modelling using the BVM Independent distributions.
The behaviour characterizes the reduction or increase in the number
of mixture components corresponding to a decrease or increase to the
first part of the message. After the $24^{\text{th}}$ iteration,
we observe that the mixture has 22 components. However, the final mixture
stabilizes in the subsequent iterations to a 21-component mixture.
After the $29^{\text{th}}$ iteration, there is no further improvement 
to the total message length and the search method terminates.

We observe a characteristic increase in the mixture size initially
followed by some perturbations that stabilize the intermediate 
mixture (step-like behaviour),
eventually resulting in an optimal mixture
(see Figure~\ref{fig:protein_bvm_mix_evolution}).
There is an initial sharp decrease in
the total message length until about 7 iterations for BVM mixtures.
Because of the multimodal nature of the directional data 
(see Figure~\ref{fig:empirical_distribution_torus}), the initial increase in the number 
of components would explain the data distribution
corresponding to those modes that are clearly distinguishable.
This leads to a substantial improvement to the total message length
as the minimal increase in the first part is dominated by the
gain in the second part.
However, towards the end of the search, 
when the increase in first part dominates
the reduction in second part, the method stops.
Thus, we see the trade-off of model complexity (as a function of the number
of components and their parameters), and the goodness-of-fit being
balanced using the search based on the MML inference framework.

\subsection{Comparison of BVM mixture models of protein data}

The existing work of MML-based mixture modelling of protein dihedral
angles by \citet{dowe1996circular} inferred 27 clusters using the BVM Independent distributions.
In contrast, our search method inferred 32 clusters.
However, their data consists of only 41,731 $(\phi,\psi)$ pairs generated 
from the protein structures known at that time.
In contrast, we have used 253,165 pairs of dihedral angles
along with a different search method
as explained previously (see Section~\ref{sec:bvm_modelling_dihedrals}).
So, there is some consensus on the rough number of component distributions
if the protein dihedral angles were modelled using BVM distributions
assuming no correlation between $\phi$ and $\psi$.

\begin{figure}[htb]
\centering
\includegraphics[width=0.75\textwidth]{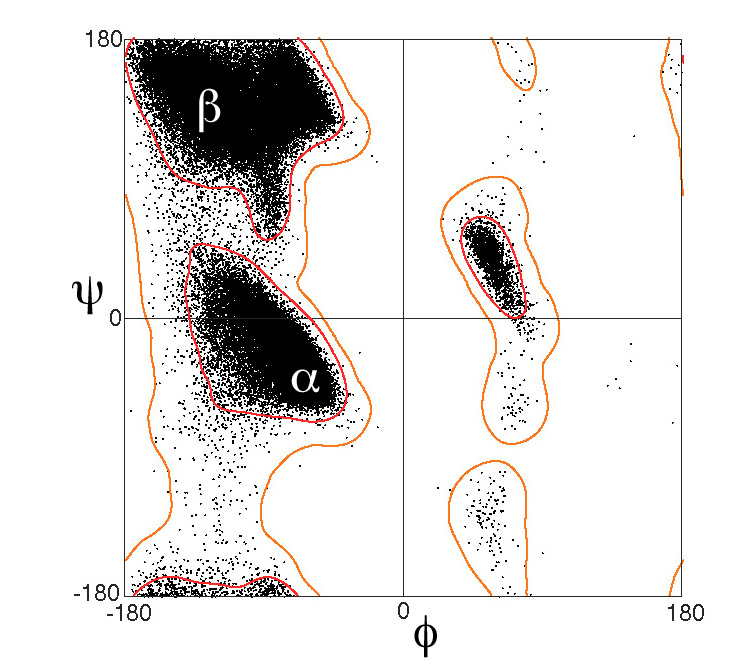} 
\caption[Models of the protein main chain dihedral angles.]
{Models of the protein main chain dihedral angles
($\phi$ and $\psi$ are in degrees).
Plot taken from \citet{lovell2003structure}.
}
\label{fig:dihedral_angles_lovell}
\end{figure}

The visualization of the dihedral angles is commonly done by 
the Ramachandran plot \citep{ramachandran1963stereochemistry}
who first analyzed the various possible protein configurations
and represented them as a two-dimensional plot.
An example of one such plot is provided in \citet{lovell2003structure}
and reproduced here (Figure~\ref{fig:dihedral_angles_lovell}). 
Such a plot is indicative of the allowed conformations that 
protein structures can adopt. There are vast spaces in the 
dihedral angle space where few data
are present. The conformations corresponding to those regions
are not possible. 
We consider the plot to explain the similarities between our inferred
mixture models and the one that is traditionally used.

\begin{figure}[!h]
\centering
\subfloat[BVM Independent MML mixture (32 components)]{\includegraphics[width=0.65\textwidth]{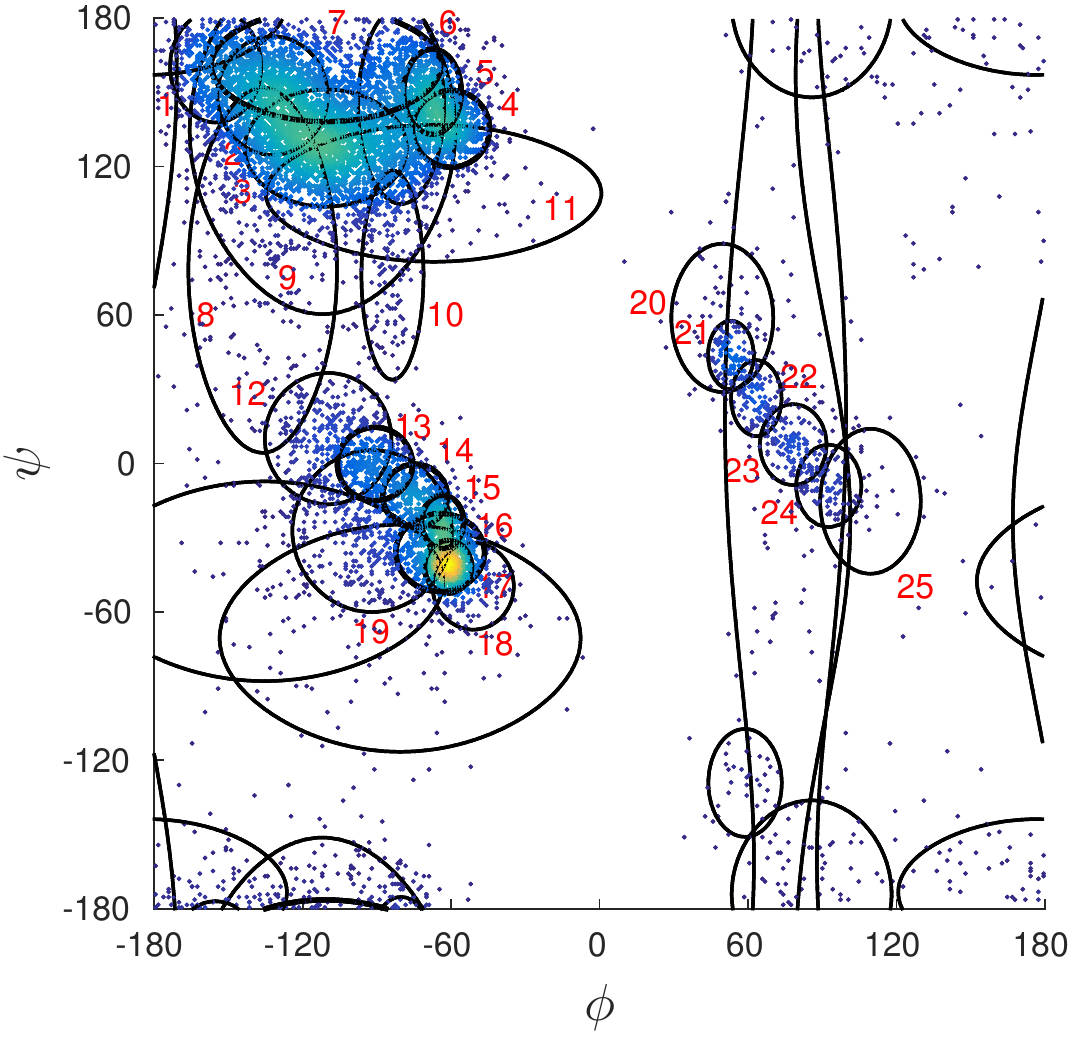}}\\
\subfloat[BVM Sine MML mixture (21 components)]{\includegraphics[width=0.65\textwidth]{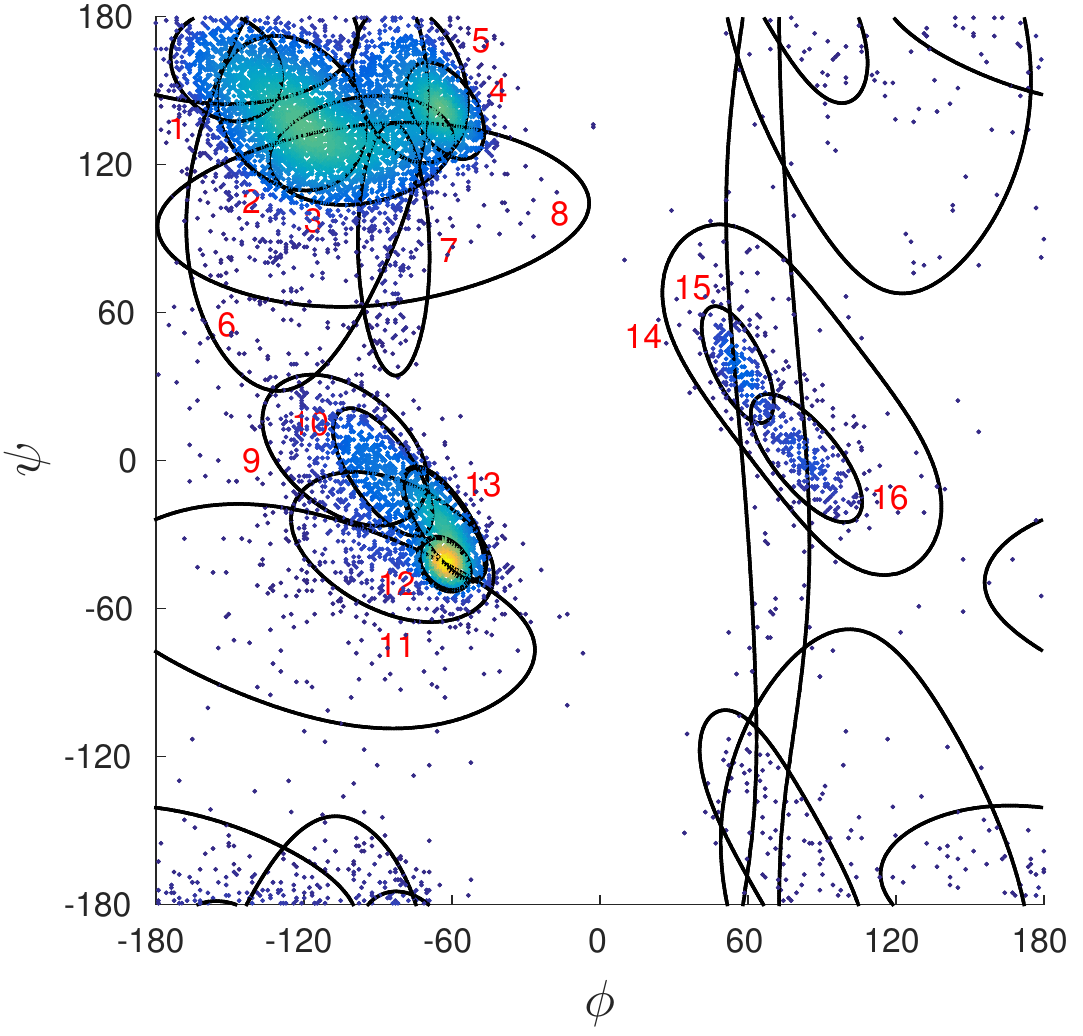}}
\caption[Models of the protein main chain dihedral angles.]
{Models of the protein main chain dihedral angles
($\phi$ and $\psi$ are in degrees).
}
\label{fig:dihedral_angles_bvm_mix}
\end{figure}

Our resulting mixtures of BVM Independent and the Sine variants
are shown in Figure~\ref{fig:dihedral_angles_bvm_mix}.
The contours of the constituent components plotted in the
$(\phi,\psi)$-space can be seen in the diagram.
For visualization purposes, we display 
the contour of each component that corresponds to 80\% of the data distribution.
The data in Figure~\ref{fig:dihedral_angles_bvm_mix}
corresponds to a random sample drawn from the empirical
distribution (same as in Figure~\ref{fig:empirical_distribution_torus}) 
visualized in the $(\phi,\psi)$-space.

In Figure~\ref{fig:dihedral_angles_lovell}, we observe that
the top-left region corresponds to the $\beta$ strands in protein structures.
The empirical distribution of dihedral angles we generated also has this characterstic.
We observe a concentrated mass in the top-left in Figure~\ref{fig:dihedral_angles_bvm_mix}.
Furthermore, our inferred mixtures are able to model this region using
the appropriate components. 
Note that smaller or highly compact contours
correspond to BVM distributions that have greater concentration 
parameters ($\kappa_1$ and $\kappa_2$ in Equation~\ref{eqn:bvm_sine}).

We note that components numbered 1-11 
(Figure~\ref{fig:dihedral_angles_bvm_mix}a) and components 1-8
(Figure~\ref{fig:dihedral_angles_bvm_mix}b)
are used to describe this region. 
These correspond to the components of the BVM Independent and
BVM Sine models respectively. Clearly, more number of components
are required to model roughly the same amount of data 
(corresponding to the $\beta$ strands) using the BVM Independent
mixture.

Similarly, in Figure~\ref{fig:dihedral_angles_lovell}, we observe
another concentration of mass in the middle-left portion of the figure.
This corresponds mainly to \emph{right-handed} $\alpha$-helices, which are very frequent
in protein structures. In Figure~\ref{fig:dihedral_angles_bvm_mix},
we have the corresponding mass and also note the dense region (bright yellow).
As per our inferred mixtures, component 17 (Figure~\ref{fig:dihedral_angles_bvm_mix}a)
and component 12 (Figure~\ref{fig:dihedral_angles_bvm_mix}b)
are used to predominantly describe this dense region. 
The other surrounding regions in the 
dihedral angle space of the right-handed helices are described by 
components 12-19 (Figure~\ref{fig:dihedral_angles_bvm_mix}a) and by
components 9-13 (Figure~\ref{fig:dihedral_angles_bvm_mix}b).
Again, we observe that the similar data is described using 8 components
by the BVM Independent mixture as opposed to 5 components
by the BVM Sine mixture.

\citet{lovell2003structure} display another region of concentrated mass
in the middle-right of Figure~\ref{fig:dihedral_angles_lovell}.
This region corresponds to the infrequent \emph{left-handed} helices in
protein structures. We see a corresponding mass in 
the empirical distribution in Figure~\ref{fig:dihedral_angles_bvm_mix}.
The components 20-25 of our inferred BVM Indenpendent mixture describe this region
(Figure~\ref{fig:dihedral_angles_bvm_mix}a).
The same region is described by components 14-16 of our inferred BVM Sine mixture 
(Figure~\ref{fig:dihedral_angles_bvm_mix}b).
Notice how this region is described by components 15 and 16.
These two components describe the dense mass within this region while 
component 14 is responsible for mainly modelling the data that is further away
from this clustered mass.
We again observe that the same region is modelled by greater number
of components when using BVM Independent distributions.

The remaining mixture components describe the insignificant mass present in
other regions of the dihedral angle space. 
The ability of our inferred mixtures to identify and describe specific regions of the protein
conformational space in a completely unsupervised setting is remarkable.
Further, we have qualified the effects of using the BVM distributions
which do not account for the correlation between the dihedral angle pairs. 
In this regard, the BVM Sine mixtures fare better when compared to mixtures
of BVM Independent distributions. We now quantify these effects
in terms of the total message length. 

Our proposed search method
to infer an optimal mixture involves evaluating the encoding cost
of the mixture parameters or the first part (model complexity), and encoding
the data using those parameters or the second part (goodness-of-fit).
The progression of the search method continues until there is no improvement to the
total message length. We observe that the resulting 21-component BVM Sine
mixture has a first part of 966 bits and a corresponding second part of
5.735 million bits (see Table~\ref{tab:bvm_mml_mix}). A BVM Independent
mixture with the same number of components has a first part of 872 bits and a
corresponding second part of 5.751 million bits. Although the model complexity
is lower for the BVM Independent mixture (difference of $\sim 94$ bits),
the BVM Sine mixture has an additional compression of $\sim 16,000$ bits
in its goodness-of-fit. Thus, the significant gain in the second part
dominates the minimal increase in the first part of the BVM Sine mixture.

Further, if we compare the 21-component BVM Independent mixture
with the inferred 32-component BVM Independent mixture,
we observe that the first part is more in the 32-component case. This is 
expected because there are more number of mixture parameters to encode
in the 32-component mixture. There is a difference of $1292 - 872 = 420$ bits
(see Table~\ref{tab:bvm_mml_mix}). However, the 32-component mixture results
in an extra compression of $\sim 15,000$ bits. So, the total message length
is lower for the 32-component mixture, and is therefore, preferred to the
21-component BVM Independent mixture.
\begin{table}[htb]
  \centering
  \caption{Message lengths of the BVM mixtures inferred on the protein dihedral angles.}
  \resizebox{\textwidth}{!}
  {
    \begin{tabular}{|c|c|c|c|c|}
    \hline
     Mixture        & Number of   &     First part        & Second part         &  Total message length                \\ \cline{4-5}
      model         & components  &  (thousands of bits)  &\multicolumn{2}{c|}{(millions of bits)} \\ \hline
    Independent     &   21        &   0.872               & 5.751         & 5.752                  \\
    Independent     &   32        &   1.292               & 5.736         & 5.737                  \\ 
    Sine            &   21        &   0.966               & 5.735         & \textbf{5.736}         \\ 
      \hline
    \end{tabular}
  }
  \label{tab:bvm_mml_mix}
\end{table}

When the inferred 32-component BVM Independent
and the 21-component BVM Sine mixtures are compared,
we observe that the total message length 
is lower for the BVM Sine mixture. In this case, both the first and 
second parts are lower for the Sine mixture leading to 
an overall gain of about $\sim 1000$ bits.
Thus, the BVM Sine mixture is more appropriate as compared to the BVM Independent
mixture in describing the protein dihedral angles.
This exercise shows how an optimal mixture model is selected
by achieving a balance between the
trade-off due to the complexity and the goodness-of-fit to the data.

Furthermore, as in the case of the vMF and \fb~distributions, we can devise null model
descriptions of protein dihedral angles based on the BVM mixtures.
For comparison, we consider a uniform distribution on the torus, which is referred
to as the uniform null model in the equation below.
\begin{equation*}
\text{Uniform Null} = -\log_2\left(\frac{\epsilon^2}{4\pi^2 Rr}\right) = 2\log_2(2\pi) - \log_2\left(\frac{\epsilon^2}{Rr}\right)\quad{\text{bits.}}
\end{equation*}
where $R$ and $r$ are the radii that define the size of the torus (see Figure~\ref{fig:torus_phi_psi}). 
When $R=r=1$, the surface area of the torus is $1/4\pi^2$.
The null models based on the BVM mixtures have the 
same form as the vMF and \fb~mixtures given as mixture distributions \citep{kent_arxiv} 
with the number of respective components being
$K=32$ and $K=21$ corresponding to the Independent and the Sine variants respectively.

Compared to the uniform model, both the BVM mixtures result in additional compression
(see Table~\ref{tab:bvm_sine_null_models}).
The message length to encode the entire collection of 253,165 dihedral angle pairs
using the uniform null model is 6.388 million bits which amounts to 25.234
bits per residue. In comparison, the BVM Independent mixture results in a
compression of 5.735 million bits which amounts to 22.656 bits per residue.
The additional compression is therefore, close to 2.58 bits per residue (on average).
The BVM Sine mixture further leads to an additional compression of
323 bits over the BVM Independent mixture.
This is equivalent to an additional saving of 0.0013 bits per residue (on average).
\begin{table}[htb]
  \centering
  \caption{Comparison of the null model encoding lengths based on the uniform distribution
on the torus,
the 32-component BVM Independent and the 21-component BVM Sine mixtures.}
  \begin{tabular}{|c|c|c|}
  \hline
  \multirow{2}{*}{Null model}      & Message length       & Bits per \\ 
                                   &  (in bits)           & residue \\ \hline
  Uniform                          & 6,388,508            & 25.2346 \\
  BVM Independent mixture          & 5,735,711            & 22.6560 \\
  BVM Sine mixture                 & \textbf{5,735,388}   & \textbf{22.6547} \\
  \hline
  \end{tabular}
  \label{tab:bvm_sine_null_models}
\end{table}

These results indicate that the BVM mixtures are superior compared to the uniform model.
This can be argued from the fact that the empirical distribution (see Figure~\ref{fig:empirical_distribution_torus})
has empty regions in the dihedral angle space. This is also confirmed
from the Ramachandran plot (Figure~\ref{fig:dihedral_angles_lovell}).
However, the BVM Independent and the BVM Sine variants are in close 
competition with each other. Noting that we need more mixture components in the Independent
case and because the Sine mixture can describe the data more effectively,
we conclude that the BVM Sine mixture supersedes the BVM Independent mixture.
The ability of the BVM Sine mixture to model correlated data leads to improved
description of the protein dihedral angles.

\section{Conclusion}

We have considered the problem of modelling directional
data using the bivariate von Mises distributions.
We have demonstrated that the MML-based estimation 
results in parameters that have a lower bias and MSE
compared to the traditional ML estimators, and contrast to
MAP estimators, they are invariant to transformations
of the parameter space.
To model empirically distributed data with multiple modes,
we have used mixtures of BVM distributions.
We have addressed the important problems of selecting
optimal number of mixture components along with their
parameters using the MML inference framework.
We employed the designed framework
to model protein dihedral angles using mixtures of
BVM distributions.
The empirical distribution of the pairs of dihedral angles represented on a toroidal surface
clearly suggests correlation between the angle pairs. As such,
the BVM Sine mixtures are shown to be appropriate. 
Both the BVM Independent and the Sine mixtures effectively model the
dihedral angle space. The ability of the search method to correctly identify
components corresponding to the regions of critical protein configurations
is remarkable. This is more so because our search method does not rely
on any prior information and infers the mixtures 
in a completely unsupervised setting.


\appendix
\section{Derivation of the KL distance between two BVM Sine distributions} 
\label{app:bvm_sine_kldiv}

The analytical form of the KL distance between two BVM Sine distributions is derived below.
For a datum $\boldx = (\theta_1,\theta_2)$, where $\theta_1,\theta_2\in[0,2\pi)$,
let $f_a(\boldx) = \text{BVM}(\mu_{a1},\mu_{a2},\kappa_{a1},\kappa_{a2},\lambda_a)$ 
and $f_b(\boldx) = \text{BVM}(\mu_{b1},\mu_{b2},\kappa_{b1},\kappa_{b2},\lambda_b))$ 
be two BVM Sine distributions
whose probability density functions
are given by Equation~\ref{eqn:bvm_sine}.
Let $c_a$ and $c_b$ be their respective normalization constants,
whose expressions are given by Equation~\ref{eqn:bvm_sine_norm_constant}. 
The computation of the BVM Sine normalization constant
is presented in Section~\ref{subsec:bvm_sine_norm_constant_derivatives}.

The KL distance between two probability distributions $f_a$ and $f_b$ is defined by 
$\expect_a\left[\log\dfrac{f_a(\boldx)}{f_b(\boldx)}\right]$.
Using the density function in Equation~\ref{eqn:bvm_sine}, we have
\begin{equation*}
\expect_a[\log f_a(\boldx)] = -\log c_a 
+ \kappa_{a1} \expect_a [\cos(\theta_1 - \mu_{a1})] 
+ \kappa_{a2} \expect_a [\cos(\theta_2 - \mu_{a2})]
+ \lambda_a \expect_a [\sin(\theta_1 - \mu_{a1})\sin(\theta_2 - \mu_{a2})]
\end{equation*}
The expressions for the above expectation terms $\expect_a[\cos(\theta_1 - \mu_{a1})]$, 
$\expect_a[\cos(\theta_2 - \mu_{a2})]$
and $\expect_a[\sin(\theta_1 - \mu_{a1})\sin(\theta_2 - \mu_{a2})]$
can be computed and
are given by Equation~\ref{eqn:expect_sincos1}.
Similarly, the expectation of $\log f_b(\boldx)$ is
\begin{equation*}
\expect_a[\log f_b(\boldx)] = -\log c_b 
+ \kappa_{b1} \expect_a [\cos(\theta_1 - \mu_{b1})] 
+ \kappa_{b2} \expect_a [\cos(\theta_2 - \mu_{b2})]
+ \lambda_b \expect_a [\sin(\theta_1 - \mu_{b1})\sin(\theta_2 - \mu_{b2})]
\end{equation*}

In order to compute $\expect_a [\cos(\theta_1 - \mu_{b1})]$, we express
$\cos(\theta_1 - \mu_{b1})$ as 
\begin{align*}
\cos(\theta_1 - \mu_{b1}) &= \cos(\theta_1 - \mu_{a1} + \mu_{a1} - \mu_{b1}) \\
&= \cos(\theta_1 - \mu_{a1}) \cos(\mu_{a1} - \mu_{b1}) -  \sin(\theta_1 - \mu_{a1}) \sin(\mu_{a1} - \mu_{b1}) 
\end{align*}
Given that $\expect_a [\sin(\theta_1 - \mu_{a1})] = 0$ (Equation~\ref{eqn:expect_sincos1}), we have
\begin{align*}
\expect_a [\cos(\theta_1 - \mu_{b1})] 
&=\cos(\mu_{a1} - \mu_{b1})\,\expect_a[\cos(\theta_1 - \mu_{a1})] \\
\text{Similarly,}\,\,
\expect_a [\cos(\theta_2 - \mu_{b2})] 
&=\cos(\mu_{a2} - \mu_{b2})\,\expect_a[\cos(\theta_2 - \mu_{a2})] 
\end{align*}

In order to compute $\expect_a [\sin(\theta_1 - \mu_{b1})\sin(\theta_2 - \mu_{b2})]$,
we express the product of the sine terms as 
\begin{equation*}
\sin(\theta_1 - \mu_{b1})\sin(\theta_2 - \mu_{b2}) 
= \sin(\theta_1 - \mu_{a1} + \mu_{a1} - \mu_{b1})\sin(\theta_2 - \mu_{a2} + \mu_{a2} - \mu_{b2}) 
\end{equation*}
Further, using the property that 
$\expect_a[\cos(\theta_1-\mu_{a1})\sin(\theta_2-\mu_{a2})]
= \expect[\sin(\theta_1-\mu_{a1})\cos(\theta_2-\mu_{a2})] = 0$ (Equation~\ref{eqn:expect_sincos2}),
we have
\begin{align*}
\expect_a [\sin(\theta_1 - \mu_{b1})\sin(\theta_2 - \mu_{b2})] 
&= \cos(\mu_{a1} - \mu_{b1})\cos(\mu_{a2} - \mu_{b2}) \expect_a [\sin(\theta_1 - \mu_{a1})\sin(\theta_2 - \mu_{a2})] \\
&+ \sin(\mu_{a1} - \mu_{b1})\sin(\mu_{a2} - \mu_{b2}) \expect_a [\cos(\theta_1 - \mu_{a1})\cos(\theta_2 - \mu_{a2})] 
\end{align*}

Then, the KL distance between the two distributions $f_a$ and $f_b$ is derived as 
\begin{align}
\expect_a\left[\log\frac{f_a(\boldx)}{f_b(\boldx)}\right] &= 
\log\frac{c_b}{c_a}
+ \{\kappa_{a1} - \kappa_{b1}\cos(\mu_{a1} -\mu_{b1})\}\,\expect_a[\cos(\theta_1 - \mu_{a1})] \notag\\
&+ \{\kappa_{a2} - \kappa_{b2}\cos(\mu_{a2} -\mu_{b2})\}\,\expect_a[\cos(\theta_2 - \mu_{a2})] \notag\\
&+ \{\lambda_a - \lambda_b \cos(\mu_{a1} -\mu_{b1}) \cos(\mu_{a2} -\mu_{b2})\}
\,\expect_a[\sin(\theta_1 - \mu_{a1})\sin(\theta_2 - \mu_{a2})]\notag\\
&- \lambda_b\sin(\mu_{a1} -\mu_{b1}) \sin(\mu_{a2} -\mu_{b2})
\label{eqn:bvm_sine_kldiv}
\end{align}
gives the analytical form of the KL distance of two BVM Sine distributions. \\

\noindent\emph{Special case ($\lambda=0$):} 
The BVM Sine distribution reduces to the product of two individual
von Mises circular distributions given by Equation~\ref{eqn:bvm_ind}.
To compute the KL distance between two BVM Independent distributions,
we can use Equation~\ref{eqn:bvm_sine_kldiv}, with $\lambda=0$.
Note that for the von Mises circular distribution, the normalization
constant is
$C(\kappa) = \dfrac{1}{2\pi I_0(\kappa)}$, where
$I_0(\kappa)$ and $I_1(\kappa)$ are the modified Bessel functions.
The KL distance between the BVM Independent distributions $f_a$ and 
$f_b$ is then given by
\begin{align}
\expect_a\left[\log\frac{f_a(\boldx)}{f_b(\boldx)}\right] &= 
\log\frac{I_0(\kappa_{b1})}{I_0(\kappa_{a1})} 
+ \frac{I_1(\kappa_{a1})}{I_0(\kappa_{a1})} 
\{ \kappa_{a1} -\kappa_{b1} \cos (\mu_{a1} - \mu_{b1}) \} \notag\\
&+ \log\frac{I_0(\kappa_{b2})}{I_0(\kappa_{a2})} 
+ \frac{I_1(\kappa_{a2})}{I_0(\kappa_{a2})} 
\{ \kappa_{a2} -\kappa_{b2} \cos (\mu_{a2} - \mu_{b2}) \} 
\label{eqn:bvm_ind_kldiv}
\end{align}

\end{document}